\documentclass{article}

\usepackage[final]{neurips_2024}
\makeatletter
\renewcommand{\footnoterule}{
  \kern-3pt
  \hrule width \textwidth height 0.4pt
  \kern 2.6pt
}
\makeatother
\usepackage{soul}
\usepackage{lineno}
\modulolinenumbers[5]
\usepackage{units}
\usepackage{tikz}
\usetikzlibrary{positioning}
\usepackage{algorithm}
\usepackage{enumitem}
\usepackage{float}
\usepackage{algpseudocode}
\usepackage{graphicx}
\usepackage{varioref}
\usepackage{longtable,pdflscape,booktabs}
\usepackage{array}
\usepackage{amsmath}
\usepackage{booktabs}
\usepackage{multirow}
\usepackage{subcaption}
\usepackage{array,arydshln}
\usepackage{amssymb}
\usepackage{hhline}
\usepackage{times,rotating}
\usepackage{comment}
\usepackage{caption}
\usepackage[hyphens,spaces,obeyspaces]{url}
\usepackage{tikz}
\usepackage{svg}
 
 \tikzset{variable/.default=}
\usepackage{tabularx,ragged2e}
\newcolumntype{T}[1]{>{\raggedright\arraybackslash}p{#1}}
\newcolumntype{M}[1]{>{\centering\arraybackslash}m{#1}}

\newcolumntype{L}[1]{>{\raggedright\let\newline\\\arraybackslash\hspace{0pt}}m{#1}}
\newcolumntype{C}[1]{>{\centering\let\newline\\\arraybackslash\hspace{0pt}}m{#1}}
\newcolumntype{R}[1]{>{\raggedleft\let\newline\\\arraybackslash\hspace{0pt}}m{#1}} 

\usepackage{tikz}
\usepackage{forest}
\usetikzlibrary{positioning}
\useforestlibrary{linguistics}
\forestapplylibrarydefaults{linguistics}
\usepackage{pifont}

\usepackage{framed} 
\usepackage{multicol} 
\usepackage{nomencl} 
\usepackage{booktabs}       
\usepackage{amsfonts}       
\usepackage{nicefrac}       
\usepackage{microtype}      
\usepackage{diagbox} 
\usepackage{dsfont}
\usepackage{bm} 
\usepackage{amsmath}
\usepackage{graphicx}
\usepackage{amsfonts}
\usepackage{subcaption}
\usepackage{stmaryrd}
\usepackage{color, colortbl}
\usepackage{breakcites}
\usepackage{pdfpages}
\usepackage{rotating}
\usepackage{dirtree} 

\usepackage[utf8]{inputenc} 
\usepackage[T1]{fontenc}    
\usepackage{hyperref} \hypersetup{
  colorlinks=true,
  linkcolor=black,
  citecolor=black,
  filecolor=black,
  urlcolor=black,
  pdfborder={0 0 0}
}      
\usepackage{url}            
\usepackage{booktabs}       
\usepackage{amsfonts}       
\usepackage{nicefrac}       
\usepackage{microtype}      
\usepackage{xcolor}         

\title{Explainability of Large Language Models: Opportunities and Challenges toward Generating Trustworthy Explanations}

\author{%
  Shahin Atakishiyev\thanks{Corresponding author. Email: shahin.atakishiyev@ualberta.ca} \\
  University of Alberta
   \And
   Housam K.B. Babiker \\
   University of Alberta
   \And
   Jiayi Dai \\
   University of Alberta
   \And
   Nawshad Farruque \\
   University of Alberta\\
   \And
   Teruaki Hayashi \\
   University of Tokyo
   \And
   Nafisa Sadaf Hriti \\
   University of Alberta
   \And
   Md Abed Rahman \\
   University of Alberta
   \And
   Iain Smith \\
   University of Alberta
   \And
   Mi-Young Kim \\
   University of Alberta
   \And
   Osmar R. Zaïane \\
   University of Alberta
   \And
   Randy Goebel \\ 
   University of Alberta
}

\begin{document}

\maketitle

\begin{abstract}
  Large language models have exhibited impressive performance across a broad range of downstream tasks in natural language processing. However, how a language model predicts the next token and generates content is not generally understandable by humans. Furthermore, these models often make errors in prediction and reasoning, known as hallucinations. These errors underscore the urgent need to better understand and interpret the intricate inner workings of language models and how they generate predictive outputs. Motivated by this gap, this paper investigates local explainability and mechanistic interpretability within Transformer-based large language models to foster trust in such models. In this regard, our paper aims to make three key contributions. First, we present a review of local explainability and mechanistic interpretability approaches and insights from relevant studies in the literature. Furthermore, we describe experimental studies on explainability and reasoning with large language models in two critical domains --- healthcare and autonomous driving --- and analyze the trust implications of such explanations for explanation receivers. Finally, we summarize current unaddressed issues in the evolving landscape of LLM explainability and outline the opportunities, critical challenges, and future directions toward generating human-aligned, trustworthy LLM explanations.
\end{abstract}

\section{Introduction} \label{sec:intro}
Pretrained large language models (LLMs) have emerged as powerful tools for a range of complex natural language processing (NLP) applications in the field of artificial intelligence (AI). Primarily based on the Transformer architecture \cite{vaswani2017attention}, recent LLMs, such as the family series from GPT \cite{brown2020language, achiam2023gpt, gpt5systemcard2025},  Vicuna \cite{vicuna2023}, LLaMA \cite{touvron2023llama, touvron2023llama2}, Mistral \cite{jiang2023mistral}, Claude \cite{anthropic_claude, anthropic2025claude45}, Qwen \cite{bai2023qwen, team2024qwen2, yang2025qwen3}, Gemini \cite{team2023gemini, comanici2025gemini}, DeepSeek \cite{bi2024deepseek, liu2024deepseek,liu2024v3deepseek, guo2025deepseek}, Grok \cite{xai2024grok, xai2024grok2, xai2025grok4} and Kimi \cite{team2025kimi1.5, team2025kimik2} have demonstrated impressive performance across many NLP tasks, and the list continues to expand. These models typically use a Transformer-based deep neural network as a base model and are augmented with an extended number of parameters, a larger volume of data, and an increased overall model size. Numerous empirical studies show that larger-size LLMs substantially outperform smaller models in terms of their emergent abilities, such as reasoning, instruction following, program execution, and general-purpose task-solving capabilities \cite{brown2020language, driess2023palm, wei2022emergent}. \\
While LLMs' emergent abilities are impressive at first glance, there are several issues with these tools \cite{singh2024rethinking}. First, almost all state-of-the-art LLMs have a common problem of producing \textit{hallucinations}, i.e., they produce convincing but falsified, factually incorrect content. Depending on the context, LLM-generated false statements can be detected easily; however, there are also cases where such statements may seem correct while actually not being evidence-grounded or confirmed without careful judgment \cite{huang2025survey, sun2025and}. Furthermore, LLMs might have biases in their generated content while responding to a specific prompt, such as in politics, culture, race, and gender-based topics \cite{haller2023opiniongpt}. Finally, LLMs remain ``black-box" models in general; people do not understand how the internal knowledge base and inner working mechanism enable these models to make their predictions. These problems hinder the applicability of such black-box AI models in safety-critical tasks and domains, and raise serious issues regarding their societal alignment and compliance with regulatory principles \cite{gabriel2020artificial}. Consequently, there is an imminent need for a thorough investigation of interpretability and explainability in LLMs. \\
{The terms \textit{interpretability} and \textit{explainability} are used interchangeably or distinguished in explainable artificial intelligence (XAI), depending on the use case and the nature of the tasks under consideration \cite{kim2021multi, saeed2023explainable, garouani2024investigating}. Traditionally, in the realm of XAI, interpretability has been defined as the extent to which a human can understand a model’s inner workings without additional post-processing. It is often associated with inherently simple models (e.g., linear regression, small decision trees) where the relationship between inputs and outputs is directly transparent. Explainability, by contrast, comes into play when a machine learning model is too complex to be inherently interpretable (e.g., deep neural networks) and refers to the use of secondary methods to explain why the model produced a particular output. In the context of LLMs,  those topics can be explored in two directions: Understanding these models 1) by their inner workings, i.e., mechanistic interpretability, and 2) based on the predictions they generate for particular tasks, i.e., local explainability (Figure \ref{fig:llm_exp_local_mech}). {Meanwhile, \textit{trust} is the willingness of end users to adopt an AI system and believe in conclusions, decisions, and suggestions produced by such a system \cite{jacovi2021formalizing, ferrario2022explainability, afroogh2024trust}. Ultimately, evolving XAI research suggests that the \textit{appropriate use} of explainability techniques might foster trust in the AI systems from the social, technical, and legal perspectives \cite{jacovi2021formalizing, afroogh2024trust, miller2019explanation}. Motivated by this synergy, our paper specifically seeks to reveal the implications of local and mechanistic explanations in LLMs from a trust viewpoint for targeted interaction partners. Given that interpretability and explainability in LLMs remain in their early stages of development, we begin by elucidating the foundational concepts within these tools and synthesizing the key insights and findings emerging from recent research within the LLM community. After analyzing the findings from the reviewed studies, we further summarize insights from empirical evidence in two safety-critical applications. Building on our critical analysis and experimental results, we finally propose future directions to advance the trustworthiness of LLM explanations. 

\subsection{Contributions}
Focusing on LLM interpretability and explainability through the lenses of trustworthiness and human-centricity, our paper makes three key contributions:
\begin{itemize} [leftmargin=1.5em]
    \item We review the local explainability and mechanistic interpretability techniques in the LLM literature, present insights from these studies, and reveal critical aspects of trustworthy LLM explanations;
    \item We draw insights from experiments in two safety-critical domains and show trust implications of LLM explanations for explanation receivers;
    \item We present unaddressed issues in the current approaches to LLM explainability and propose guidelines toward achieving eight fundamental goals of a trustworthy LLM with explanations.  
\end{itemize}
\subsection{Structure of the Paper} 
The paper is organized into the following sections. After the introduction, Section 2 presents background information on LLMs. Section 3 reviews related studies on LLM explainability and demonstrates how our paper bridges gaps in the current literature. Section 4 presents a detailed review of local explainability and mechanistic interpretability approaches in  LLMs. Critical aspects in presenting trustworthy LLM explanations and their assessment are covered in Sections 5 and 6, respectively. After that, Section 7 presents empirical insights from two safety-critical domains, and

\begin{figure*}[!ht]
 \centering
    \includegraphics[width=\linewidth]{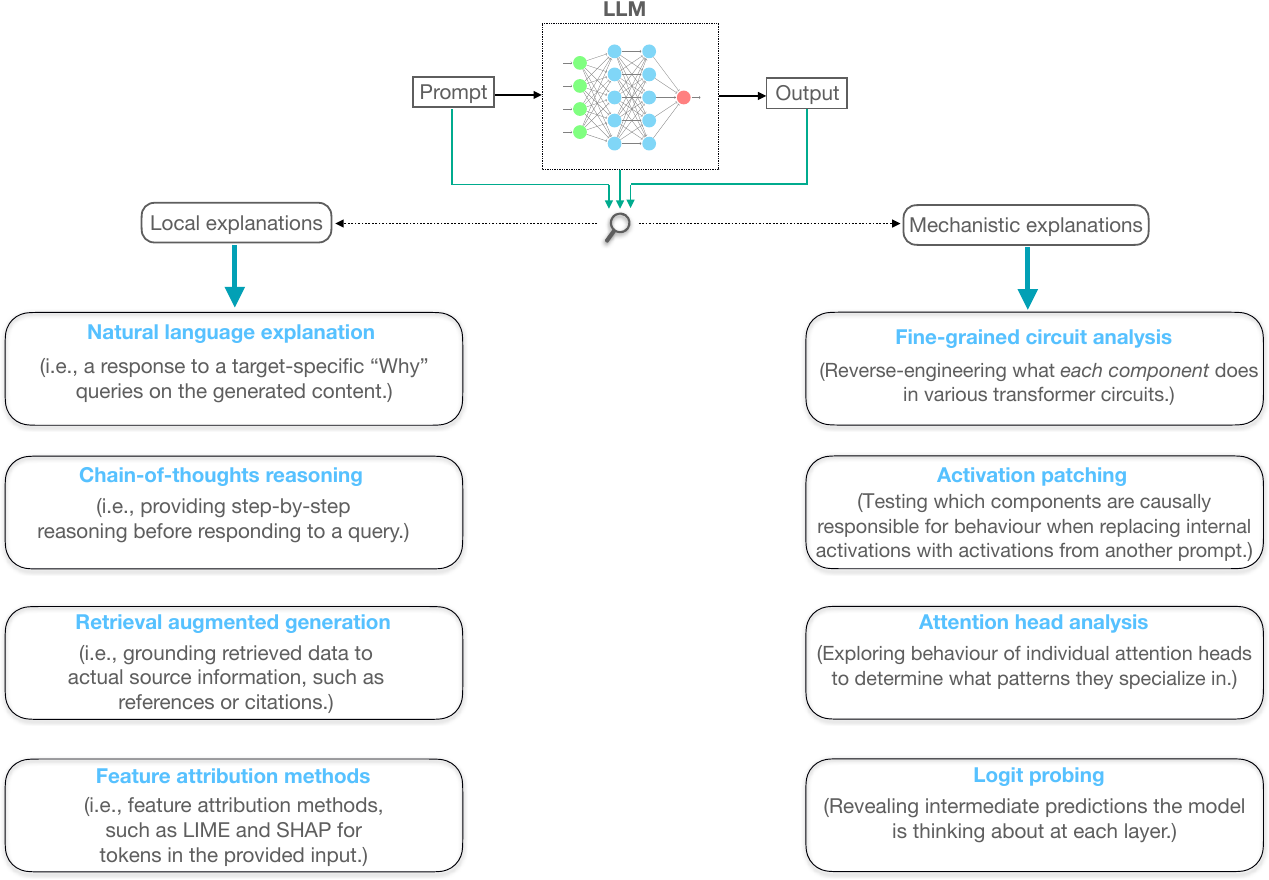}
    \caption{Local explainability and mechanistic interpretability approaches in LLMs.}
    \label{fig:llm_exp_local_mech}
\end{figure*}
key findings from these studies, in conjunction with the existing literature, are presented in Section 8. Finally, Section 9 proposes potential future directions for advancing trustworthiness in LLM explanations, and Section 10 concludes the paper with an overall summary.        
 
\section{Background of Language Models}
\subsection{The Rise of LLMs}
The history of LLMs dates back to the early 1980s and all the way through 2026, with several evolutions based on their training mechanisms and capabilities. Unsurprisingly, all these models are based on the main theme of language modeling, i.e., predicting the probability of a word occurring given its surrounding context. Earlier statistical learning models \cite{brown1992class,ney1994structuring} were developed on a purely Markov assumption, i.e., the probability of a word occurrence is wholly dependent on the n-grams preceding it. However, these methods were not successful in predicting words accurately when the preceding context is large, which creates a huge computational burden of calculating an exponential number of transition probabilities. Subsequently, deep learning models that can learn distributed word representations more efficiently have been shown to exploit the back-propagation technique proposed by \cite{rumelhart1986learning,bengio2000neural}. Later, these methods were improved by posing them as a self-supervised classification task with a contrastive loss function and negative sampling; this improvement led to the development of language models such as Word2Vec \cite{mikolov2013distributed} and FastText \cite{bojanowski2017enriching}. The resulting distributed representations primarily focus on how a word is connected to its neighboring words. Moreover, this method can help achieve high accuracy in several NLP domains when used to pre-train those models through various techniques, such as retrofitting \cite{faruqui2014retrofitting}, domain-specific embedding creation and mapping \cite{farruque2019augmenting}. Subsequently, even more efficiency improvements emerged for task-specific context-aware embedding representation building. These include contextual word embedding methods, where a word's embedding vector becomes dynamic and changes based on its surrounding context, enabling the creation of an embedding representation for an entire sentence. In this research trajectory, the most popular methods were Google's Universal Sentence Embedding \cite{cer2018universal} and Allen AI's ELMO \cite{DBLP:journals/corr/abs-1802-05365}. The basic architecture used here was based on the Bi-LSTM method to capture a word's context in both directions, or a simple MLP, first pre-trained in an unlabeled corpus and then fine-tuned on various mainstream NLP tasks. \\
The next revolution in the language modeling area emerged when Vaswani et al. \cite{vaswani2017attention} published their paper on the Self-Attention mechanism, which not only helped with learning an end-to-end high-performing model for language modeling, but also helped with identifying the salience of participating words for next word prediction or other downstream classification tasks. The first models using this architecture were different variations of Google's BERT \cite{kenton2019bert} (such as ALBERT \cite{Lan2020ALBERT} and RoBERTa \cite{liu2019roberta}), decoder-only or GPT models \cite{wu2023brief} (GPT-1,2, and 3), and models with a BERT-like encoder with a GPT-like Decoder, such as BART \cite{lewis2020bart} from Meta. These models, with several enhancements to the costly Self-Attention weight computation, revolutionized the applied NLP industry across most mainstream NLP tasks from 2018 to 2021, including Machine Translation, Summarization, and Text Classification. Still, the primary advantage of these BERT-based models stemmed from the powerful hardware (GPUs, TPUs) that were used to train these highly parameterized models. This included computing with millions of parameters and then fine-tuning with several downstream NLP tasks. Although these models demonstrated strong word prediction capabilities through bidirectional context and produced high-quality sentence embeddings for tasks such as text classification, they still lagged in generative ability—particularly in sustaining open-domain conversational exchanges with humans. This remained the case until the end of 2022, when OpenAI first released their conversational model, ChatGPT \cite{openai2022chatgpt}, based on GPT-3.5. Although ChatGPT used the same Transformer architecture proposed by Google, it delivered significant improvements in the generative text domain because of its decoder-only training mechanism, which primarily focuses on optimizing the model for chat conversation continuation. So, these models not only used very large pre-training corpora but also underwent text-generation-specific fine-tuning, which made them very efficient. Such Transformer-based families of models typically have billions of parameters (and are therefore called large language models or LLMs) and exhibit enhanced capabilities, or so-called emergent capabilities, through in-context learning \cite{wei2022emergent}. \\
Further improvements have been achieved by fine-tuning these models via reinforcement learning with human feedback (RLHF) \cite{chaudhari2025rlhf}, enabling them to learn from high-quality examples curated by human annotators. Recently, the key players in this family of models include proprietary GPT-4 \cite{achiam2023gpt}, GPT-4 Omni \cite{hurst2024gpt}, GPT-5 \cite{gpt5systemcard2025}, Google's Gemini \cite{team2023gemini}, open-source LLaMA models from Meta \cite{touvron2023llama, touvron2023llama2}, and DeepSeek-R1 \cite{guo2025deepseek}. Due to the powerful in-context learning capabilities of these models and the reduction of hallucination via multi-modal fine-tuning, the LLM community has observed significant progress in the areas of prompt-based or zero-shot LLM models, especially in the areas of Retrieval Augmented Generation \cite{lewis2020retrieval, chen2024benchmarking} and Agentic Task Completion \cite{singh2024enhancing}. Although these are considerable advancements in the area of text generation compared to earlier language models, it remains unclear what enables these recent LLMs to acquire emergent capabilities, making it challenging to debug and understand the inner workings of these models. While techniques, such as the chain-of-thought (CoT), enable conversing with these models and aim to find out how they came up with particular reasoning, such methods are not always accurate, as it remains unclear whether the output of these models is faithful to the reasoning they provide to users \cite{turpin2023language, arcuschin2025chain}. All these challenges and fundamental gaps underscore the need for more in-depth investigations into the true working mechanisms of LLMs.

\subsection{The Transformer architecture}
To understand how language models generate output from prompts, it is necessary to examine the internal elements and computational flow and pathways within these deep neural network models. The architecture used to build models with this generative ability for LLMs is called a \textit{Transformer} \cite{vaswani2017attention}. This architecture processes sequential data, allowing an LLM to capture contextual relationships between all input elements simultaneously, enabling tasks such as natural language understanding and generation. Mathematically, the architecture has the following elements and computational pathways. Let's denote a sequence of $n$ items ($\mathbf{x}_1, \mathbf{x}_2, \cdots \mathbf{x}_n$) by $\mathbf{X} \in \mathbb{R}^{n \times d}$. Here, $d$ is the embedding dimension representing each item. Self-attention aims to capture the interaction amongst all $n$ items by encoding each item by using three learnable weight matrices to transform Queries ($\mathbf{W}^Q \in \mathbb{R}^{d \times d_q}$), Keys ($\mathbf{W}^K \in \mathbb{R}^{d \times d_k}$) and Values ($\mathbf{W}^V \in \mathbb{R}^{d \times d_v}$), in which $d_q=d_k$. Here, the input sequence $\mathbf{X}$ is first projected onto these weight matrices to obtain $\mathbf{Q}=\mathbf{X}\mathbf{W}^Q$, $\mathbf{K}=\mathbf{X}\mathbf{W}^K$ and $\mathbf{V}=\mathbf{X}\mathbf{W}^V$. The final output $\mathbf{Z}\in\mathbb{R}^{n \times d_v}$ of
the self-attention layer is then calculated as follows:
\begin{equation}
\mathbf{Z}= \mathbf{softmax}\left (\frac{\mathbf{Q}\mathbf{K}^T}{\sqrt{d_q}}\right )\mathbf{V}
\end{equation}

\begin{figure*}[h!]
 \centering
    \includegraphics[width=.5\linewidth]{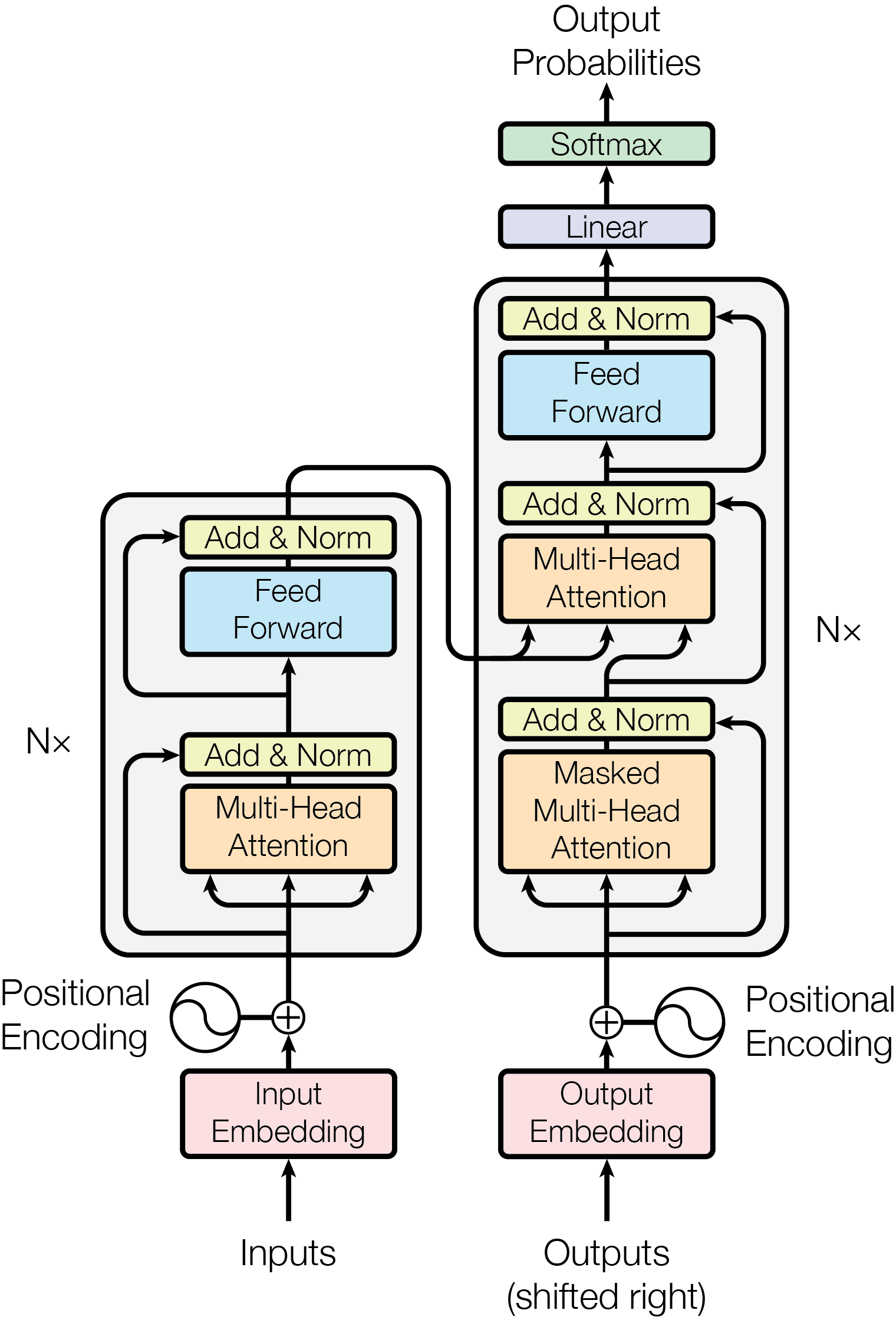}
    \caption{The architecture of Transformer. Source: \cite{vaswani2017attention}. }
    \label{fig:Transformer_model}
\end{figure*} 
\hspace{-0.10cm}
For a given item, $\mathbf{x}_n$, in the sequence, the role of self-attention is to calculate the dot product of the query with all keys, where the softmax operator is further used to get the attention scores. Each item, $\mathbf{x}_n$, eventually becomes the weighted sum of all items in the sequence $\mathbf{X}$, in which weights are given by the attention scores. Figure \ref{fig:Transformer_model} shows the network architecture of the original attention-based Transformer.

\section{Related Work}
Several studies have recently explored the concepts of explainability and interpretability in the context of LLMs. Notably, Zhao et al.'s work \cite{zhao2024explainability} is the first detailed study on explainability for LLMs. The authors present techniques for local and global explanations coupled with model training. They categorize various methods based on conventional fine-tuning-based (i.e., feature attribution explanation, probing-based explanation, mechanistic interpretability, and concept-based explanation) and prompting-based paradigms (i.e., in-context learning, chain-of-thought prompting, and representation engineering).\\
Singh et al. \cite{singh2024rethinking} analyze the interpretation of LLMs in two ways: 1) interpreting an entire LLM model itself, including its output generation, and 2)  using an LLM for producing individual explanations for each output. They propose, for a specific output of an LLM, that a \textit{local explanation} can be explained via a combination of feature attribution (e.g., using SHAP \cite{lundberg2017unified}), natural language explanation, prediction decomposition (i.e., chain-of-thought-reasoning), and data grounding (i.e., retrieval augmented generation (RAG)). For the \textit{global/mechanistic explanation} of an entire LLM model, they propose the use of attribution for LLM internals along with attention head summaries, coupled with some form of algorithmic understanding and data influence. \\
Luo and Specia \cite{luo2024understanding} have further expanded the concepts proposed by Singh et al. \cite{singh2024rethinking}'s work by revealing the role of explanations in directly improving the LLM itself. In \cite{luo2024understanding}, the authors summarize local and global methods for LLM interpretability and also show that an LLM's output can provide an opportunity to further enhance the LLM via model editing and other methods for enhancing a model's capability. They posit that LLM outputs can be used to debug existing models and provide hints for refining and improving the model design.\\
Furthermore, Liu et al. \cite{liu2024large} have evaluated LLMs from a causality perspective, with a focus on reasoning, fairness, safety, and multi-modality. The authors demonstrate that causal reasoning can enhance the reasoning capacity, fairness, and safety of LLMs, ultimately enabling these tools to generate explanations and elucidate how LLMs arrive at their conclusions. \\
Finally, some recent studies have reviewed general approaches to explaining LLM outcomes \cite{volkov2024local} and the computational pathways behind the inner workings of transformer-based LLMs \cite{ferrando2024primer, rai2024practical, sharkey2025open, gantla2025exploring} as scoping reviews. \\
{While all these investigations provide valuable insights into explainability and interpretability of LLMs with a specific focus, the evolving landscape of research in understanding LLMs' working mechanisms requires a deeper investigation to build \textit{trust} in these tools. In this sense, our paper complements previous studies with three significant new analyses. First, we examine the concepts of \textit{tacit} and \textit{explicit} knowledge, two distinct forms of knowledge that differ in how LLMs store, utilize, and generate information. Furthermore, explaining \textit{fact}, \textit{belief}, and \textit{grey-zone information} with an LLM’s knowledge base is also a crucial topic requiring further analysis. Finally, the trustworthiness of LLM explanations with respect to explanation receivers, i.e., explainees, also needs rigorous exploration to ensure explainees understand and use such explanations in a meaningful and purposeful manner. In this regard, we analyze in depth the role of explainability in achieving the fundamental goals of \textit{TrustLLM} \cite{huang2024trustllm}, an internationally adopted framework that benchmarks trust in LLMs across eight key dimensions. Hence, our paper addresses these three significant details as a complement to the current landscape of explainability in LLMs, and presents opportunities and challenges toward generating trustworthy LLM explanations.

\section{Explainability within LLMs}
This section revisits the expected properties of explainability and distinguishes local explainability and mechanistic interpretability concepts within LLMs.

\subsection{LLMs meet XAI}
XAI has emerged as a potential solution to achieve transparent and trustworthy AI systems, particularly in safety-critical applications \cite{saeed2023explainable, hossain2025explainable, atakishiyev2024explainable}. Historically, preliminary studies on XAI have leveraged explanations as a reasoning approach in expert systems or as symbolic representations for neural networks \cite{ goebel2018explainable, confalonieri2021historical,  garcez2023neurosymbolic, buchanan1984rule}. But the rising dominance of deep learning and extensive use and deployment of deep neural network-based models have added further complexity,  and increased the need for AI explanations. According to \cite{saeed2023explainable}, explanations in modern AI systems should consider five essential perspectives: regulatory, scientific, industrial, developmental, and social perspectives. The regulatory compliance perspective primarily ensures the ``right to explanation" from a legal perspective as first articulated by the General Data Protection Regulation (GDPR) of the European Union \cite{regulation2016regulation}. The key point of this recital is that an AI system must justify its specific conclusions or decisions, particularly in critical domains and applications. Furthermore, XAI can help extract explicit, useful information from black-box models trained on large datasets, where the extracted knowledge can lead to further scientific discovery within the field. In addition, as many industrial entities make customer satisfaction as a top priority in automated systems, AI explanations can have value in improving user trust by providing transparency in autonomous decision-making. Moreover, from a developmental viewpoint, explanations can help debug black-box AI systems, discover potential faults and flaws in existing systems, and incrementally improve both reliability and trust. Finally, from a societal perspective, one can better understand what an AI system aims for rather than focusing on what it has been trained for. These non-exhaustive points of view reveal that the value of explanations can be measured in various ways, depending on the needs and expectations of a targeted audience. \\
Having recently been applied across many domains and critical tasks \cite{chang2024survey, raiaan2024review}, all the above-mentioned explainability perspectives inherently align with LLMs. While emergent abilities of LLMs help enable them to reason, summarize, generalize, and make predictions across a variety of downstream tasks, model errors --- primarily known as hallucinations --- are serious issues with these models. In particular, in safety-critical applications and domains, hallucinations may drastically impact the outcome of tasks, resulting in serious trust, transparency, and accountability issues among the stakeholders involved \cite{huang2024trustllm, wang2024navigating}. {Such issues necessitate rigorous regulatory compliance requirements for the design, technical strength, and human factors considerations of LLMs as pervasive AI tools \cite{dubey2024nested, gyevnar2025ai}. In this sense, persona-adaptable strategies are critical in LLM regulation because LLMs can dynamically shift their outputs based on perceived user persona, such as age, cultural background, or emotional state, creating context-dependent risks including targeted manipulation, biased recommendations, or personalized disinformation \cite{hattab2024persona, ghandeharioun2024s, li2026persona}. Ultimately, XAI as a design implication in LLMs could transform transparency from a regulatory checkbox into a functional necessity that guides data curation, training objectives, and deployment interfaces for targeted interaction partners \cite{hattab2024persona}. Consequently, explainability and trustworthiness in LLMs are deeply intertwined, and explainability, at least to a feasible extent, is a significant topic deserving careful exploration for achieving trust in LLMs. We discuss the explainability-trust synergy in more detail in Subsection 8.3.\\

\subsection{Local explainability of LLMs} 
Analogous to other learned models, LLM explanations for their predictive outputs can be classified as \textit{local explanations} and \textit{global explanations}. Local explanations refer to explanations provided in response to individual explicit outputs generated by an LLM \cite{zhao2024explainability,volkov2024local}. More formally, we can define a formulation of LLM explainability as follows. Let us use the following notation:
\begin{itemize}
    \item \( f: \mathcal{X} \to \mathcal{Y} \) be a pretrained LLM, where \( \mathcal{X} \) is the input space (e.g., token sequences) and \( \mathcal{Y} \) is the output space (i.e., the token with the highest probability).
    \item \( x \in \mathcal{X} \) be a specific input, i.e., a prompt.
    \item \( f(x) \) is the model's output on input \( x \).
    \item \( h^{(l)} \in \mathbb{R}^{d_l} \) is the hidden representation at layer \( l \in \{1, \dots, L\} \).
    \item \( z^{(l)} \in \mathbb{R}^{d_l} \) is the pre-activation value (e.g., logits) at layer \( l \).
    \item \( T \) is the number of tokens in the prompt.
\end{itemize}
Based on this notation, a method \( \mathcal{E}_{\text{local}} \) provides a local explanation for an input–output pair \( (x, f(x)) \) by constructing a simpler interpretable function \( g \in \mathcal{G} \) such that:

\[
g \approx f \quad \text{in a neighborhood } \mathcal{N}_x \subset \mathcal{X} \text{ around } x
\]

and

\[
\mathcal{E}_{\text{local}}(x; f) := g
\]

subject to the general objective, such as analogous to the one in LIME \cite{ribeiro2016should}:

\[
\underset{g \in \mathcal{G}}{\arg\min} \; \mathcal{L}_{\text{fidelity}}(f, g; \mathcal{N}_x) + \lambda \cdot \Omega(g)
\]

in which:
\begin{itemize}
    \item \( \mathcal{L}_{\text{fidelity}} \) measures the closeness of \( g \) to \( f \) in \( \mathcal{N}_x \).
    \item \( \Omega(g) \) penalizes the complexity of the explanation function.
    \item \( \lambda \geq 0 \) is a regularization parameter.
\end{itemize}
Essentially, the local explainability method must provide an explanation for why the model generated a specific output in response to a user-entered prompt. At the highest level, local explanations can be distinguished as natural language explanations, chain-of-thought reasoning, retrieval-augmented generation, feature attribution methods, or some combination of those methods \cite{zhao2024explainability}. We describe each of these approaches as follows:  \\
(1) \textit{Natural language explanations:} Given a user-provided prompt, the model generates an explanatory description for the prompt in natural language \cite{singh2024rethinking}.  In this case, the LLM outputs both the predicted answer and an explanation, often structured with requests like ``Explain your answer'' in the prompt.

     \begin{table}[t!]
\caption{Selected representative studies on local explanation techniques within LLMs}
\captionsetup[table]{position=above}
\centering
\resizebox{\textwidth}{!}{%
\begin{tabular}{cC{6cm}C{8cm}C{6cm}}
\specialrule{.2em}{.1em}{.1em}
\textbf{Study} & \textbf{Goal of the study} & \textbf{LLM used} & \textbf{Key findings/takeaways}\\
\specialrule{.2em}{.1em}{.1em}


\multicolumn{4}{c}{\textit{Natural language explanation}}\\
\hline
Huang et al., \cite{huang2023can}, 2023 &  LLM's ability to explain its predictions & ChatGPT & ChatGPT's self-explanations differ from the compared metrics while being computationally less costly to produce.\\
\hline
Tanneru et al., \cite{tanneru2024quantifying}, 2024 & Introducing metrics to measure the uncertainty in LLM-generated explanations & InstructGPT, GPT-3.5, and GPT-4 & Probing uncertainty correlates with explanation faithfulness, while verbalized confidence may be misleading.\\
\hline
Chen et al., \cite{chen2025towards}, 2025 & Enhancing consistency of natural language explanations with explanation-consistency finetuning &  LLaMA-2 13B & Explanation-consistency finetuning enhances consistency of explanations in finetuning datasets and generalizes to out-of-distribution datasets as well.\\
\hline
Chuang et al., \cite{chuang2026faithlm}, 2026 & Improving the faithfulness of natural language explanations without task-specific heuristics or token masking in LLMs & Vicuna-7B, Phi-2, GPT-3.5-Turbo, Claude-2  & Intertwining intervention-based evaluation with iterative optimization provides a principled route toward trustworthy LLM explanations.\\
\hline


\multicolumn{4}{c}{\textit{Chain-of-Thought reasoning}}\\
\hline
Wei et al., \cite{wei2022chain}, 2022 & Understanding the role of intermediate reasoning on the ability of LLMs to perform complex reasoning & GPT-3, LaMDA, PaLM, UL2 20B, Codex & Prompting LLMs with intermediate reasoning steps improves performance on complex tasks.\\
\hline
Lyu et al., \cite{lyu2023faithful}, 2023 & How to generate faithful reasoning for LLM predictions & Codex & Using symbolic reasoning chain and deterministic solver enhances faithfulness of chain-of-thought reasoning.\\
\hline
Turpin et al., \cite{turpin2023language}, 2023 & Critical examination of the faithfulness of LLM explanations when using CoT prompting & GPT-3.5 and Claude 1.0 & CoT explanations can be misleading and may not always reflect the true reasoning behind predictions.\\
\hline
Arcuschin et al., \cite{arcuschin2025chain}, 2025 & Investigating unfaithful CoTs on realistic prompts & GPT, Gemini, DeepSeek, LLaMA series (15 models) & Even frontier models like GPT-4o and Gemini 2.5 Pro may produce CoTs that sound logical but don’t match the actual decision process.\\
\hline
Chen et al., \cite{chen2026does}, 2026 & Understanding CoT reasoning with sparse autoencoding & Pythia-70M, Pythia-2.8B & CoT might induce interpretable internal structures in frontier LLMs.\\
\hline

\multicolumn{4}{c}{\textit{Retrieval-augmented generation}}\\
\hline
Yuan et al., \cite{yuan2024rag}, 2024 & Providing human-understandable justifications for driving decisions & Vicuna1.5-7B & RAG-Driver improves interpretability of autonomous driving systems by combining a multi-modal LLM with retrieval-augmented in-context learning.\\
\hline
Li et al., \cite{li2025g}, 2025 & Framework for personalized and explainable recommendations & LLaMA family & LLMs augmented with structured graph data move beyond generic recommendations to transparent, user-specific ones.\\
\hline


\multicolumn{4}{c}{\textit{Feature-attribution methods}}\\
\hline
Wu et al., \cite{wu2020perturbed}, 2020 & Analyzing linguistic knowledge embedded in BERT via perturbed masking & BERT & Perturbed Masking provides a faithful, parameter-free lens to reveal internal mechanisms of BERT.\\
\hline
Mohebbi et al., \cite{mohebbi-etal-2021-exploring}, 2021 & How individual token representations within BERT contribute to probing tasks & BERT & BERT encodes meaningful knowledge in specific token representations.\\
\hline
Enguehard, \cite{enguehard2023sequential}, 2023 & Refining Integrated Gradients (IG) for explainability & BERT, DistilBERT, RoBERTa & Sequential Integrated Gradients (SIG) provides a more accurate and intuitive explanation method.\\
\hline
Kokalj et al., \cite{kokalj2021bert}, 2021 & Adapting SHAP to transformer-based LLMs & BERT & TransSHAP bridges SHAP and transformers with a structure-aware explanation technique.\\
\hline
Wu and Ong, \cite{wu2021explaining}, 2021 & How well attribution methods explain BERT’s decision-making in classification & BERT & Attribution methods generalize well across tasks with shared semantics.\\
\hline

\end{tabular}%
}
\label{tab:llm_local_explanations}
\end{table}
{Meanwhile, there are significant challenges with faithfulness of natural language explanations in LLMs. Such explanations are generally a distinct artifact and a post-hoc narrative generated by the LLM to convince the explainee instead of being a verifiable log of the model's internal computation \cite{parcalabescu2024measuring, alon2026faithfulness}. The core reason is that LLMs are generally trained to predict the next token in a sequence, where this procedure involves learning to yield plausible output. However, the model does not have direct access to its own internal reasoning; it only has access to its own hidden states and the input text. Such a built-in mechanism may cause LLM explanations to appear as  \textit{correlation vs. causation} \cite{matton2025walk} and \textit{confabulation}, which refers to hallucination of reasoning \cite{sui2024confabulation}. Consequently, explainees always need to carefully inspect where the explanation indeed supports an LLM's prediction with logical reasoning and accurate references.\\
(2) \textit{Chain-of-thought reasoning (CoT):} This approach factors complex problems into a series of sequential steps or intermediate thoughts that sequentially lead to an answer for the provided prompt \cite{wei2022chain}. Rather than directly predicting an answer, the explanation process walks through intermediate reasoning steps that lead to the final decision, aiming to make its internal logic explicit for a specific input. This makes CoT a form of local explanation, as it focuses on clarifying a single predictive output rather than describe a model’s behavior across all inputs. CoT is especially effective in producing explanations for complex reasoning tasks such as math word problems, commonsense reasoning, and multi-hop question answering. For instance, given a math question, the model might output: “We start with 10 apples, eat 3, then collect 5 more. So, in the end we have (10 - 3 + 5) = 12 apples,” followed by the final answer. CoT explanations can be prompted with cues like ``Think step by step,”  or generated through few-shot prompting, or supervised training on datasets with annotated reasoning paths.  Compared to other explanation methods, such as feature attributions, CoT can offer more understandable justifications for a model’s behavior on individual examples. However, similar to natural language explanations, some recent investigations show that CoT may not always present faithful explanations for specific prompts despite seeming convincing in support of the output prediction \cite{arcuschin2025chain, barez-chain-2025}. This nuance is again referenced to the correlation vs. causation \cite{matton2025walk} and confabulation \cite{sui2024confabulation} issues mentioned above. For instance, for our highlighted apple example above, it is not truly clear whether the model indeed performed the mathematical subtraction or addition operations for the instruction of ``Think step by step,'' or if it performed pattern matching with the final answer 12, based on the learning from its own training data. Hence, faithfulness of CoT also remains an open research topic in the LLM and XAI communities, requiring further exploration. \\
(3) \textit{Retrieval-augmented generation (RAG):} This is a technique that enhances the capabilities of an LLM by integrating relevant material from external knowledge bases (e.g., references, citations) into the generation process \cite{lewis2020retrieval, gao2023retrieval}. The retrieval component of RAG focuses on retrieving relevant documents or data from an external knowledge base, as indicated in the user's query/prompt. Augmentation refers to the incorporation of context and additional details. These key points ultimately enable the generation of a response that incorporates both the original query and the augmented information. The RAG process differs from purely generative methods, which rely solely on internal parameters; it grounds the explanation in potentially verifiable facts, which intuitively enhances trust and reduces hallucination.\\
(4) \textit{Feature attribution-based explanations:} These explanations aim to quantify the
contribution of each input feature (e.g., words, tokens) to a model’s generated prediction. This category includes perturbation-based methods, gradient-based methods, surrogate models, and decomposition-based methods \cite{zhao2024explainability}. 
\begin{itemize}
    \item Perturbation-based methods modify input features and evaluate the output of the model against such modifications.
    \item Gradient-based methods reveal feature importance by leveraging the partial derivatives of the model's output with respect to input features (but requires some precision in defining such partial derivatives).
    \item Surrogate feature attribution-based explanation methods, on the other hand, use simpler, interpretable models to approximate the behavior of complex LLMs and provide insights into how different features contribute to their predictions (i.e., LIME \cite{ribeiro2016should}, SHAP \cite{lundberg2017unified}).
    \item Finally, decomposition-based methods explain the outputs of LLMs by breaking their reasoning into smaller, interpretable components, such as layer-based or individual neuron-based analysis.
\end{itemize}
Furthermore, attention-based explanations (e.g., heatmaps \& visualization) and example-based explanations (counterfactual analysis \cite{cheng2024interactive} and human-adversarial examples \cite{paulus2024advprompter}) can be considered as an analysis of local LLM predictions. Table \ref{tab:llm_local_explanations} shows representative examples of local explanation generation methods in LLMs from the relevant studies. \\
Overall, from the broad spectrum, local explanation generation approaches follow three main paradigms by design \cite{zhao2024explainability}: the explain-then-predict approach (e.g., CoT reasoning), the predict-then-explain (post-hoc explanation), and the joint predict-explain mechanisms (e.g., jointly training explanation and prediction heads). While the same LLM-generated explanations are favorable for the model's predictions, there are a few approaches incorporating another model to explain the behavior of a target LLM, coined ``LLM-as-a-Judge", as such ``judge" LLMs might have a better ability to provide cost-effective and consistent evaluation \cite{gu2024survey}.

\subsection{Mechanistic interpretability of LLMs}

Global explanations, also sometimes referred to as "mechanistic explanations," in the context of LLMs describe the model's internal working mechanism by explaining the predictive behavior of the entire model \cite{singh2024rethinking, bricken2023monosemanticity, templeton2024scaling}. In general, generating mechanistic explanations of LLMs is more challenging than generating local explanations.  A significant part of the problem is that accurately conveying internal functions requires an embedded self-model, but existing LLMs do not possess such a representation.  While local explanations focus on justifying a specific output, mechanistic interpretability deals with more complex tasks and aims to provide a holistic explanation of how LLM works in its entirety. Within the notation provided above,  a mechanistic explanation \( \mathcal{E}_{\text{mech}} \) provides a mapping:
\begin{figure*}[h!]
 \centering
    \includegraphics[width=.95\linewidth]{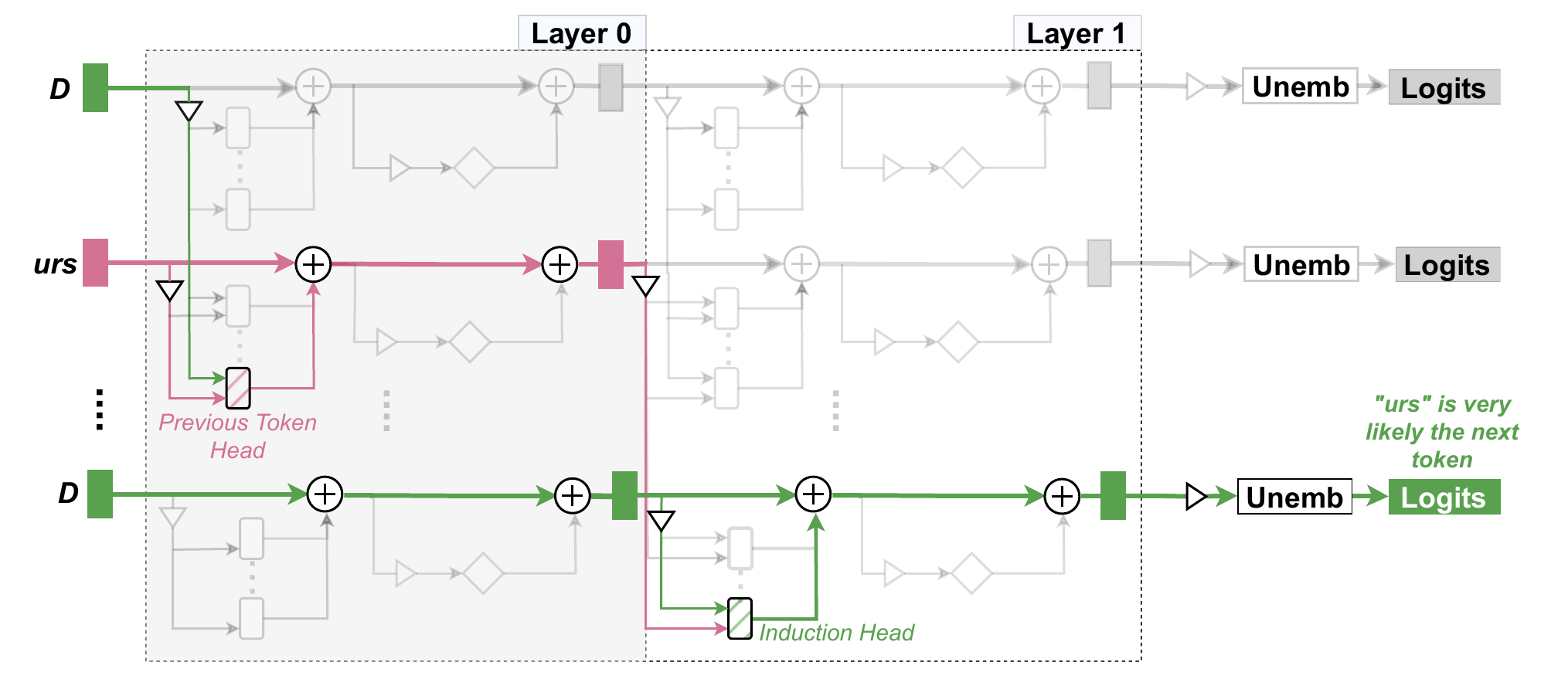}
    \caption{An example of an induction circuit provided by Rai et al. \cite{rai2024practical}, based on Elhage et al.'s \cite{elhage2021mathematical} findings, showing outputs of two attention heads as nodes, and a link between them as edges, enabling the model to generalize from repeated patterns. }
    \label{fig:induction_circuit}
\end{figure*}
\[
\mathcal{E}_{\text{mech}}: \{W^{(l)}, b^{(l)}, h^{(l)}\} \to \mathcal{S}
\]
Here \( \mathcal{S} \) is a structured representation of functional roles of components, such as attention heads, neurons, and circuits. It is noteworthy to mention that these most common concepts are specific to the Transformer architecture used to create most of the existing LLM internal representation models, and so do not easily convey any human-interpretable explanations. From the latter perspective, whether Transformer architectures are the only appropriate foundation model is currently a question of significant debate, and further research reveals the need for a broader spectrum of foundation architectures (e.g., \cite{garcez2023neurosymbolic, wang2024towards, marra2024statistical}.
Formally, mechanistic interpretability in Transformer seeks to identify functions \( \phi_k \) such that:

\[
f(x) \approx F(\phi_1(x), \phi_2(x), \dots, \phi_K(x))
\]

In which
\begin{itemize}
    \item \( \phi_k(x) \) represents a functional sub-component of the model.
    \item \( F \) is a known or inferred composition function approximating \( f \).
\end{itemize}

These functions are often discovered via probing or causal intervention methods (e.g., activation patching), such as:

\[
\phi_k(x) := \psi(h^{(l)}_i(x)) \quad \text{for some } \psi: \mathbb{R} \to \{0, 1\}
\] A broader scope of mechanistic interpretability of LLMs generally falls into the umbrella of two key focus areas: \textit{circuit} and \textit{feature} analysis \cite{rai2024practical}. Features, which serve as the fundamental units of representation, are interpretable input-related attributes encoded in the language model’s activations \cite{gantla2025exploring}. They are ``interpretable'' because they have a natural interpretation in their application domain. For instance, when a user enters a prompt containing specific tokens, the language model can predict responses related to these features, which it has already learned during the pretraining stage. Therefore, decoding such features can be considered as one way of mechanistic interpretability in LLMs. \\ 
While feature-based analysis is one approach to LLM interpretability, understanding the entire LLM's behavior goes beyond the analysis of features and the interconnection of an LLM's subgraphs. These interconnected components are responsible for guiding specific tasks,  which are important to understand
the entire model’s operating principles. Such subgraphs within an LLM are referred to as \textit{circuits} and several types of circuits have recently been proposed and discovered through LLM interpretability research, such as induction circuits (generalizing from repeated patterns), copy circuits (copying tokens from earlier contexts), indirect object identification (IOI) circuits (resolving pronouns

\renewcommand{\arraystretch}{1.2} 
\scriptsize
\begin{longtable}{>{\centering\arraybackslash}p{2.5cm} 
                >{\centering\arraybackslash}p{4cm} 
                >{\centering\arraybackslash}p{2.5cm} 
                >{\centering\arraybackslash}p{5cm}}
    \caption{Empirical studies on mechanistic interpretability of LLMs, focusing on attention head analysis, fine-grained circuit analysis, logit probing, activation patching, as well as a combination of these methods} \label{tab:mech_int} \\
    \toprule
    \textbf{Study} & \textbf{Goal of the study} & \textbf{LLM used} & \textbf{Key findings/takeaways} \\
    \midrule
    \endfirsthead

    \caption[]{(Continued)} \\
    \toprule
    \textbf{Study} & \textbf{Goal of the study} & \textbf{LLM used} & \textbf{Key findings/takeaways} \\
   
    \endhead

    \midrule
    \multicolumn{4}{r}{\textit{Continued on next page}} \\
    \midrule
    \endfoot

    \bottomrule
    \endlastfoot

			Elhage et al., \cite{elhage2021mathematical}, 2021 &  Reverse engineering transformers with two or less layers and solely have attention blocks  &  GPT-3 & Induction heads can help explain in-context learning in small models \\
			
            \hline
            Olsson et al., \cite{olsson2022context}, 2022 &  Reverse engineering transformers with larger (\textgreater 2) attention blocks  &  GPT-2, GPT-3 &``Induction heads might constitute the mechanism for the actual majority of all in-context learning in large transformer models."\\
		\hline
            Wang et al., \cite{wang2023interpretability}, 2023 &  Understanding LLMs via their circuits   &  GPT-2 Small & Identification of attention heads compensating for the loss of function of other heads, and heads contributing negatively to the next token prediction.\\
            \hline
            Bills et al., \cite{bills2023language}, 2023 &  Understanding what patterns in text cause a neuron to activate &  GPT-2, GPT-4 & LLMs can be used to interpret and explain the behavior of individual neurons within LLMs.\\
            \hline
            Goldowsky-Dill et al., \cite{goldowsky2023localizing}, 2023 &  Generalizing path patching to test hypotheses containing any number of paths from
input to output  &  GPT-2 & ``Path patching is an expressive formalism for localization claims that is both
principled and sufficiently efficient to run on real models"\\
            \hline
            Gurnee et al., \cite{gurnee2023finding}, 2023 &  Understanding how high-level human-interpretable features are represented within the internal neuron activations of LLMs  &  EleutherAI’s Pythia suite  & Sparse probing is an effective methodology to locate neurons contributing to interpretable structures.\\
            \hline
            Wu et al., \cite{wu2024interpretability}, 2023 &  Efficient search for interpretable causal structure in large language models &  Alpaca-7B & Boundless distributed alignment search solves a simple numerical reasoning problem in a human-interpretable manner.\\
            \hline
            Hase et al., \cite{hase2024does}, 2023 &  Understanding where to manipulate knowledge in language models &  GPT-J & ``Model edit success is essentially unrelated to where factual information is stored in models, as measured by Causal Tracing".\\
            \hline
            Li et al., \cite{li2024inference}, 2023 & Improving the truthfulness of language model outputs  &  LLaMA, Alpaca, Vicuna & Subset of attention heads plays a major role in the truthfulness of model outputs.\\
            \hline
            Hou et al., \cite{hou2023towards}, 2023 &  Exploring mechanistic interpretability of language models for multi-step reasoning tasks  &  GPT-2, LLaMA & `` We can often detect the
information in the reasoning tree from the LM’s
attention patterns."\\
            \hline
            Conmy et al., \cite{conmy2023towards}, 2023 & Finding the connections between the abstract neural network units that form a circuit.  &  GPT-2 Small & Automatic circuit discovery algorithm conducts all the activation patching experiments essential to discover which circuit is responsible for the model's behavior".\\\hline
            Friedman et al., \cite{friedman2024learning}, 2023 & Modifying Transformer that can be trained via gradient-based optimization and converted
into a human-readable program &  GPT-family &Transformer Programs can help trace information flows between model components and different positions.\\
            \hline
            Hanna et al., \cite{hanna2024does}, 2023 & Identifying circuits in an LLM model that performs mathematical tasks  &  GPT-2 Small & GPT-2 small’s final multi-layer perceptrons boost model's mathematical reasoning ability.\\
            \hline
            Zhang and Nanda \cite{zhang2023towards}, 2024 &  Understanding impact of methodological details in activation patching  &  GPT-2 small, GPT-2 XL  & Variations in metrics and methods in activation patching in language models might lead to different interpretability results.\\\hline
            Gould et al., \cite{gould2023successor}, 2024 &  Discovering successor heads, a class of attention heads that lead detecting interpretable model features &  GPT-2, Pythia, and LLaMA-2 & ``Successor heads exhibit a weak form of universality, arising in models across different architectures and scale."\\
            \hline
            Lee et al., \cite{pmlr-v235-lee24a}, 2024 &  Understanding how toxicity is represented and elicited in a pre-trained language
model  &  GPT2-medium, LLaMA-2-7B & Subtracting specific vectors in MLP blocks from the residual stream can eliminate toxic outputs.\\
            \hline
            Jain et al., \cite{jain2024mechanistically}, 2024 &  Understanding effects of fine-tuning on procedurally defined tasks  &  GPT-3.5 & Finetuning minimally
changes a pretrained language model's capabilities in procedurally defined tasks.\\
            \hline
            Ren et al., \cite{ren-etal-2024-identifying}, 2024 &  Understanding in-context learning of LLMs  &  InternLM2-1.8B & ``Specific attention heads encode
syntactic dependencies and semantic relationships in natural languages"\\
            \hline
            Prakash et al., \cite{prakash2024finetuning}, 2024 &  Understanding impact of fine-tuning
on the internal mechanisms implemented in language models.  &  LLaMA-7B, Vicuna 7-B, Goat-7B, FLoat-7B & Functionality of the base model's circuit remains unchanged in fine-tuned models\\
            \hline
            Lan et al., \cite{lan2024towards}, 2024 & Comparing circuits for  sequence
continuation tasks on Arabic numerals, number words, and months  &  GPT-2 Small, LLaMA-2-7B & ``Semantically related sequences
rely on shared circuit subgraphs with analogous roles".\\
            \hline
            Todd et al., \cite{todd2024function}, 2024 &  Understanding the role of function references in transformer-based language models  &  GPT-J, GPT-Neox, LLaMA-7B, 13B, 70B & Function vectors can be
explicitly extracted from a small number of attention heads and represent the demonstrated task within a hidden state.\\
            \hline
            Niu et al., \cite{niu2024what}, 2024 &  Analying the ability of language models to recall facts from a training
corpus via knowledge neuron &  BERT, GPT-2, LLaMA-2 & ``Knowledge neuron thesis does not adequately explain the process of factual expression."\\
            \hline
            Singh et al., \cite{pmlr-v235-singh24c}, 2024 &  Understanding emergence dynamics of induction heads in a controlled setting  &  GPT-2, GPT-3  & ``Induction heads operate additively,
with multiple heads used to learn the ICL task more quickly."\\
            \hline
            Ghandeharioun et al., \cite{ghandeharioun2024patchscopes}, 2024 &  Using the language model itself to explain its internal representations in natural language  &  LLaMA-2 13B, Vicuna-13B, GPT-J 6B, Pythia-12B & Various kinds of information from LLM representations can be queried via Patchscopes in natural language.\\
            \hline
            Zhou et al., \cite{zhou2024alignment}, 2024 &  Explaining LLM safety through intermediate hidden states  &  LLaMA-2, LLaMA-3, Mistral, Vicuna, Falcon  & LLMs tend to learn ethical concepts during the pretraining phase rather than alignment and can detect malicious or normal inputs in the early layers.\\
            \hline
            Ferrando and Voita \cite{ferrando2024information}, 2024 &  Interpreting LLM predictions by extracting the dominant part of the overall information flow  &  LLaMA-2 & ``Some model components can be specialized on specific domains such as coding or multilingual texts."\\
            \hline  
            He et al., \cite{he2025learning}, 2024 &  The emergence of in-context learning and skill composition.
in a collection of modular arithmetic tasks  &  GPT-style architectures & ``The smallest model capable of out-of-distribution generalization requires two transformer blocks, while for deeper models, the out-of-distribution generalization phase is transient, necessitating early stopping." \\
\hline

         Lindsey et al., \cite{lindsey2025biology}, 2025 &  Applying attribution graphs to study reasoning capabilities of Claude 3.5 Haiku  & Claude 3.5 Haiku & Attribution graphs could help reveal misrepresentative and undesirable CoT reasoning.\\
         \hline

         Chen et al., \cite{chen2025persona}, 2025 &   Identifying and controlling character traits in LLMs via persona vectors  & Qwen2.5-7B-Instruct,  LLaMA-3.1-8B-Instruct & Persona vectors can help detect personality shifts in LLMs.\\
\hline

         Lindsey \cite{lindsey2025introspection}, 2025 &  Investigating emergent introspective awareness in LLMs  & Claude Opus 4 and 4.1 & LLMs have functional awareness of their internal states to some extent, but also are sensitive to post-training techniques.\\
         \hline

         Ahmad et al., \cite{ahmad2026beyond}, 2025 &  Understanding
interpretability of transformer circuits via singular vectors  & GPT-2 Small & Individual singular directions in parameter space might discover interpretable sub-computations.\\
\hline

         Im et al., \cite{im2026how}, 2026 &  Understanding how transformers learn to associate tokens  & Pythia-1.4B & Each set of weights of Transformer has closed-form expressions
in terms of bigram, token-interchangeability, and context mappings.\\
\hline

         Golimblevskaia et al., \cite{golimblevskaia2026circuit}, 2026 &  Circuit discovery for understanding model computations  & GPT-2 Small, Gemma-2-2B, LLaMA-3.2-1B & Transcoders enable efficient incorporation of structural information for robust and scalable interpretability in LLMs.\\
         \hline

         Chen et al., \cite{chen2026decomposing}, 2026 &  Decomposing LLMs into explicit input-to-output computational paths for mechanistic interpretability & GPT-2, GPT-Neo-2.7B, LLaMA-2-7B, OLMo-7B & Jet expansions establish interpretability approaches in the approximation theory and enable modular inspection into the LLM.\\
                  \hline

         Fraser-Taliente et al., \cite{fraser-taliente2026nla}, 2026 &  Generating unsupervised explanations of LLM activations  & Claude Opus 4.6, Claude Haiku 4.5, Claude Haiku 3.5 NLA & Natural language encoder explanations show the signs of possible interpretations of model internals in LLMs and tend to become more informative over training.
\end{longtable}
\normalsize
via syntactic structure), and logit attribution circuits (assigning influence on final token prediction to specific tokens in context). \\
Interpretability techniques within these focus areas include an analysis of an LLM's internal elements and the way these elements form a network, such as attention head analysis, fine-grained circuit analysis, and the impact of various forms of training data distribution \cite{singh2024rethinking}. Meanwhile, a challenge arises from the interactions between neuron units, where each neuron can activate in response to a range of different concepts. For example, a neuron might activate for specific tokens in a language (e.g., Korean or academic citations), making it difficult to generalize or assign a single meaning to that neuron. This phenomenon has been called ``polysemanticity" \cite{scherlis2022polysemanticity}. As a result, a neuron can capture a mixture of information, making it difficult for humans to interpret it in a meaningful and generalized way \cite{templeton2024scaling}. This behavior is often due to information mixing or superposition, where neurons do not represent a single, clear feature. Instead, features are represented by unique combinations of activations in multiple neurons.\\
Another relevant idea for understanding the behavior of LLMs is the use of a dictionary learning approach with sparse autoencoders. The authors in \cite{bricken2023monosemanticity} focus on a single layer from a transformer model with a 512-neuron MLP layer, decomposing its activations into interpretable representations by training sparse auto-encoders on these activations. The findings indicate that sparse autoencoders can effectively uncover interpretable, relatively universal features from neural network activations, providing insights that are not visible at the neuron level alone. For example, sparse auto-encoders applied across different language models yield similar features, supporting the idea of shared, universal representations. Although considering only 512 neurons, the study shows that these neurons can represent thousands of distinct characteristics \cite{bricken2023monosemanticity}.\\
In another example, Rai et al. \cite{rai2024practical} present an example of an induction circuit in Figure \ref{fig:induction_circuit}, based on Elhage et al.'s \cite{elhage2021mathematical} findings on a toy LLM. This circuit involves two attention heads - one called the previous token head and the other the induction head. These heads act as nodes, while the data flow between them, represented by input and output activations, forms the edges of the circuit. The purpose of the circuit is to recognize and continue repeated sequences in the input text (for example, turning ``Mr D urs ley was thin and bold. Mr D” into “urs”). In this process, the previous token head captures that the string “urs” typically follows the token “D” in earlier parts of the text, and the induction head uses this information to boost the likelihood of predicting “urs” as the next token.\\
Overall, important properties for reverse engineering a deep neural network are decomposability and linearity. Decomposability means that representations can be broken down into individual interpretable features, allowing each feature to be analyzed independently, making it easier for humans to understand. Linearity, on the other hand, means that each feature can be represented as a specific direction in the network's internal space, typically relying on the concept of vector directions \cite{elhage2022toy}.\\ 
A deeper analysis of the related work reveals that understanding the entire LLM’s behavior requires a combination of component- and computation-based techniques within these models. Furthermore, another critical consideration is whether the same or similar LLM features and circuits appear in other LLMs \cite{rai2024practical}. This nuance raises a question about the \textit{universality} of LLM interpretability, meaning to what degree similar LLM features and circuits are observed in different LLMs. Table \ref{tab:mech_int}  shows a scoping review of different studies on the mechanistic interpretability of LLMs. As seen, each study has focused on specific LLMs to discover their inner working principle. Within this context, as an example, a recent investigation by \cite{chandna2025dissecting} shows that GPT-2 and LLaMA-2 encode bias differently, and their internal structures respond differently to interventions.  Another study, which focuses on inference stages and robustness through layer-level interventions \cite{lad2024mechanistic}, has determined that various LLMs, such as GPT and Pythia, exhibit different levels of robustness and modularity, with implications for the general safety and optimization of LLMs. Such analyses and Table \ref{tab:mech_int} give a reason for a more careful investigation of the interpretability concept in LLMs. In fact, the way these neural dialogue models generate responses to prompts might be dependent on both the pretraining data and the construction mechanism of such models \cite{dziri2022origin}. \\
Overall, despite all this significant progress, scaling mechanistic interpretability to frontier LLMs remains difficult due to the huge complexity of circuit discovery, the representational phenomenon of superposition, and the intense computational demands from the causality methods \cite{naseem2026mechanistic}. First, isolating a specific circuit for a behavior is a combinatorial search that makes extensive iteration over components an extremely costly practice \cite{basu2025on}. Furthermore, compressing unrelated concepts into overlapping representations may break the link between a neuron and an understandable feature from the superposition perspective \cite{somvanshi2026bridging}. Finally, a dedicated causality technique must perform thousands of forward/backward passes on LLMs to compute causal effects, where this operation is a computationally expensive process \cite{sharkey2025open, naseem2026mechanistic, somvanshi2026bridging}. Currently, most of the frontier LLMs have different training data and architectural design, and such features necessitate individual, model-specific mechanistic interpretability analysis with the three factors mentioned above, rather than seeking universality. Thus, until further scalable solutions to circuit discovery, superposition, and computational inefficiency emerge, our ability to mechanistically understand frontier LLMs might remain limited.

\section{Critical Epistemic Properties of LLMs for Generating Trustworthy Explanations}
\subsection{LLM explanations for facts, beliefs, and the grey zone}

Facts and beliefs are spread across a dichotomy in human reasoning: a similar dichotomy affects our ability to comprehend them in LLM explanations. A fact is something that can be confirmed as objectively true, either verified or proven through evidence, a documented source, observations, or experience, and is not influenced by personal feelings, subjective opinions, or interpretations \cite{armstrong1973belief}. For instance, the statement ``\textit{Akkra is the capital city of Ghana."} is a \textit{fact} as this piece of information can easily be verified from government or official records. In another example in from logic, ``All men are mortal; Aristotle is a man; Therefore, Aristotle is mortal." The implication is direct, and the evidence could not imply the opposite ``Aristotle is immortal." However, this is clear since all the facts can be confirmed. A belief, on the other hand, is an acceptance that something is true or exists, often without requiring definitive proof, and the presence of facts in a belief is hazy \cite{price2002belief}. For instance, the statement ``\textit{Democracy is the best form of government."} can be considered a belief as it reflects a subjective judgment influenced by cultural, ethical, or personal values. However, some might argue that an alternative government structure, such as technocracy, is a more preferred structure as the decision makers are experts in their field within such a government. So, beliefs often reflect personal perspectives.  The formal literature on belief provides a more technical definition of belief as something which is consistent with a collection of supporting beliefs, even if not generally true (e.g., \cite{alchourron1985logic}).  A final subtlety is that there is a grey zone where contradictions emerge, where there could be conflicting evidence against a presented statement. For example, while one might argue that ``\textit{Eating eggs every day improves human health,}" there are also conflicting scientific studies showing that higher regular intakes of eggs were linked to increased risks of cardiovascular disease and mortality \cite{zhuang2021egg}. These aspects are directly related to local LLM explanations, so that when an LLM presents explanatory information on its predictions, the target user must be aware of whether such pieces of information are facts, beliefs, or potentially on the borderline.

\subsection{Tacit and explicit explanations by LLMs: Show vs tell}
LLMs, as knowledge-based models, produce outputs with varying knowledge granularity depending on the prompt design. The knowledge these models retain could be \textit{explicit}, i.e.,  as easily documentable and communicable, but also \textit{tacit}, meaning such kind of knowledge arises from unspecified inference or is experiential and difficult to articulate. For example, when a novice asks \textit{how to bike}, humans can’t fully explain how to balance or steer to that person, as riding a bike is typically learned via experimenting through trial and error. This is an example of tacit knowledge. On the other hand, when one asks for \textit{how to make a banana cake}, this question can be answered more easily as relevant measurements and steps for cooking a banana cake are written in a cookbook. This is an example of explicit knowledge. Tacit knowledge, popularized by Polanyi, refers specifically to the intuitive understanding behind an expert's judgment or skill and is rooted in personal experience that is difficult to verbalize or formalize \cite{polanyi, polanyi2009tacit}. Such ``intuitive'' knowledge is based on personal insight and experience, which is subjective and cannot be easily expressed in symbols such as language, mathematical formulas, or charts and tables; it is distinguished from explicit knowledge that can be more easily expressed in words or other symbols.\\
Current LLMs utilize a representation model that learns patterns from textual data, using them as a basis for language generation and understanding. These models can encode ``implicit rules" or cultural nuances from large amounts of data, but their processes and internal structures are often black boxes, as tacit knowledge is, by definition, subjective, barely textualized knowledge that makes up a large part of our common knowledge.\\
LLMs could have an evolving role in clarifying tacit explanations by acting as intermediaries between implicit knowledge and explicit understanding in several ways. First, LLMs can be used to transform implicit or context-dependent ideas into explicit language. For instance, if one enters a prompt like \textit{how can I be a successful entrepreneur like  [Person]}, an LLM can infer the underlying principles of successful entrepreneurship and articulate them as actionable advice, translating observed behavior into explicit steps. Furthermore, LLMs could analyze and translate context as cross-cultural and cross-domain mediators by uncovering why certain expressions, actions, or phrases might carry particular meanings in different cultures. For instance, LLMs may reveal the significance of communicative differences, such as silence, in Finnish versus Italian conversations \cite{gabbatore2023silent}. Finally, an LLM can adjust its response to a prompt considering a user's comprehension of the provided response (e.g., explain [the concept] like  I am five versus explain [the concept] to an expert).  Overall, from a broad spectrum, tacit and explicit explanations could be differentiated in the manifestation of the \textit{show vs tell} dilemma: Tacit explanations \textit{show} by explaining through actions or context, while explicit explanations \textit{tell} through words, rules, or data.  Hence, LLMs can be considered \textit{amplifiers of tacit understanding}, making implicit knowledge more intelligible and facilitating users' discussions on a topic, while also augmenting their creativity and intelligence.   
\section{Evaluating LLM Explanations}
Understanding the accuracy, correctness, and faithfulness of LLM explanations is crucial for interaction partners to trust and responsibly use them. Hence, LLM explanations need careful evaluation to meet the human-centric requirements discussed above. In general, there are currently two dominant evaluation methods for LLM explanations: human evaluation and automatic evaluation \cite{zhao2024explainability, chang2024survey}. The following subsections describe these two evaluation approaches.

\begin{table}[t!]

\caption{The state-of-the-art automatic evaluation approaches for LLMs}
\captionsetup[table]{position=above}
\centering
\resizebox{\textwidth}{!}{%
\begin{tabular}{cC{6cm}C{8cm}}
\specialrule{.08em}{.1em}{.1em}
\textbf{Evaluation method} & \textbf{Mathematical formulation} & \textbf{Conceptual description} \\
\specialrule{.08em}{.1em}{.1em}

G-EVAL \cite{liu2023g} & Aggregates weighted token log-probabilities into a normalized, probability-based score and guides scoring via CoT. & LLM as a judge emulates human reasoning and evaluates final outputs on custom criteria using CoT prompting. \\
\hline
LLM-EVAL \cite{chen2023llm} & Maximizes correlation with human ratings by analyzing rationales vs numeric scores. & Uses a single prompt together with a unified multi-criteria evaluation schema, such as relevance, appropriateness, and grammar. \\
\hline
REM \cite{yang2024rem} & Leverages a voting mechanism on preferences from multiple LLM judges to evaluate and compare voting patterns to produce a consensus ranking. & Multi‑LLM voting mitigates individual judge biases and leads to a stable ranking. \\
\hline
PandaLM \cite{wang2024pandalm} & Fine‑tuned judge model, such as LLaMA-7B and Pythia-6.9B, outputs a preference score and a rationale between two candidate responses. & Underscores critical
evaluation aspects, such as instruction following and formality, and detects and corrects grammatical issues and logical fallacies. \\
\hline
HELM \cite{liang2023holistic} & Aggregates a multi‑metric system, such as accuracy, calibration, robustness, fairness, and efficiency into standardized, multi-dimensional scores. & Holistically compares LLMs across various tasks and metrics as a unified framework. \\
\hline
AGIEval \cite{zhong2024agieval} & Evaluates LLMs on human-centric tasks against real-world standardized exams with straightforward probability and accuracy formulations. &  Rigorously assesses the general reasoning and problem-solving abilities of foundation models. \\
\hline
PICO \cite{ning2025pico} & Uses consistency optimization and assigns a learnable capability parameter to each LLM to maximize consistency in peer review. & Mimics the human peer‑review principles where LLMs evaluate each other’s answers, similar to academic peer review. \\
\hline
MT‑Bench \cite{zheng2023judging} &  Judge LLM (GPT‑4) assigns numerical scores (1–10) per rubric to assess a target LLM in 80 multi‑turn questions across 8 domains. & Evaluates conversational ability, contextual memory, and instruction following capabilities of LLMs in multi‑turn dialogues. \\
\hline
AlpacaEval \cite{alpaca_eval} & Judge LLM (GPT-4) computes pairwise win rate against a baseline across a test set of prompts. & Provides a fast, affordable, and human-aligned way to gauge the instruction-following and conversational capabilities of LLMs. \\
\hline
\end{tabular}%
}
\label{tab:llm_automatic_eval}
\end{table}
\subsection{Human Evaluation} Human evaluation of LLMs refers to the manual assessment of the correctness and quality of LLM-generated output via human judgment. Depending on the nature of the task and the intelligibility of explanations, human participants can be people with diverse technical backgrounds and domain knowledge, from ordinary users to experts. Overall, human evaluation of LLM explanations has three primary factors: the number of evaluators, evaluators' expertise level in that specific task, and the criteria for evaluating explanations \cite{chang2024survey, zhao2024explainability}. Given that people may have diverse functional and cognitive capabilities along with cultural and subjective preferences, their assessment of an explanation may be different. Consequently, having a sufficiently high number of people in judging LLM explanations may intuitively lead to more grounded and solid conclusions. Moreover, domain expertise may play a crucial role in assessing the quality of the explanation, and such studies should first identify people suitable for each specific explanation assessment. Finally, evaluation rubrics should include key factors for evaluating LLM explanations, as these metrics provide a framework for drawing conclusions from them.\\
While delivering LLM predictions and underlying explanations to target groups for their evaluation, the effectiveness of explanation conveyance depends not only on the model’s internal fidelity but also on the design of the appropriate human–machine interface (HMI) and human–computer interaction (HCI) \cite{he2025conversational, karny2026neural, zhou2026improving}. First, the interface must bridge the gap between the LLM’s complex reasoning and the user’s mental model to make explanations understandable \cite{jeck2025tell}. Second, HCI research underscores the importance of tailoring the interaction modality to the target group’s domain and expertise \cite{mindlin2025towards}. Third, the interface should support iterative follow-ups, allowing users to ask clarifying questions, challenge the LLM’s rationale, and adjust the granularity of explanations in real time \cite{zhou2026improving}. Ultimately, these requirements make HMI serve as a bidirectional passage. It translates LLM outputs into human‑understandable forms and captures user feedback to refine future

\begin{figure*}[!ht]
 \centering
    \includegraphics[width=\linewidth]{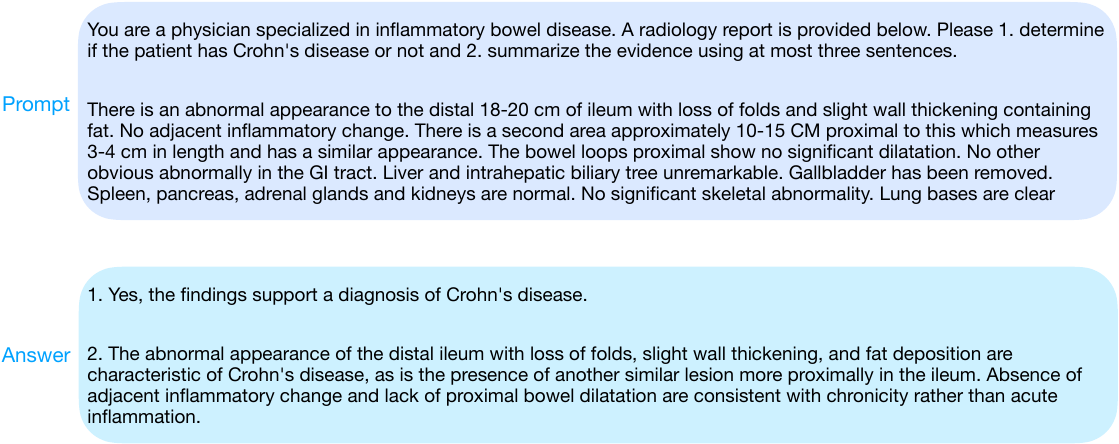}
    \caption{ChatGPT provides a diagnostic prediction and the corresponding evidence when provided with a radiology report on Crohn's disease. The prompt was tested on October 31st, 2024.}
    \label{fig:crohn_dis_prediction.pdf}
\end{figure*}
explanations. Consequently, HMIs play a significant role in transforming LLMs from opaque black boxes into collaborative tools and, hence, in human evaluation of these tools.

\subsection{Automatic Evaluation} Automation evaluation of LLM outputs typically involves using established metrics and tools to evaluate the quality and correctness of LLM-generated explanations automatically. These approaches are primarily used to provide a basis for standardized assessment, and save time and avoid potential impacts of humans' subjective preferences \cite{zhao2024explainability, chang2024survey}. Depending on the task and requirements, various automatic evaluation tools can be used for LLM explanations \cite{hu2024unveiling}. However, at its essence, automatic evaluation approaches must present accurate and factually correct explanations \cite{chang2024survey}: these evaluation approaches are \textit{accuracy} (e.g., Exact match, ROUGE score \cite{lin2004rouge}), \textit{faithfulness} \cite{matton2025walk}, \textit{calibrations} (e.g., classification accuracy, such as area under curve (AOC)), \textit{fairness} (e.g., equalized odds difference), and \textit{robustness} (e.g., attack success rate \cite{wang2021adversarial} and performance drop rate \cite{zhu2023promptrobust}). There have recently been attempts to develop methods for automating such evaluation approaches. For instance, G-Eval \cite{liu2023g} has been proposed to evaluate natural language generation (NLG) outputs with GPT-4 for better human alignment. Furthermore, LLM-EVAL \cite{chen2023llm} assesses open-domain conversations with LLMs. In addition, REM \cite{yang2024rem} leverages the collective intelligence approach using numerous LLMs through a ranking mechanism to evaluate LLM outputs. Other notable automatic evaluation approaches include PandaLM \cite{wang2024pandalm}, HELM \cite{liang2023holistic}, AGIEval \cite{zhong2024agieval}, PICO \cite{ning2025pico}, MT-Bench \cite{zheng2023judging} and AlpacaEval \cite{alpaca_eval}. More details are provided at Table \ref{tab:llm_automatic_eval}.  

\section{Empirical insights from critical applications}
Having presented local and mechanistic explanations within LLMs, along with critical epistemic LLM properties and assessment methods for explanations, we now unify these concepts in experimental insights from two safety-critical applications. The following subsections present empirical studies from two domains—healthcare and autonomous driving—and describe implications of LLM explanations from the trust viewpoint for targeted interaction partners in these domains.
\subsection{Insights from Healthcare}
XAI has been widely adopted in the medical and healthcare domain to identify the most contributing input features for the decision-support function of predictive models (e.g., for both disease diagnosis and prognosis). The features identified as important then serve as explanations or evidence for the predictions. As a predictive model trained on a particular dataset might exploit some features biased towards that dataset, a model with high predictive performance in a particular test dataset might not necessarily perform well with new data (e.g., the medical data from a different hospital or province); it is essential to gain insights on what features are leading to the predictions, so that a human decision-maker can determine the trustworthiness of the predictions. For example, in the 

\begin{figure*}[!ht]
 \centering
    \includegraphics[width=\linewidth]{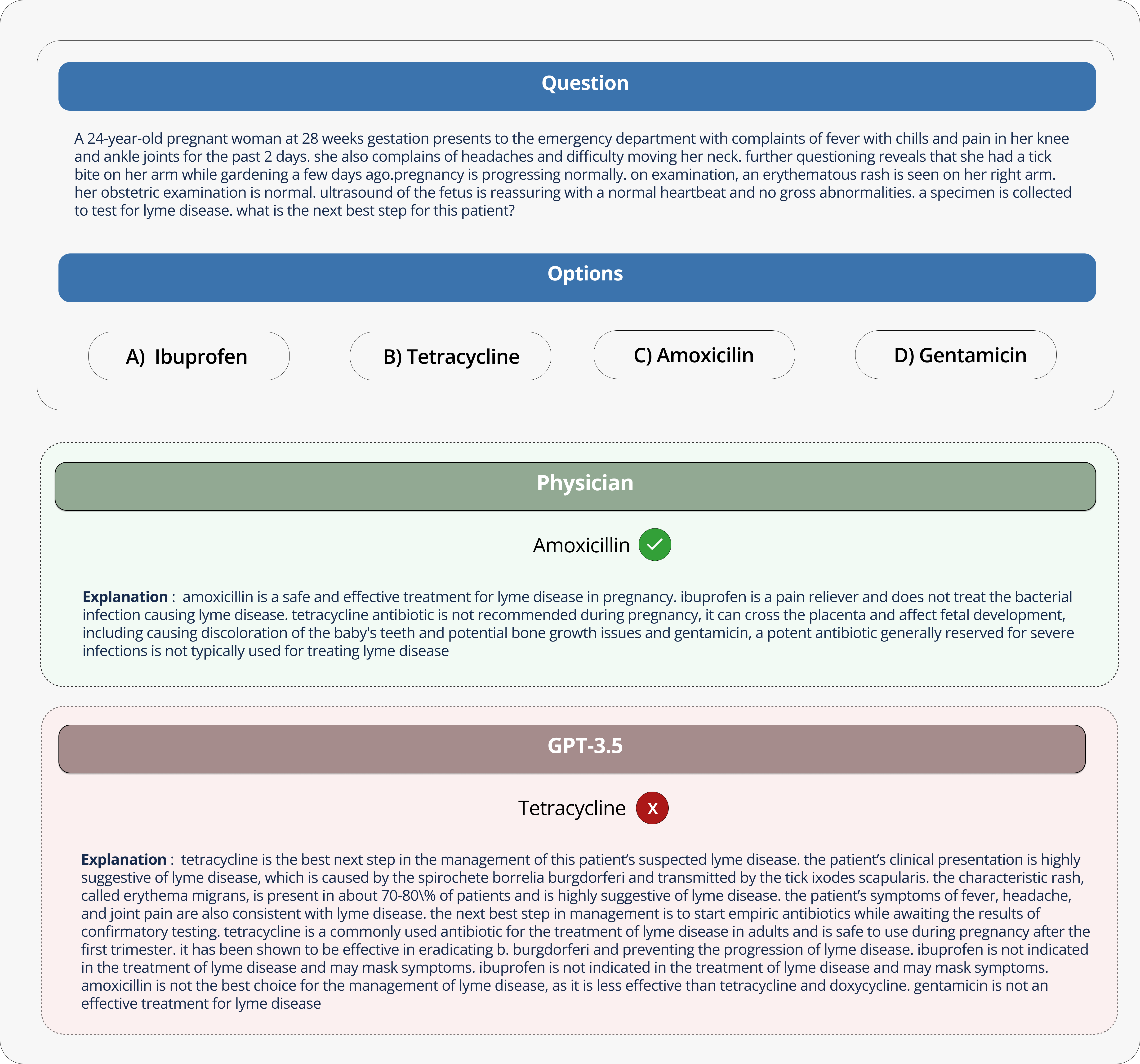}
    \caption{An example of a serious hallucination of GPT 3.5 in clinical diagnosis on the Med-HALT benchmark: Source: \cite{pal2023med}.}
    \label{fig:gpt_halluc_example}
\end{figure*}
particular case of survival analysis in medicine \cite{doi:10.1191/026921600701536192, kim2019deep} - where the task is to predict the time for a particular event to occur given an individual (e.g., death for a cancer patient) from various modalities of data (e.g., medical imaging, textual reports and tabular data) - post-hoc XAI methods, such as SHAP  and LIME are commonly employed to interpret how machine learning models make survival predictions for patients with cancer or those undergoing organ transplantation \cite{KOVALEV2020106164, Moncada-Torres2021-vq, terminassian2024explainableaisurvivalanalysis, KRZYZINSKI2023110234, doi:10.1200/CCI.20.00172, Liu2024-gf}. In the study of applying XAI to diagnostic predictions \cite{diagnostics12020237, JiayiDissertation, 10091536, 10.3389/fpubh.2022.874455}, the goal is to assist physicians and
doctors with their diagnostic decisions by not only providing predictions using machine learning-based models but also evidence for the predictions. Particularly in life-critical domains like medicine, explanations must accurately reflect the reasons behind a predictive model's decision-making (i.e., the faithfulness of the explanations). We claim that the faithfulness of explanations lies in the explicitness of the control over the predictive models' decision-making. To demonstrate this idea and the differences in the faithfulness of the explanations provided by popular existing XAI methods, we discuss the following three examples: (1) asking an LLM to provide justifications for its own predictions, (2) using LIME \cite{ribeiro2016should} to provide feature-importance values for individual tokens in a post-hoc manner and (3) using rationale extraction that learns to extract subsets of features as explanations via a select-predict architecture \cite{lei-etal-2016-rationalizing, dai2022interactive}. The corresponding examples are separately provided in Figures \ref{fig:crohn_dis_prediction.pdf}, \ref{fig:gpt_halluc_example}, \ref{fig:LIME_BERT_811}, and \ref{fig:RE_CTE}.
\paragraph{Asking an LLM to justify itself}
While asking an LLM to provide an answer for a given question, it is very intuitive to also ask the LLM to provide justifications due to its capability of generating human-readable texts, which is an example of natural language explanation or CoT reasoning

\begin{figure*}[!ht]
 \centering
    \includegraphics[width=12.5 cm]{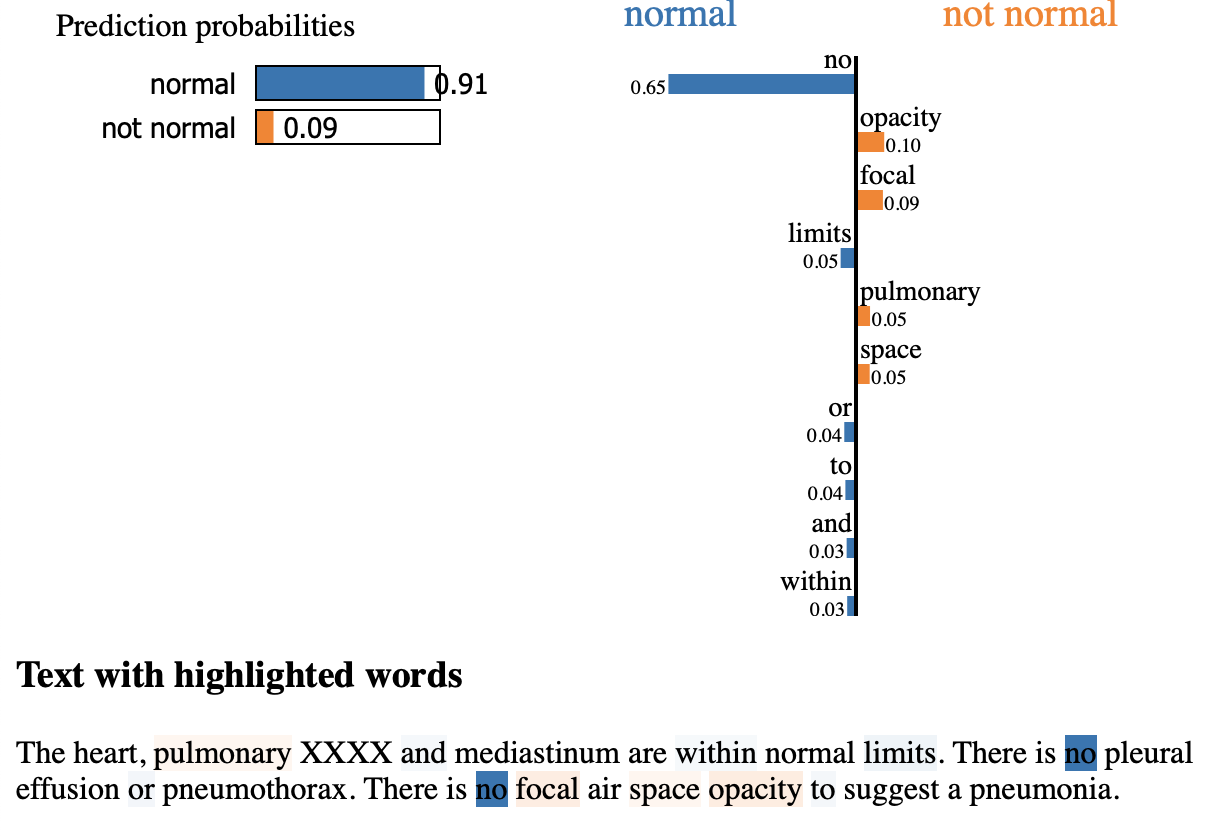}
    \caption{Running LIME on classifying a medical note with BERT.}
    \label{fig:LIME_BERT_811}
\end{figure*}
 described in Subsection 4.2. The resulting LLM-generated explanations might seem convincing to the users in some cases.
As an example, in Figure \ref{fig:crohn_dis_prediction.pdf}, ChatGPT was able to identify some abnormality on the distal ileum, which was in support of the diagnostic prediction of Crohn's disease. However, in another example shown in Figure \ref{fig:gpt_halluc_example},
GPT 3.5 fails to make a correct prediction along with its underlying explanation, which is a serious case of a hallucination \cite{pal2023med}. These two examples highlight the core problem with an LLM's general working principle: a major drawback of any explanation provided by an LLM via auto-regressive next-token prediction, where the explanation is simply a collection of the most probable tokens and lacks explicit control over the inference process towards the answer. This means we can not guarantee that the explanation is truly the reason for the answer. Hence, given the design of current LLMs, we could conclude that the faithfulness of an LLM's explanations can not be fulfilled by simply asking the LLM itself to explain. We previously noted that Transformer-based foundation models do not have any capacity for self-representation. Furthermore, another critical drawback of LLM-generated explanations is the parametric memory of LLMs for knowledge representation. This means that self-explanations generated by an LLM can hardly be verified without referring to external data sources.
\paragraph{Post-hoc explanation via linear approximation} LIME, as well as SHAP, as post-hoc XAI tools, have gained the most attention and popularity due to their model-agnostic capability of providing feature
importance scores as explanations. As shown in Figure \ref{fig:LIME_BERT_811}, LIME suggested that the word ``no" was the highest contributing feature for a BERT classifier to predict no abnormality. The identification of contributing features could provide some explanations for the decision-making of the predictive model. The mechanism of LIME is fairly simple. Given an input and a model (e.g., a medical report and a deep neural network, separately), to explain the model's output on this input, LIME learns a linear model as a surrogate of the original model in the space created by perturbing the features of the input. Then the coefficients of the linear model are treated as importance scores for the input features. Linear approximation, to a certain degree, gives insights on the behavior of the original model, but can not completely replace the original model (i.e., the approximation creates errors). For example, the perturbation in LIME is performed by randomly masking the features of the original input, which does not guarantee fair consideration of all subsets of features. In short, the explanations provided by a surrogate model may not truthfully explain the original model.
\paragraph{Rationale extraction that learns to explain}
Rationale extraction aims to learn an intermediate representation for selecting the features of training inputs that are identified as critical for making

\begin{figure*}[!ht]
 \centering
    \includegraphics[width=\linewidth]{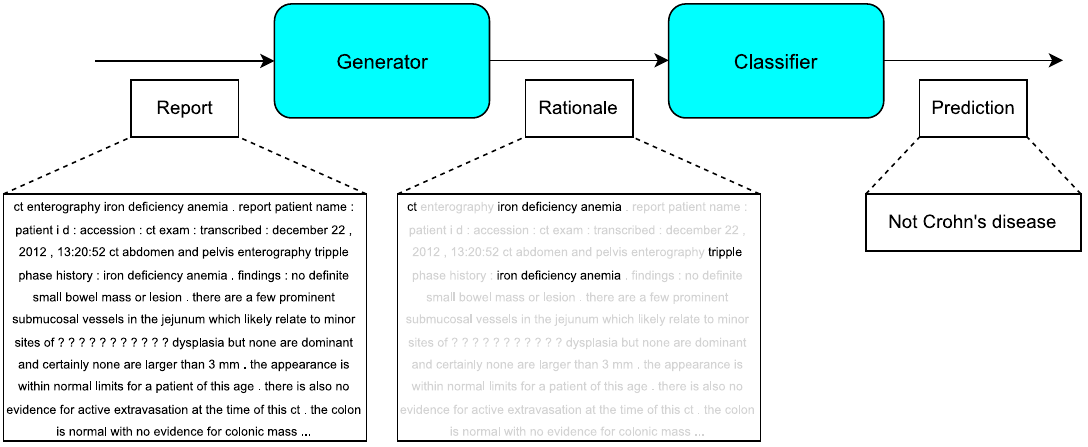}
    \caption{Rationale extraction selects a subset of features as a rationale for a diagnostic prediction on Crohn's disease from a computed tomography enterography report, where the prediction is correct and the corresponding explanation justifies the prediction. The example is from \cite{Dai_Kim_Sutton_Mitchell_Goebel_Baumgart_2025}.}
    \label{fig:RE_CTE}
\end{figure*}
predictions. The selected features are then used as the reason for a prediction. A rationale extraction architecture is typically composed of two neural networks, which are trained jointly. The first neural network (i.e., the generator) learns to select a subset of features from the original input, and the second neural network (i.e., the classifier) learns to make a prediction based on the selected subset
of features. Rationale extraction is performed with only instance-level supervision (as in a classic supervised learning setting) to search for the most important features. For example, in Figure \ref{fig:RE_CTE}, only some words were extracted from the original radiology report by the generator to be used as input for the classifier to make a final prediction. This architecture controls a neural network's decision-making by forcing the encoder to make a prediction solely based on the selected features, which guarantees that the selected features are truly the reason for the decision-making. These values indicate the contribution of individual features to the model’s decision. However, it is essential that these values accurately reflect the internal decision-making process of the model rather than being an approximation. So, while LLM-generated justifications can be intuitive and user-friendly, which brings great opportunities for providing more trustworthy AI-assisted medical diagnosis and prognosis, their faithfulness may vary. Further observations have highlighted that CoT reasoning with LLMs is often mistakenly considered as real reasoning, and can be misaligned with the models' actual computation \cite{barez-chain-2025}. Hence, it is essential to ensure these justifications are not merely plausible-sounding but are rooted in the actual reasoning process of the model. Thus, the faithfulness issue presents a fundamental challenge in applying LLM-based explanations in life-critical domains.

\subsection{Insights from Autonomous Driving}
Interactive explanations with human-machine interaction have recently been an area of interest for building trust in an autonomous vehicle (AV) \cite{marcu2024lingoqa, atakishiyev2024incorporating}. Understanding how an AV understands its operational surroundings and makes real-time decisions can help establish trust with onboard passengers. The question-answering (QA) mechanism, particularly in terms of visual question answering (VQA) and video question answering (VideoQA), has been an area of interest for providing human-interpretable explanations for justifying AV actions in a variety of actual and simulated traffic scenes. In this sense, we experiment with a multimodal LLM applied to an autonomous driving dataset and ask the model follow-up questions to justify its answers. As the focus of this paper is explainability from a language aspect, we focus on analyzing model predictions from the latter perspective. We use the Video-LLaMA2 multimodal LLM \cite{cheng2024videollama} to summarize temporal driving scenes from the Berkeley DeepDrive Attention (BDD-A) dataset \cite{xia2019predicting}, a video dataset featuring dynamic traffic and critical situations. We carefully sample four driving scenes from the dataset and ask explanatory questions for each scenario shown in Figure \ref{fig:traffic_scenarios}. The goal is to determine whether the model captures dynamic traffic environments with timing sensitivity and can explain its prediction of traffic scenes via intelligible, linguistic reasoning. 
Figure \ref{fig:traffic_scenarios} presents our experimental results. As seen, we first prompt the model to describe a scene and further ask causal, contrastive, and counterfactual questions. While the traditional approach for evaluating explanations has been to consider either

\begin{figure*}[!ht]
 \centering
    \includegraphics[width=\linewidth]{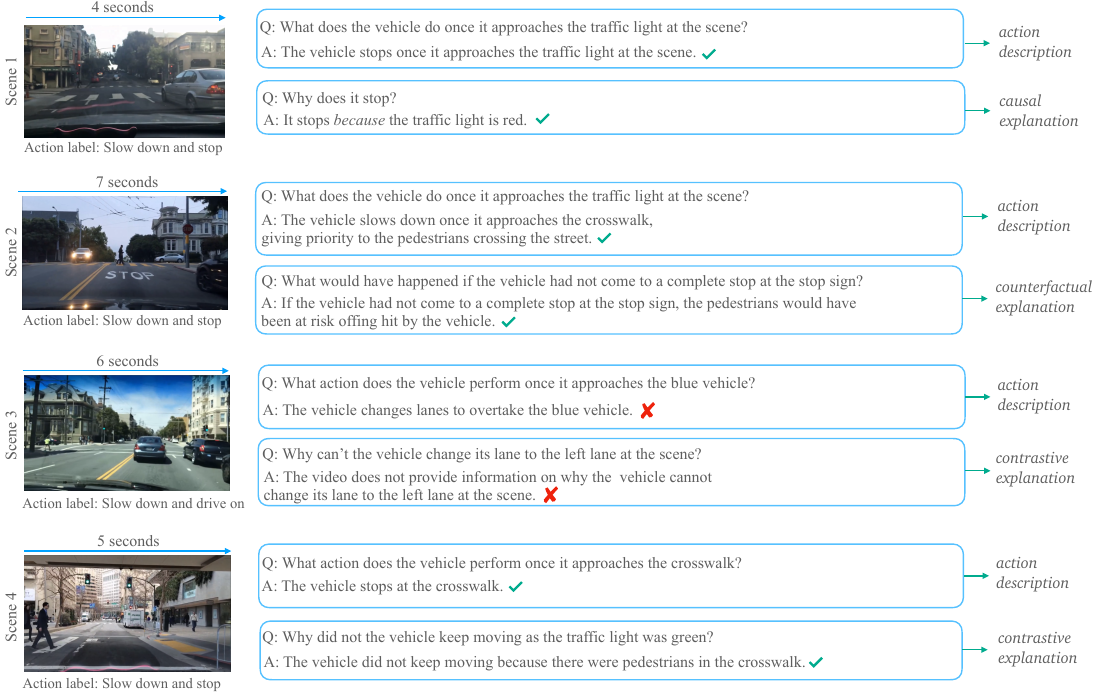}
    \caption{Scene understanding with the Video-LLaMA2 multimodal LLM: We first ask the model to explain action decisions/traffic scenes and further follow up with causal, contrastive, and counterfactual questions to stress test faithfulness of the previous responses.}
    \label{fig:traffic_scenarios}
\end{figure*}
descriptive or causal questions starting with the ``What" and ``Why" keywords, we instead argue that the model should also justify more challenging prompts ---  contrastive and counterfactual questions, as they are robustness test for models \cite{stepin2021survey},  to ensure that its emergent ability does not remain limited to simple tests. Traditionally, LLMs are constructed by fine-tuning labeled datasets of instructional prompts and corresponding outputs, and their generalization ability is usually tested using their built-in mechanism. However, we argue that a \textit{local explanation} is \textit{faithful} if the model also predicts correct answers in stress test prompts. In the selected scenes, once the model produces a response, we ask causal, counterfactual, and contrastive questions and observe that the model produces correct responses to these questions in Scenes 1, 2, and 4. In Scene 3, the model initially produces an incorrect explanation and cannot rationalize a contrastive query as the follow-up to the first query. Based on these insights, we can summarize that a local LLM explanation in this example can be considered trustworthy if 1) it passes the causal reasoning test, 2) it can justify contrastive and counterfactual questions correctly as a stress test, 3) it understands, distinguishes, and justify questions requiring fact or belief-based subjective answers accordingly, and 4) it avoids presenting falsified explanation in case it just cannot answer the question. The fourth property should be emphasized: depending on the data sources an LLM model is pretrained on, it may not capture information beyond its knowledge base, and acknowledging this limitation by simply avoiding the presentation of a fictitious explanation is as significant as the first three properties. Hence, we argue that at least these four properties should be inherent requirements of trustworthy local LLM explanations.

\section{Key findings and insights}
\subsection{LLM explanations have levels of granularity}

Our analysis of the presented empirical studies and insights from the literature shows that intelligibility of explanations may change from user to user, depending on their technical knowledge and how they use such explanations. Consequently, we categorize LLM explanations with \textit{three levels of granularity}: \\
1) \textit{Coarse-grained explanations}: Explanations within these levels can be understood by everyone, such as general users and regulators, as they do not contain deep technical information and rather

\begin{figure*}[!ht]
 \centering
    \includegraphics[width=\linewidth]{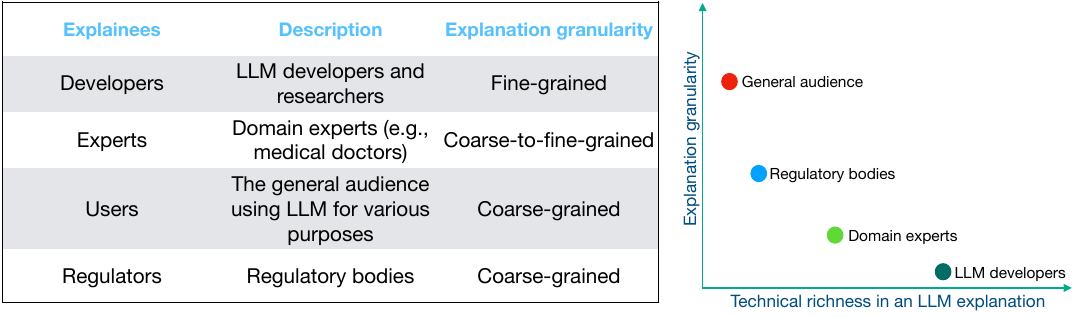}
    \caption{Granularity of LLM explanations with respect to the explainees}
    \label{fig:exp_gran}
\end{figure*}
focus on being intelligible and understandable by everyone. For instance, referring to Scenario 1 in Figure \ref{fig:traffic_scenarios}, the causal explanation to the ``Why?" question can be considered as an explanation with coarse granularity. \\
2) \textit{Coarse-to-fine-grained explanations}: Such explanations are primarily targeted for domain experts. Coarse-to-fine-grained explanations describe a model's predictions in which LLM first provides a coarse (high-level) overview of its reasoning — summarizing general patterns, modalities, or feature groups that drive a decision — and then progressively reveals fine-grained (detailed) evidence, such as specific features, regions, or textual tokens that substantiate the prediction. For instance, in the medical domain, LLM explanations for doctors are most effective when they follow a coarse-to-fine-grained structure, as clinical reasoning itself naturally progresses in that way: from general first impressions to specific, grounded evidence. For instance, if an LLM predicts that a patient has community-acquired pneumonia, a model's underlying explanation, such as ``\textit{The model identified a pneumonia-like pattern based on fever, cough symptoms, and chest X-ray abnormalities}," provides preliminary reasoning on the outcome. As a deeper detail,  ``\textit{Model's attention map localizes to a 2 cm opacity in the right lower lobe on the chest X-ray, and to the textual phrase ‘new infiltrate’ in the provided radiology report.}” would be fine-grained reasoning to support such an outcome, as potential symptoms of community-acquired pneumonia \cite{prina2015community}. Thus, such adaptive granularity, as in the medical field example, could also support trust in specific domains far more effectively than either purely coarse or purely fine explanations alone.  \\
3) \textit{Fine-grained explanations}: These explanations are highly detailed and technical, often delving into the inner workings of an LLM model, such as individual weights and activations in neural networks. All explanations on the mechanistic interpretability of LLMs are good examples of fine-grained granularity. This level is generally targeted at LLM developers or experts who need to understand the model at a granular level to debug or improve it and understand the inner workings of the entire model.\\

\subsection{LLM explanations must pass stress tests}
As stress tests are regularly used to evaluate the robustness and reliability of AI systems, LLM outputs should also be assessed with relevant checkpoints. As presented in the experiment in Figure \ref{fig:traffic_scenarios}, \textit{counterfactual reasoning} is one form of stress test exploring alternative scenarios for causal reasoning. Furthermore, LLM explanations can also be evaluated against \textit{benchmark datasets} dedicated to evaluating the robustness of various forms of explanations. A recent study shows that replacing keywords in questions results in the deterioration of the reasoning capabilities of seven well-known LLMs from OpenAI, Gemini, and LLaMA families \cite{wang-etal-2024-mmlu}. For this purpose, this study presents the MMLU-SR benchmark, enabling reasoning and understanding ability of LLM models by focusing on ``Question Only,” ``Answer Only,” and “Question and Answer” scenarios to ensure an LLM simply does not predict the next token but truly understands the concept and context in such predictions. Finally, it is important to test LLMs via specific \textit{stress prompts} to evaluate the capability of such models in the real world, where stress cases are highly anticipated. StressPrompt, a novel benchmark by \cite{shen2025stressprompt}, shows that carefully crafted specific prompts can alter the internal states of LLM models, leading to incorrect responses in critical applications. Analogously, another recent benchmark, Humanity's Last Exam \cite{phan2025humanity}, released in early 2025, shows that even the most advanced LLM models such as GPT-5 \cite{gpt5systemcard2025}, Grok-2 \cite{xai2024grok2}, Grok-4 \cite{xai2025grok4}, GPT-4o \cite{hurst2024gpt}, Gemini 2.5 \cite{comanici2025gemini}, Claude 4.5 Sonnet \cite{anthropic2025claude45}, and DeepSeek-R1 \cite{guo2025deepseek} perform below or around 25\% accuracy in predicting correct answers to 2,500

\begin{figure*}[!ht]
 \centering
    \includegraphics[width=\linewidth]{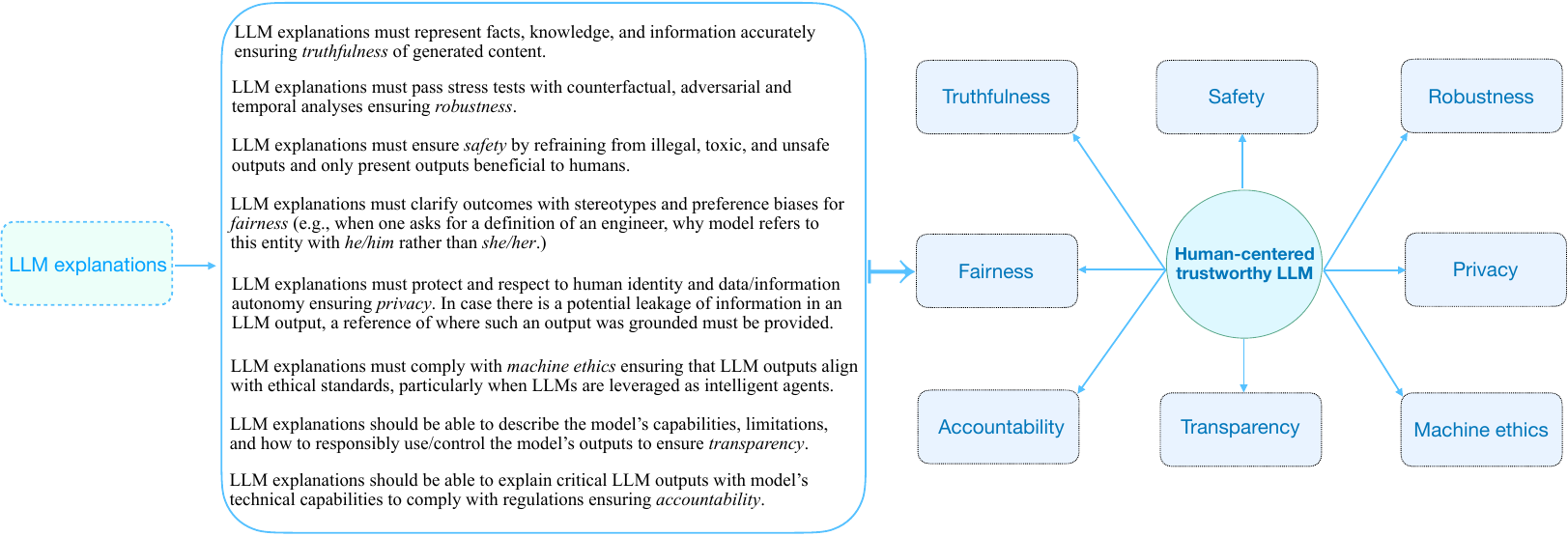}
    \caption{The eight principles of a human-centered trustworthy LLM and the role of LLM explanations in achieving these goals.}
    \label{fig:trustworhty_llm}
\end{figure*}
domain-specific difficult questions across various disciplines. Hence, the robustness of LLM explanations might be holistically assessed by the degree to which explanations remain consistent, faithful, and resistant to manipulation or noise across variations in input, context, or adversarial attacks, supported by the examples provided in Subsection 7.2 and \cite{shen2025stressprompt}. Whether targeted explainees are laypeople, general-domain users, or domain experts, truly understanding prompts/questions and predicting faithful, factually correct explanations for such queries is a paramount feature of a trustworthy LLM. 

\subsection{LLM explanations must support fundamental principles of TrustLLM}

A benchmark recently established by numerous institutions for trustworthy LLMs, named \textit{TrustLLM}, identifies eight fundamental principles of such models \cite{huang2024trustllm}. According to TrustLLM, a language model must consider \textit{safety, truthfulness, fairness, robustness, privacy, machine ethics, transparency}, and \textit{accountability} in its underlying working mechanism. These concepts have been investigated in general XAI research \cite{arrieta2020explainable, longo2024explainable}, so LLM explanations, in turn, must also align with these principles while being delivered to interaction partners. Below, we describe these nuances individually and in detail: \\
\textbf{Safety}: LLMs must ensure that their generated content does not harm a user in any way. The harmful prompt can be targeted at any user group or system in terms of jailbreak prompts, toxic queries, and the presentation of convincing but fictitious content in a safety-critical application. LLM explanations can help mitigate this undesirable behavior in a certain way. First, a \textit{built-in defense mechanism} can trigger an LLM to avoid presenting information about harmful queries (e.g., \textit{How can I make a \textless dangerous-item\textgreater ?}). Furthermore, if a user queries about a safety-critical topic (e.g., \textit{What medicine can I adopt for a \textless disease\textgreater and how many doses?}), a hallucination of an LLM in the response could have dire consequences if a user solely takes an action based on this response. To prevent such a case, an LLM can recommend a specific source or reference for a user's further look, as an explanation, which is a typical RAG approach. Recent acts and regulatory principles, such as the European Union's AI Act \cite{eu2024aiact}, the Singapore Consensus on Global AI Safety Research Priorities \cite{bengio2025singapore}, and International AI Safety Report \cite{bengio2025international} unanimously emphasize building safety guardrails on the LLM-generated content and underscore the substance of underlying explanations for what such models generate.\\
\textbf{Truthfulness:} \cite{huang2024trustllm} defines the term truthfulness as the \textit{accurate representation of information, facts, and results}. As LLMs are pre-trained with massive information on the Internet, (unintentional) hallucinations with their abilities may be related to non-factual information, outdated data, or simply relying on language prior to generate falsified content. Furthermore, an incorrect LLM output may also be the result of training an LLM model with intentionally falsified information as well \cite{pan2023risk, zhou2023synthetic}. \cite{huang2024trustllm} proposes that performing \textit{sycophancy test} \cite{fanous2025syceval, malmqvist2025sycophancy, li2025causally} in terms of persona and preference and testing them with adversarial queries are also crucial to validate the truthfulness of LLM outputs. In such cases, an LLM explanation can provide corrective measures, show incorrect details in the prompt, and present fact-based information by referring users to a specific source of information, as described in the previous section. \\
\textbf{Fairness:} Ensuring fairness in explanations has been one of the significant goals of XAI research, in general \cite{arrieta2020explainable}. The concern with the fairness issue is that LLM outputs might be biased, such as being against or inclined toward specific demographic groups, genders, and political figures \cite{huang2024trustllm}. So, LLM explanations must ensure the generated output does not contain discriminatory statements and present examples with equity when needed.\\
\textbf{Robustness:} We have covered the robustness perspective of LLM outputs with the experimental analysis above, but as a reiteration, we can conclude that LLM explanations should be robust against counterfactual analysis and stress prompts both in terms of informational content and in prompts requiring responses from a temporal analysis perspective.\\
\textbf{Privacy:} The primary concern with privacy in LLMs is data breaches in generated content and the jailbreaking of LLMs to spread confidential information \cite{huang2024trustllm, huang2022large, han2024medical, kim2024propile}. So, how can an LLM explanation help detect potential data breaches or leakage of confidential information? A simple solution can be to provide a reference or source information from which the content was generated. This way, the user could analyze the source information and judge whether LLM itself or the reference it was based on is a reason to spread privacy-concerning output. \\
\textbf{Machine Ethics:} The primary principles of machine ethics in AI research have been linked with Asimov's ``three laws of robotics" \cite{asimov1950runaround}. These principles have evolved over the decades and have further been adjusted or augmented to some extent \cite{moor2006nature}. Applying these principles to trustworthy LLMs, \cite{huang2024trustllm} posits that machine ethics principles can be investigated in terms of \textit{implicit ethics} (alignment with humans' general ethical standards),  \textit{explicit ethics} (an LLM's behavior in various ethical situations), and \textit{emotional awareness} (an LLM's ability to identify its true mission, abilities, and understand humans' emotions). Linking these concepts to explanations, a model can, for example, provide causal reasoning on why it made a particular prediction, and if possible, describe potential implications of the generated content to enlighten users with the potential sensitivity of the presented information.\\
\textbf{Transparency}: Transparency of an LLM output generally refers to the ability to understand and explain how an LLM model generates its outputs and decisions \cite{liao2024ai}. At first glance, this is a considerably challenging topic because of the intricate structure of an LLM model and the massive data on which they are trained. Furthermore, LLM predictions or decisions have or require various degrees of transparency \cite{huang2024trustllm}; hence, the level of transparency should be explored with human factors consideration with respect to who requires explanation with granularity \cite{barman2024beyond}. Although this is a general approach to transparent LLMs, the primary role of an explanation is to help various stakeholders achieve their desiderata with the LLM output presented to them \cite{langer2021we}.\\
\textbf{Accountability:} Finally, as LLMs are increasingly deployed across various domains and organizations for different purposes, the topic of accountability naturally arises as an actual problem with the usage of these models. Particularly, when an LLM makes a critical mistake, whether the organization that developed and deployed this model or a user that relied on the conclusion arrived at by such an LLM raises serious legal culpability issues (e.g., the example of an airline company's chatbot providing incorrect information on flight scheduling \footnote{https://www.cbc.ca/news/canada/british-columbia/air-canada-chatbot-lawsuit-1.7116416}). Consequently, AI acts on responsible AI development and deployment \cite{eu2024aiact, casonato2024ai} underscore accountability as a crucial feature of intelligent systems, underscoring \textit{ a high-risk AI system should be capable of explaining its output with its technical capabilities,"} and LLMs, as rapidly growing AI tools, must comply with these principles. Hence, based on these eight identified dimensions, we might infer that LLM explanations could serve as amplifiers, connecting the dots between a model's internal operations and its external trustworthiness, making those unanimously adopted principles verifiable.

\section{Future Directions}
Although the growing interest in mechanistic interpretability and local explainability of LLMs has been successful to some extent, there remain significant directions underexplored in generating trustworthy LLM explanations. In this regard, this section presents three directions for future exploration, necessitating collaborative work between LLM and XAI communities: \\
\textbf{Analyzing true logical reasoning in LLM explanations:} LLMs' remarkable content generation ability on different downstream language tasks are impressive at the first glance. However, the key question is whether these models truly reason over the prompt. There have been few studies on understanding particular logical reasoning abilities of language models, such as via inductive reasoning \cite{sinha2019clutrr}, temporal logic \cite{hahn2021teaching}, and deductive reasoning \cite{softreasoner_transformer}; however, holistic evaluation of language models via various first-order and non-monotonic logic approaches still remains unexplored. \textit{LogicBench} \cite{parmar2024logicbench} has recently been introduced to test and ameliorate logical reasoning abilities of advanced LLMs to some extent. Experimental results show that even advanced models from the GPT-family series (e.g., GPT-4 \cite{achiam2023gpt}), Gemini-Pro \cite{team2023gemini}, and LLaMA2 \cite{touvron2023llama2} do not perform well on LogicBench. In particular, they fail at complex reasoning tasks and sometimes overlook contextual information crucial to supporting the conclusion reached with sound reasoning. Furthermore, a recent study \cite{verma2025teaching} shows that guiding LLMs through explicit logical reasoning steps via Planning Domain Definition Language (PDDL)-based representation, such as checking action applicability, modeling state transitions, and verifying plan validity, could potentially bridge the gap between LLMs' general reasoning and the logical precision needed for automated planning. Hence, further exploration of LLMs' true reasoning ability is a significant future direction. \\
\textbf{Leveraging concept-bottleneck models for enhancing LLM interpretability:} The great majority of current LLMs produce output in a straightforward way based on the prompt, which does not specifically incorporate human-defined concepts into the reasoning process to arrive at the conclusions. \textit{Concept bottleneck models}, as a potential solution, have recently been introduced to enchance the interpretability of end-to-end deep neural networks, where the model first predicts the human-defined \textit{concepts}, and further uses only those predicted concepts in a \textit{bottleneck layer} to
make a final prediction \cite{koh2020concept, yuksekgonul2023posthoc, panousis2025coarse} (e.g., referring to an example from \cite{koh2020concept}: ``\textit{If the model did not think there was a bone spur in the x-ray, would it still predict severe arthritis?"}). The topic has been investigated across several domains to provide human-understandable explanatory information, such as in autonomous driving \cite{kenny2024explainable}, clinical tasks \cite{pang2024integrating}, and various object classification tasks \cite{yuksekgonul2023posthoc} with some empirical successes. However, concept bottleneck models have not been deeply investigated in the realm of LLMs and remain limited to a few but promising studies \cite{sun2025concept}. By breaking down intricate data into comprehensible components, such models could allow researchers and practitioners to scrutinize and understand the reasoning behind LLMs' outputs. In particular, for safety-critical applications and tasks, integrating concept bottleneck models into the reasoning process can help domain-specific LLMs be more transparent, human-understandable, and trustworthy. \\
\textbf{Performance-interpretability trade-off: Could LLMs suffer from it?} Model performance vs interpretability has been an actual, ongoing, and debatable topic in XAI research over the last couple of years. The ungrounded belief is that complex, interpretable ML models likely have low performance/accuracy \cite{rudin2019stop}. However, this nuance remains as a general view as the evaluation of explanation can be task-specific and human-centric, and depending on the nature of the task, this trade-off could be correct or incorrect to some extent \cite{arrieta2020explainable, saeed2023explainable, rudin2019stop}. Thinking of LLMs as a deep neural network-based model, that trade-off also arises with respect to these models' working mechanisms. In a recent study, \cite{kenny2024regulation} investigated the impact of incorporating interpretability constraints into LLMs, finding that while the interpretable model enhances human task efficiency and, in their case, fosters appropriately calibrated confidence, it also leads to a decline in the model's performance on an insurance liability classification task. Thus, as the topic is still in its infancy, the literature remains scarce in exploring interpretability-accuracy trade-offs in LLMs from a broader perspective and warrants further exploration.

\section{Conclusion}
In this paper, we have presented a comprehensive analysis of interpretability and explainability in the context of LLMs. Delving into the local explainability and mechanistic interpretability of LLMs, our study sheds light on opportunities and challenges for generating human-centered and trustworthy explanations. In conjunction with findings from related studies, our paper presents insights from safety-critical applications and outlines key points and steps to achieve faithful explanations while considering human factors. As general-purpose and domain-specific LLMs continue to proliferate across diverse applications, the need for trustworthy LLM explanations becomes increasingly critical. We believe that the proposed guidelines can help researchers and engaged target groups improve the human alignment and faithfulness of LLMs and promote the responsible use of these models in applied domains and tasks.  
\section*{Acknowledgment}
We acknowledge support from the Alberta Machine Intelligence Institute
(Amii), from the Computing Science Department of the University of Alberta,
and the Natural Sciences and Engineering Research Council of Canada (NSERC).

\bibliographystyle{unsrt}
\bibliography{main}

@article{marra2024statistical,
  title={{From statistical relational to neurosymbolic artificial intelligence: A survey}},
  author={Marra, Giuseppe and Duman{\v{c}}i{\'c}, Sebastijan and Manhaeve, Robin and De Raedt, Luc},
  journal={Artificial Intelligence},
  volume={328},
  pages={104062},
  year={2024},
  publisher={Elsevier}
}

@article{garcez2023neurosymbolic,
  title={Neurosymbolic {AI}: the 3rd wave},
  author={d'Avila Garcez, Artur S. and Lamb, Lu{\'\i}s C.},
  journal={Artificial Intelligence Review},
  volume={56},
  number={11},
  pages={12387--12406},
  year={2023},
  publisher={Springer},
  doi={10.1007/s10462-023-10448-w}
}

@article{alchourron1985logic,
  author = {Carlos E. Alchourrón and Peter Gärdenfors and David Makinson},
  title = {On the Logic of Theory Change: Partial Meet Contraction and Revision Functions},
  journal = {Journal of Symbolic Logic},
  volume = {50},
  number = {2},
  pages = {510--530},
  year = {1985},
  doi = {10.2307/2274239},
  publisher = {Cambridge University Press}
}

@article{ney1994structuring,
  title={On structuring probabilistic dependences in stochastic language modelling},
  author={Ney, Hermann and Essen, Ute and Kneser, Reinhard},
  journal={Computer Speech \& Language},
  volume={8},
  number={1},
  pages={1--38},
  year={1994},
  publisher={Elsevier}
}

@article{mikolov2013distributed,
  title={{Distributed Representations of Words and Phrases and their Compositionality}},
  author={Mikolov, Tomas and Sutskever, Ilya and Chen, Kai and Corrado, Greg S and Dean, Jeff},
  journal={Advances in Neural Information Processing Systems},
  volume={26},
  year={2013}
}

@article{faruqui2014retrofitting,
  title={Retrofitting word vectors to semantic lexicons},
  author={Faruqui, Manaal and Dodge, Jesse and Jauhar, Sujay K and Dyer, Chris and Hovy, Eduard and Smith, Noah A},
  journal={arXiv preprint arXiv:1411.4166},
  year={2014}
}

@article{vaswani2017attention,
  title={{Attention Is All You Need}},
  author={Vaswani, A},
  journal={Advances in Neural Information Processing Systems},
  year={2017}
}

@inproceedings{kenton2019bert,
  title={{BERT: Pre-training of Deep Bidirectional Transformers for Language Understanding}},
  author={Devlin, Jacob and Chang, Ming-Wei and Lee, Kenton and Toutanova, Kristina},
  booktitle={{Proceedings of the 2019 Conference of the North American Chapter of the Association for Computational Linguistics: Human Language Technologies, Volume 1 (Long and Short Papers)}},
  pages={4171--4186},
  year={2019}
}

@article{lewis2020retrieval,
  title={{Retrieval-Augmented Generation for Knowledge-Intensive NLP Tasks}},
  author={Lewis, Patrick and Perez, Ethan and Piktus, Aleksandra and Petroni, Fabio and Karpukhin, Vladimir and Goyal, Naman and K{\"u}ttler, Heinrich and Lewis, Mike and Yih, Wen-tau and Rockt{\"a}schel, Tim and others},
  journal={Advances in Neural Information Processing Systems},
  volume={33},
  pages={9459--9474},
  year={2020}
}

@article{wu2023brief,
  title={A brief overview of ChatGPT: The history, status quo and potential future development},
  author={Wu, Tianyu and He, Shizhu and Liu, Jingping and Sun, Siqi and Liu, Kang and Han, Qing-Long and Tang, Yang},
  journal={IEEE/CAA Journal of Automatica Sinica},
  volume={10},
  number={5},
  pages={1122--1136},
  year={2023},
  publisher={IEEE}
}

@article{achiam2023gpt,
  title={{GPT-4 Technical Report}},
  author={Achiam, Josh and Adler, Steven and Agarwal, Sandhini and Ahmad, Lama and Akkaya, Ilge and Aleman, Florencia Leoni and Almeida, Diogo and Altenschmidt, Janko and Altman, Sam and Anadkat, Shyamal and others},
  journal={arXiv preprint arXiv:2303.08774},
  year={2023}
}

@article{DBLP:journals/corr/abs-1802-05365,
  author       = {Matthew E. Peters and
                  Mark Neumann and
                  Mohit Iyyer and
                  Matt Gardner and
                  Christopher Clark and
                  Kenton Lee and
                  Luke Zettlemoyer},
  title        = {Deep contextualized word representations},
  journal      = {CoRR},
  volume       = {abs/1802.05365},
  year         = {2018},
  url          = {http://arxiv.org/abs/1802.05365},
  eprinttype    = {arXiv},
  eprint       = {1802.05365},
  bibsource    = {dblp computer science bibliography, https://dblp.org}
}

@inproceedings{cer2018universal,
  title={Universal sentence encoder for English},
  author={Cer, Daniel and Yang, Yinfei and Kong, Sheng-yi and Hua, Nan and Limtiaco, Nicole and John, Rhomni St and Constant, Noah and Guajardo-Cespedes, Mario and Yuan, Steve and Tar, Chris and others},
  booktitle={Proceedings of the 2018 conference on empirical methods in natural language processing: system demonstrations},
  pages={169--174},
  year={2018}
}

@inproceedings{farruque2019augmenting,
  title={Augmenting semantic representation of depressive language: From forums to microblogs},
  author={Farruque, Nawshad and Zaiane, Osmar and Goebel, Randy},
  booktitle={Joint European conference on machine learning and knowledge discovery in databases},
  pages={359--375},
  year={2019},
  organization={Springer}
}

@article{bojanowski2017enriching,
  title={Enriching Word Vectors with Subword Information},
  author={Bojanowski, Piotr and Grave, Edouard and Joulin, Armand and Mikolov, Tomas},
  journal={Transactions of the Association for Computational Linguistics},
  volume={5},
  year={2017},
  issn={2307-387X},
  pages={135--146}
}

@article{bengio2000neural,
  title={{A Neural Probabilistic Language Model}},
  author={Bengio, Yoshua and Ducharme, R{\'e}jean and Vincent, Pascal},
  journal={Advances in Neural Information Processing Systems},
  volume={13},
  year={2000}
}

@article{rumelhart1986learning,
  title={Learning representations by back-propagating errors},
  author={Rumelhart, David E and Hinton, Geoffrey E and Williams, Ronald J},
  journal={Nature},
  volume={323},
  number={6088},
  pages={533--536},
  year={1986},
  publisher={Nature Publishing Group UK London}
}

@article{brown1992class,
  title={Class-based n-gram models of natural language},
  author={Brown, Peter F and Della Pietra, Vincent J and Desouza, Peter V and Lai, Jennifer C and Mercer, Robert L},
  journal={Computational linguistics},
  volume={18},
  number={4},
  pages={467--480},
  year={1992}
}

@article{luo2024understanding,
  title={{From Understanding to Utilization: A Survey on Explainability for Large Language Models}},
  author={Luo, Haoyan and Specia, Lucia},
  journal={arXiv preprint arXiv:2401.12874},
  year={2024}
}

@article{singh2024rethinking,
  title={{Rethinking Interpretability in the Era of Large Language Models}},
  author={Singh, Chandan and Inala, Jeevana Priya and Galley, Michel and Caruana, Rich and Gao, Jianfeng},
  journal={arXiv preprint arXiv:2402.01761},
  year={2024}
}

@inproceedings{singh2024enhancing,
  title={Enhancing AI Systems with Agentic Workflows Patterns in Large Language Model},
  author={Singh, Aditi and Ehtesham, Abul and Kumar, Saket and Khoei, Tala Talaei},
  booktitle={2024 IEEE World AI IoT Congress (AIIoT)},
  pages={527--532},
  year={2024},
  organization={IEEE}
}

@article{elhage2022toy,
  title={Toy models of superposition},
  author={Elhage, Nelson and Hume, Tristan and Olsson, Catherine and Schiefer, Nicholas and Henighan, Tom and Kravec, Shauna and Hatfield-Dodds, Zac and Lasenby, Robert and Drain, Dawn and Chen, Carol and others},
  journal={arXiv preprint arXiv:2209.10652},
  year={2022}
}

@inproceedings{chen2024benchmarking,
  title={{Benchmarking Large Language Models in Retrieval-Augmented Generation}},
  author={Chen, Jiawei and Lin, Hongyu and Han, Xianpei and Sun, Le},
  booktitle={Proceedings of the AAAI Conference on Artificial Intelligence},
  volume={38},
  number={16},
  pages={17754--17762},
  year={2024}
}

@article{gao2023retrieval,
  title={{Retrieval-Augmented Generation for Large Language Models: A Survey}},
  author={Gao, Yunfan and Xiong, Yun and Gao, Xinyu and Jia, Kangxiang and Pan, Jinliu and Bi, Yuxi and Dai, Yi and Sun, Jiawei and Wang, Haofen},
  journal={arXiv preprint arXiv:2312.10997},
  year={2023}
}

@inproceedings{goebel2018explainable,
  title={{Explainable AI: the new 42?}},
  author={Goebel, Randy and Chander, Ajay and Holzinger, Katharina and Lecue, Freddy and Akata, Zeynep and Stumpf, Simone and Kieseberg, Peter and Holzinger, Andreas},
  booktitle={{International Cross-Domain Conference for Machine Learning and Knowledge Extraction}},
  pages={295--303},
  year={2018},
  organization={Springer}
}

@article{confalonieri2021historical,
  title={{A historical perspective of explainable Artificial Intelligence}},
  author={Confalonieri, Roberto and Coba, Ludovik and Wagner, Benedikt and Besold, Tarek R},
  journal={Wiley Interdisciplinary Reviews: Data Mining and Knowledge Discovery},
  volume={11},
  number={1},
  pages={e1391},
  year={2021},
  publisher={Wiley Online Library}
}

@article{kim2021multi,
  title={{A Multi-Component Framework for the Analysis and Design of Explainable Artificial Intelligence}},
  author={Kim, Mi-Young and Atakishiyev, Shahin and Babiker, Housam Khalifa Bashier and Farruque, Nawshad and Goebel, Randy and Za{\"\i}ane, Osmar R and Motallebi, Mohammad-Hossein and Rabelo, Juliano and Syed, Talat and Yao, Hengshuai and others},
  journal={Machine Learning and Knowledge Extraction},
  volume={3},
  number={4},
  pages={900--921},
  year={2021},
  publisher={MDPI}
}

@article{zhao2024explainability,
  title={{Explainability for Large Language Models: A Survey}},
  author={Zhao, Haiyan and Chen, Hanjie and Yang, Fan and Liu, Ninghao and Deng, Huiqi and Cai, Hengyi and Wang, Shuaiqiang and Yin, Dawei and Du, Mengnan},
  journal={ACM Transactions on Intelligent Systems and Technology},
  volume={15},
  number={2},
  pages={1--38},
  year={2024},
  publisher={ACM New York, NY}
}

@article{brown2020language,
  title={Language models are few-shot learners},
  author={Brown, Tom and Mann, Benjamin and Ryder, Nick and Subbiah, Melanie and Kaplan, Jared D and Dhariwal, Prafulla and Neelakantan, Arvind and Shyam, Pranav and Sastry, Girish and Askell, Amanda and others},
  journal={Advances in Neural Information Processing Systems},
  volume={33},
  pages={1877--1901},
  year={2020}
}

@article{touvron2023llama2,
  title={{Llama 2: Open Foundation and Fine-Tuned Chat Models
}},
  author={Touvron, Hugo and Martin, Louis and Stone, Kevin and Albert, Peter and Almahairi, Amjad and Babaei, Yasmine and Bashlykov, Nikolay and Batra, Soumya and Bhargava, Prajjwal and Bhosale, Shruti and others},
  journal={arXiv preprint arXiv:2307.09288},
  year={2023}
}

@misc{anthropic_claude,
  title={{Introducing Claude}},
  author={Anthropic},
  howpublished = {https://www.anthropic.com/news/introducing-claude},
  year={2023}
}

@misc{vicuna2023,
    title = {{Vicuna: An Open-Source Chatbot Impressing GPT-4 with 90\%* ChatGPT Quality}},
    url = {https://lmsys.org/blog/2023-03-30-vicuna/},
    author = {Chiang, Wei-Lin and Li, Zhuohan and Lin, Zi and Sheng, Ying and Wu, Zhanghao and Zhang, Hao and Zheng, Lianmin and Zhuang, Siyuan and Zhuang, Yonghao and Gonzalez, Joseph E. and Stoica, Ion and Xing, Eric P.},
    month = {March},
    year = {2023}
}

@article{touvron2023llama,
  title={{LLaMA: Open and Efficient Foundation Language Models}},
  author={Touvron, Hugo and Lavril, Thibaut and Izacard, Gautier and Martinet, Xavier and Lachaux, Marie-Anne and Lacroix, Timoth{\'e}e and Rozi{\`e}re, Baptiste and Goyal, Naman and Hambro, Eric and Azhar, Faisal and others},
  journal={arXiv preprint arXiv:2302.13971},
  year={2023}
}

@inproceedings{driess2023palm,
  title={{PaLM-E: An Embodied Multimodal Language Model}},
  author={Driess, Danny and Xia, Fei and Sajjadi, Mehdi SM and Lynch, Corey and Chowdhery, Aakanksha and Ichter, Brian and Wahid, Ayzaan and Tompson, Jonathan and Vuong, Quan and Yu, Tianhe and others},
  booktitle={International Conference on Machine Learning},
  pages={8469--8488},
  year={2023},
  organization={PMLR}
}

@article{team2023gemini,
  title={{Gemini: A Family of Highly Capable Multimodal Models}},
  author={Anil, Rohan and Borgeaud, Sebastian and Wu, Yonghui and Alayrac, Jean-Baptiste and Yu, Jiahui and Soricut, Radu and Schalkwyk, Johan and Dai, Andrew M and Hauth, Anja and others},
  journal={arXiv preprint arXiv:2312.11805},
  year={2023}
}

@article{wei2022emergent,
  title={{Emergent Abilities of Large Language Models}},
  author={Wei, Jason and Tay, Yi and Bommasani, Rishi and Raffel, Colin and Zoph, Barret and Borgeaud, Sebastian and Yogatama, Dani and Bosma, Maarten and Zhou, Denny and Metzler, Donald and others},
  journal={Transactions on Machine Learning Research},
  year={2022}
}

@article{haller2023opiniongpt,
  title={{OpinionGPT: Modelling Explicit Biases in Instruction-Tuned LLMs}},
  author={Haller, Patrick and Aynetdinov, Ansar and Akbik, Alan},
  journal={arXiv preprint arXiv:2309.03876},
  year={2023}
}

@article{liu2024large,
  title={{Large Language Models and Causal Inference in Collaboration: A Comprehensive Survey}},
  author={Liu, Xiaoyu and Xu, Paiheng and Wu, Junda and Yuan, Jiaxin and Yang, Yifan and Zhou, Yuhang and Liu, Fuxiao and Guan, Tianrui and Wang, Haoliang and Yu, Tong and others},
  journal={arXiv preprint arXiv:2403.09606},
  year={2024}
}

@article{doi:10.1191/026921600701536192,
author = {Antonio Viganò and Marlene Dorgan and Jeanette Buckingham and Eduardo Bruera and Maria E Suarez-Almazor},
title ={Survival prediction in terminal cancer patients: a systematic review of the medical literature},

journal = {Palliative Medicine},
volume = {14},
number = {5},
pages = {363-374},
year = {2000},
doi = {10.1191/026921600701536192},
    note ={PMID: 11064783},

URL = { 
        https://doi.org/10.1191/026921600701536192
},
eprint = { 
        https://doi.org/10.1191/026921600701536192
}
}

@article{kim2019deep,
  title={Deep learning-based survival prediction of oral cancer patients},
  author={Kim, Dong Wook and Lee, Sanghoon and Kwon, Sunmo and Nam, Woong and Cha, In-Ho and Kim, Hyung Jun},
  journal={Scientific Reports},
  volume={9},
  number={1},
  pages={6994},
  year={2019},
  publisher={Nature Publishing Group UK London}
}

@article{huang2023can,
  title={{Can Large Language Models Explain Themselves? A Study of LLM-Generated Self-Explanations}},
  author={Huang, Shiyuan and Mamidanna, Siddarth and Jangam, Shreedhar and Zhou, Yilun and Gilpin, Leilani H},
  journal={arXiv preprint arXiv:2310.11207},
  year={2023}
}

@article{KOVALEV2020106164,
title = {SurvLIME: A method for explaining machine learning survival models},
journal = {Knowledge-Based Systems},
volume = {203},
pages = {106164},
year = {2020},
issn = {0950-7051},
doi = {https://doi.org/10.1016/j.knosys.2020.106164},
url = {https://www.sciencedirect.com/science/article/pii/S0950705120304044},
author = {Maxim S. Kovalev and Lev V. Utkin and Ernest M. Kasimov}
}

@ARTICLE{Moncada-Torres2021-vq,
  title     = "Explainable machine learning can outperform Cox regression
               predictions and provide insights in breast cancer survival",
  author    = "Moncada-Torres, Arturo and van Maaren, Marissa C and Hendriks,
               Mathijs P and Siesling, Sabine and Geleijnse, Gijs",
  journal   = "Sci. Rep.",
  publisher = "Springer Science and Business Media LLC",
  volume    =  11,
  number    =  1,
  pages     = "6968",
  month     =  mar,
  year      =  2021,
  copyright = "https://creativecommons.org/licenses/by/4.0",
  language  = "en"
}

@misc{terminassian2024explainableaisurvivalanalysis,
      title={Explainable AI for survival analysis: a median-SHAP approach}, 
      author={Lucile Ter-Minassian and Sahra Ghalebikesabi and Karla Diaz-Ordaz and Chris Holmes},
      year={2024},
      eprint={2402.00072},
      archivePrefix={arXiv},
      primaryClass={cs.LG},
      url={https://arxiv.org/abs/2402.00072}, 
}

@article{KRZYZINSKI2023110234,
title = {SurvSHAP(t): Time-dependent explanations of machine learning survival models},
journal = {Knowledge-Based Systems},
volume = {262},
pages = {110234},
year = {2023},
issn = {0950-7051},
author = {Mateusz Krzyziński and Mikołaj Spytek and Hubert Baniecki and Przemysław Biecek},
}

@article{doi:10.1200/CCI.20.00172,
author = {Sundrani, Sameer and Lu, James},
title = {Computing the Hazard Ratios Associated With Explanatory Variables Using Machine Learning Models of Survival Data},
journal = {JCO Clinical Cancer Informatics},
volume = {},
number = {5},
pages = {364-378},
year = {2021},
doi = {10.1200/CCI.20.00172},
    note ={PMID: 33797958},

URL = { 
    
        https://doi.org/10.1200/CCI.20.00172
    
    

},
eprint = { 
    
        https://doi.org/10.1200/CCI.20.00172
    
    

},
}

@ARTICLE{Liu2024-gf,
  title     = "Predicting skin cancer risk from facial images with an
               explainable artificial intelligence ({XAI}) based approach: a
               proof-of-concept study",
  author    = "Liu, Xianjing and Sangers, Tobias E and Nijsten, Tamar and
               Kayser, Manfred and Pardo, Luba M and Wolvius, Eppo B and
               Roshchupkin, Gennady V and Wakkee, Marlies",
  journal   = "EClinicalMedicine",
  publisher = "Elsevier BV",
  volume    =  71,
  number    =  102550,
  pages     = "102550",
  month     =  may,
  year      =  2024,
  copyright = "http://creativecommons.org/licenses/by-nc-nd/4.0/",
  language  = "en"
}

@article{polanyi,
  title={Personal Knowledge : Towards a Post-Critical Philosophy},
  author={Michael Polanyi},
  journal={Chicago: University of Chicago Press},
  year={1958}
}

@article{saeed2023explainable,
  title={{Explainable AI (XAI): A systematic meta-survey of current challenges and future opportunities}},
  author={Saeed, Waddah and Omlin, Christian},
  journal={Knowledge-Based Systems},
  volume={263},
  pages={110273},
  year={2023},
  publisher={Elsevier}
}

@book{buchanan1984rule,
  title={Rule based expert systems: the mycin experiments of the stanford heuristic programming project (the Addison-Wesley series in artificial intelligence)},
  author={Buchanan, Bruce G and Shortliffe, Edward H},
  year={1984},
  publisher={Addison-Wesley Longman Publishing Co., Inc.}
}

@article{chang2024survey,
  title={{A Survey on Evaluation of Large Language Models}},
  author={Chang, Yupeng and Wang, Xu and Wang, Jindong and Wu, Yuan and Yang, Linyi and Zhu, Kaijie and Chen, Hao and Yi, Xiaoyuan and Wang, Cunxiang and Wang, Yidong and others},
  journal={ACM Transactions on Intelligent Systems and Technology},
  volume={15},
  number={3},
  pages={1--45},
  year={2024},
  publisher={ACM New York, NY}
}

@article{raiaan2024review,
  title={{A Review on Large Language Models: Architectures, Applications, Taxonomies, Open Issues and Challenges}},
  author={Raiaan, Mohaimenul Azam Khan and Mukta, Md Saddam Hossain and Fatema, Kaniz and Fahad, Nur Mohammad and Sakib, Sadman and Mim, Most Marufatul Jannat and Ahmad, Jubaer and Ali, Mohammed Eunus and Azam, Sami},
  journal={IEEE Access},
  year={2024},
  publisher={IEEE}
}

@inproceedings{ribeiro2016should,
  title={{"Why Should I Trust You?": Explaining the Predictions of Any Classifier}},
  author={Ribeiro, Marco Tulio and Singh, Sameer and Guestrin, Carlos},
  booktitle={Proceedings of the 22nd ACM SIGKDD International Conference on Knowledge Discovery and Data Mining},
  pages={1135--1144},
  year={2016}
}

@article{bricken2023monosemanticity,
       title={{Towards Monosemanticity: Decomposing Language Models With Dictionary Learning}},
       author={Bricken, Trenton and Templeton, Adly and Batson, Joshua and Chen, Brian and Jermyn, Adam and Conerly, Tom and Turner, Nick and Anil, Cem and Denison, Carson and Askell, Amanda and Lasenby, Robert and Wu, Yifan and Kravec, Shauna and Schiefer, Nicholas and Maxwell, Tim and Joseph, Nicholas and Hatfield-Dodds, Zac and Tamkin, Alex and Nguyen, Karina and McLean, Brayden and Burke, Josiah E and Hume, Tristan and Carter, Shan and Henighan, Tom and Olah, Christopher},
       year={2023},
       journal={Transformer Circuits Thread},
       note={https://transformer-circuits.pub/2023/monosemantic-features/index.html}
    }

@article{templeton2024scaling,
       title={{Scaling Monosemanticity: Extracting Interpretable Features from Claude 3 Sonnet}},
       author={Templeton, Adly and Conerly, Tom and Marcus, Jonathan and Lindsey, Jack and Bricken, Trenton and Chen, Brian and Pearce, Adam and Citro, Craig and Ameisen, Emmanuel and Jones, Andy and Cunningham, Hoagy and Turner, Nicholas L and McDougall, Callum and MacDiarmid, Monte and Freeman, C. Daniel and Sumers, Theodore R. and Rees, Edward and Batson, Joshua and Jermyn, Adam and Carter, Shan and Olah, Chris and Henighan, Tom},
       year={2024},
       journal={Transformer Circuits Thread},
       url={https://transformer-circuits.pub/2024/scaling-monosemanticity/index.html}
    }

@mastersthesis{JiayiDissertation,
    author = {Dai, Jiayi},
    title = {{Rationale Extraction and Crohn’s Disease Detection from Computed Tomography Enterography Reports}},
    school = {University of Alberta},
    year = {2023}
}

@Article{diagnostics12020237,
AUTHOR = {Zhang, Yiming and Weng, Ying and Lund, Jonathan},
TITLE = {Applications of Explainable Artificial Intelligence in Diagnosis and Surgery},
JOURNAL = {Diagnostics},
VOLUME = {12},
YEAR = {2022},
NUMBER = {2},
ARTICLE-NUMBER = {237},
URL = {https://www.mdpi.com/2075-4418/12/2/237},
PubMedID = {35204328},
ISSN = {2075-4418},
DOI = {10.3390/diagnostics12020237}
}

@ARTICLE{10091536,
  author={Moreno-Sánchez, Pedro A.},
  journal={IEEE Access}, 
  title={Data-Driven Early Diagnosis of Chronic Kidney Disease: Development and Evaluation of an Explainable AI Model}, 
  year={2023},
  volume={11},
  number={},
  pages={38359-38369},
  doi={10.1109/ACCESS.2023.3264270}}

@ARTICLE{10.3389/fpubh.2022.874455,

AUTHOR={Gong, Houwu  and Wang, Miye  and Zhang, Hanxue  and Elahe, Md Fazla  and Jin, Min },

TITLE={An Explainable AI Approach for the Rapid Diagnosis of COVID-19 Using Ensemble Learning Algorithms},

JOURNAL={Frontiers in Public Health},

VOLUME={10},

YEAR={2022},

URL={https://www.frontiersin.org/journals/public-health/articles/10.3389/fpubh.2022.874455},

DOI={10.3389/fpubh.2022.874455},

ISSN={2296-2565}}

@inproceedings{
dai2022interactive,
title={Interactive Rationale Extraction for Text Classification},
author={Jiayi Dai and Mi-Young Kim and Randy Goebel},
booktitle={Workshop on Trustworthy and Socially Responsible Machine Learning, NeurIPS},
year={2022},
url={https://openreview.net/forum?id=zaJsDuwwdlJ}
}

@inproceedings{lei-etal-2016-rationalizing,
    title = "Rationalizing Neural Predictions",
    author = "Lei, Tao  and
      Barzilay, Regina  and
      Jaakkola, Tommi",
    editor = "Su, Jian  and
      Duh, Kevin  and
      Carreras, Xavier",
    booktitle = "Proceedings of the 2016 Conference on Empirical Methods in Natural Language Processing",
    month = nov,
    year = "2016",
    address = "Austin, Texas",
    publisher = "Association for Computational Linguistics",
    url = "https://aclanthology.org/D16-1011",
    doi = "10.18653/v1/D16-1011",
    pages = "107--117",
}

@inproceedings{xia2019predicting,
  title={{Predicting Driver Attention in Critical Situations}},
  author={Xia, Ye and Zhang, Danqing and Kim, Jinkyu and Nakayama, Ken and Zipser, Karl and Whitney, David},
  booktitle={Computer Vision--ACCV 2018: 14th Asian Conference on Computer Vision, Perth, Australia, December 2--6, 2018, Revised Selected Papers, Part V 14},
  pages={658--674},
  year={2019},
  organization={Springer}
}

@article{atakishiyev2024explainable,
  title={{Explainable Artificial Intelligence for Autonomous Driving: A Comprehensive Overview and Field Guide for Future Research Directions}},
  author={Atakishiyev, Shahin and Salameh, Mohammad and Yao, Hengshuai and Goebel, Randy},
  journal={IEEE Access},
  volume={12},
  number={},
  pages={101603-101625},
  year={2024},
  publisher={IEEE}
}

@INPROCEEDINGS{atakishiyev2024incorporating,
  author={Atakishiyev, Shahin and Salameh, Mohammad and Goebel, Randy},
  booktitle={2024 IEEE Intelligent Vehicles Symposium (IV)}, 
  title={{Incorporating Explanations into Human-Machine Interfaces for Trust and Situation Awareness in Autonomous Vehicles}}, 
  pages={2948-2955},
  year={2024},
  }

@article{cheng2024videollama,
  title={{VideoLLaMA 2: Advancing Spatial-Temporal Modeling and Audio Understanding in Video-LLMs}},
  author={Cheng, Zesen and Leng, Sicong and Zhang, Hang and Xin, Yifei and Li, Xin and Chen, Guanzheng and Zhu, Yongxin and Zhang, Wenqi and Luo, Ziyang and Zhao, Deli and others},
  journal={arXiv preprint arXiv:2406.07476},
  year={2024}
}

@article{stepin2021survey,
  title={{A Survey of Contrastive and Counterfactual
Explanation Generation Methods for Explainable
Artificial Intelligence}},
  author={Stepin, Ilia and Alonso, Jose M and Catala, Alejandro and Pereira-Fari{\~n}a, Mart{\'\i}n},
  journal={IEEE Access},
  volume={9},
  pages={11974--12001},
  year={2021},
  publisher={IEEE}
}

@article{wei2022chain,
  title={{Chain-of-Thought Prompting Elicits Reasoning in Large Language Models}},
  author={Wei, Jason and Wang, Xuezhi and Schuurmans, Dale and Bosma, Maarten and Xia, Fei and Chi, Ed and Le, Quoc V and Zhou, Denny and others},
  journal={Advances in Neural Information Processing Systems},
  volume={35},
  pages={24824--24837},
  year={2022}
}

@article{arrieta2020explainable,
  title={{Explainable Artificial Intelligence (XAI): Concepts, taxonomies, opportunities and challenges toward responsible AI}},
  author={Arrieta, Alejandro Barredo and D{\'\i}az-Rodr{\'\i}guez, Natalia and Del Ser, Javier and Bennetot, Adrien and Tabik, Siham and Barbado, Alberto and Garc{\'\i}a, Salvador and Gil-L{\'o}pez, Sergio and Molina, Daniel and Benjamins, Richard and others},
  journal={Information Fusion},
  volume={58},
  pages={82--115},
  year={2020},
  publisher={Elsevier}
}

@article{elhage2021mathematical,
   title={A Mathematical Framework for Transformer Circuits},
   author={Elhage, Nelson and Nanda, Neel and Olsson, Catherine and Henighan, Tom and Joseph, Nicholas and Mann, Ben and Askell, Amanda and Bai, Yuntao and Chen, Anna and Conerly, Tom and DasSarma, Nova and Drain, Dawn and Ganguli, Deep and Hatfield-Dodds, Zac and Hernandez, Danny and Jones, Andy and Kernion, Jackson and Lovitt, Liane and Ndousse, Kamal and Amodei, Dario and Brown, Tom and Clark, Jack and Kaplan, Jared and McCandlish, Sam and Olah, Chris},
   year={2021},
   journal={Transformer Circuits Thread},
   note={https://transformer-circuits.pub/2021/framework/index.html}
}

@article{olsson2022context,
   title={In-context Learning and Induction Heads},
   author={Olsson, Catherine and Elhage, Nelson and Nanda, Neel and Joseph, Nicholas and DasSarma, Nova and Henighan, Tom and Mann, Ben and Askell, Amanda and Bai, Yuntao and Chen, Anna and Conerly, Tom and Drain, Dawn and Ganguli, Deep and Hatfield-Dodds, Zac and Hernandez, Danny and Johnston, Scott and Jones, Andy and Kernion, Jackson and Lovitt, Liane and Ndousse, Kamal and Amodei, Dario and Brown, Tom and Clark, Jack and Kaplan, Jared and McCandlish, Sam and Olah, Chris},
   year={2022},
   journal={Transformer Circuits Thread},
   note={https://transformer-circuits.pub/2022/in-context-learning-and-induction-heads/index.html}
}

@inproceedings{wang2023interpretability,
title={Interpretability in the Wild: a Circuit for Indirect Object Identification in {GPT}-2 Small},
author={Kevin Ro Wang and Alexandre Variengien and Arthur Conmy and Buck Shlegeris and Jacob Steinhardt},
booktitle={The Eleventh International Conference on Learning Representations },
year={2023},
url={https://openreview.net/forum?id=NpsVSN6o4ul}
}

@misc{bills2023language,
         title={Language models can explain neurons in language models},
         author={
            Bills, Steven and Cammarata, Nick and Mossing, Dan and Tillman, Henk and Gao, Leo and Goh, Gabriel and Sutskever, Ilya and Leike, Jan and Wu, Jeff and Saunders, William
         },
         year={2023},
         howpublished = {\url{https://openaipublic.blob.core.windows.net/neuron-explainer/paper/index.html}}
      }

@article{goldowsky2023localizing,
  title={Localizing model behavior with path patching},
  author={Goldowsky-Dill, Nicholas and MacLeod, Chris and Sato, Lucas and Arora, Aryaman},
  journal={arXiv preprint arXiv:2304.05969},
  year={2023}
}

@article{gurnee2023finding,
  title={Finding neurons in a haystack: Case studies with sparse probing},
  author={Gurnee, Wes and Nanda, Neel and Pauly, Matthew and Harvey, Katherine and Troitskii, Dmitrii and Bertsimas, Dimitris},
  journal={Transactions on Machine Learning Research},
  year={2023}
}

@article{wu2024interpretability,
  title={Interpretability at scale: Identifying causal mechanisms in alpaca},
  author={Wu, Zhengxuan and Geiger, Atticus and Icard, Thomas and Potts, Christopher and Goodman, Noah},
  journal={Advances in Neural Information Processing Systems},
  volume={36},
  year={2024}
}

@article{hase2024does,
  title={{Does Localization Inform Editing? Surprising
Differences in Causality-Based Localization vs.
Knowledge Editing in Language Models}},
  author={Hase, Peter and Bansal, Mohit and Kim, Been and Ghandeharioun, Asma},
  journal={Advances in Neural Information Processing Systems},
  volume={36},
  year={2024}
}

@article{li2024inference,
  title={{Inference-Time Intervention:
Eliciting Truthful Answers from a Language Model}},
  author={Li, Kenneth and Patel, Oam and Vi{\'e}gas, Fernanda and Pfister, Hanspeter and Wattenberg, Martin},
  journal={Advances in Neural Information Processing Systems},
  volume={36},
  year={2024}
}

@inproceedings{hou2023towards,
  title={Towards a Mechanistic Interpretation of Multi-Step Reasoning Capabilities of Language Models},
  author={Hou, Yifan and Li, Jiaoda and Fei, Yu and Stolfo, Alessandro and Zhou, Wangchunshu and Zeng, Guangtao and Bosselut, Antoine and Sachan, Mrinmaya},
  booktitle={Proceedings of the 2023 Conference on Empirical Methods in Natural Language Processing},
  pages={4902--4919},
  year={2023}
}

@article{conmy2023towards,
  title={Towards automated circuit discovery for mechanistic interpretability},
  author={Conmy, Arthur and Mavor-Parker, Augustine and Lynch, Aengus and Heimersheim, Stefan and Garriga-Alonso, Adri{\`a}},
  journal={Advances in Neural Information Processing Systems},
  volume={36},
  pages={16318--16352},
  year={2023}
}

@article{friedman2024learning,
  title={Learning transformer programs},
  author={Friedman, Dan and Wettig, Alexander and Chen, Danqi},
  journal={Advances in Neural Information Processing Systems},
  volume={36},
  year={2024}
}

@article{hanna2024does,
  title={How does GPT-2 compute greater-than?: Interpreting mathematical abilities in a pre-trained language model},
  author={Hanna, Michael and Liu, Ollie and Variengien, Alexandre},
  journal={Advances in Neural Information Processing Systems},
  volume={36},
  year={2024}
}

@article{zhang2023towards,
  title={Towards best practices of activation patching in language models: Metrics and methods},
  author={Zhang, Fred and Nanda, Neel},
  journal={International Conference on Learning Representations},
  year={2024}
}

@article{gould2023successor,
  title={Successor heads: Recurring, interpretable attention heads in the wild},
  author={Gould, Rhys and Ong, Euan and Ogden, George and Conmy, Arthur},
  journal={International Conference on Learning Representations},
  year={2024}
}

@InProceedings{pmlr-v235-lee24a,
  title = 	 {A Mechanistic Understanding of Alignment Algorithms: A Case Study on {DPO} and Toxicity},
  author =       {Lee, Andrew and Bai, Xiaoyan and Pres, Itamar and Wattenberg, Martin and Kummerfeld, Jonathan K. and Mihalcea, Rada},
  booktitle = 	 {Proceedings of the 41st International Conference on Machine Learning},
  pages = 	 {26361--26378},
  year = 	 {2024},
  series = 	 {Proceedings of Machine Learning Research},
  publisher =    {PMLR},
}

@inproceedings{
jain2024mechanistically,
title={Mechanistically analyzing the effects of fine-tuning on procedurally defined tasks},
author={Samyak Jain and Robert Kirk and Ekdeep Singh Lubana and Robert P. Dick and Hidenori Tanaka and Tim Rockt{\"a}schel and Edward Grefenstette and David Krueger},
booktitle={International Conference on Learning Representations},
year={2024}
}

@inproceedings{ren-etal-2024-identifying,
    title = "Identifying Semantic Induction Heads to Understand In-Context Learning",
    author = "Ren, Jie  and
      Guo, Qipeng  and
      Yan, Hang  and
      Liu, Dongrui  and
      Zhang, Quanshi  and
      Qiu, Xipeng  and
      Lin, Dahua",
    booktitle = "Findings of the Association for Computational Linguistics: ACL 2024",
    year = "2024",
    url = "https://aclanthology.org/2024.findings-acl.412/",
    pages = "6916--6932"    
}

@inproceedings{
prakash2024finetuning,
title={Fine-Tuning Enhances Existing Mechanisms: A Case Study on Entity Tracking},
author={Nikhil Prakash and Tamar Rott Shaham and Tal Haklay and Yonatan Belinkov and David Bau},
booktitle={International Conference on Learning Representations},
year={2024}
}

@inproceedings{lan2024towards,
  title={Towards Interpretable Sequence Continuation: Analyzing Shared Circuits in Large Language Models},
  author={Lan, Michael and Torr, Philip and Barez, Fazl},
  booktitle={Proceedings of the 2024 Conference on Empirical Methods in Natural Language Processing},
  pages={12576--12601},
  year={2024}
}

@inproceedings{
todd2024function,
title={Function Vectors in Large Language Models},
author={Eric Todd and Millicent Li and Arnab Sen Sharma and Aaron Mueller and Byron C Wallace and David Bau},
booktitle={International Conference on Learning Representations},
year={2024}
}

@inproceedings{
niu2024what,
title={What does the Knowledge Neuron Thesis Have to do with Knowledge?},
author={Jingcheng Niu and Andrew Liu and Zining Zhu and Gerald Penn},
booktitle={International Conference on Learning Representations},
year={2024}
}

@InProceedings{pmlr-v235-singh24c,
  title = 	 {What needs to go right for an induction head? {A} mechanistic study of in-context learning circuits and their formation},
  author =       {Singh, Aaditya K and Moskovitz, Ted and Hill, Felix and Chan, Stephanie C.Y. and Saxe, Andrew M},
  booktitle = 	 {Proceedings of the 41st International Conference on Machine Learning},
  pages = 	 {45637--45662},
  year = 	 {2024},
  series = 	 {Proceedings of Machine Learning Research},
  publisher =    {PMLR}
}

@inproceedings{
ghandeharioun2024patchscopes,
title={{Patchscopes: A Unifying Framework for Inspecting Hidden Representations of Language Models}},
author={Asma Ghandeharioun and Avi Caciularu and Adam Pearce and Lucas Dixon and Mor Geva},
booktitle={International Conference on Machine Learning},
year={2024}
}

@inproceedings{wang-etal-2024-mmlu,
    title = "{MMLU}-{SR}: A Benchmark for Stress-Testing Reasoning Capability of Large Language Models",
    author = "Wang, Wentian  and
      Jain, Sarthak  and
      Kantor, Paul  and
      Feldman, Jacob  and
      Gallos, Lazaros  and
      Wang, Hao",
    booktitle = "Proceedings of the 2nd GenBench Workshop on Generalisation (Benchmarking) in NLP",
    year = "2024",
    address = "Miami, Florida, USA",
    publisher = "Association for Computational Linguistics",
    pages = "69--85"
}

@article{phan2025humanity,
  title={{Humanity's Last Exam}},
  author={Phan, Long and Gatti, Alice and Han, Ziwen and Li, Nathaniel and Hu, Josephina and Zhang, Hugh and Shi, Sean and Choi, Michael and Agrawal, Anish and Chopra, Arnav and others},
  journal={arXiv preprint arXiv:2501.14249},
  year={2025}
}

@inproceedings{pan2023risk,
  title={{On the Risk of Misinformation Pollution with Large Language Models}},
  author={Pan, Yikang and Pan, Liangming and Chen, Wenhu and Nakov, Preslav and Kan, Min-Yen and Wang, William},
  booktitle={Findings of the Association for Computational Linguistics: EMNLP 2023},
  pages={1389--1403},
  year={2023}
}

@inproceedings{zhou2023synthetic,
  title={{Synthetic Lies: Understanding AI-Generated Misinformation and Evaluating Algorithmic and Human Solutions}},
  author={Zhou, Jiawei and Zhang, Yixuan and Luo, Qianni and Parker, Andrea G and De Choudhury, Munmun},
  booktitle={Proceedings of the 2023 CHI Conference on Human Factors in Computing Systems},
  pages={1--20},
  year={2023}
}

@inproceedings{sinha2019clutrr,
  title={{CLUTRR: A Diagnostic Benchmark for Inductive Reasoning from Text}},
  author={Sinha, Koustuv and Sodhani, Shagun and Dong, Jin and Pineau, Joelle and Hamilton, William L},
  booktitle={Proceedings of the 2019 Conference on Empirical Methods in Natural Language Processing and the 9th International Joint Conference on Natural Language Processing (EMNLP-IJCNLP)},
  pages={4506--4515},
  year={2019}
}

@inproceedings{
hahn2021teaching,
title={{Teaching Temporal Logics to Neural Networks}},
author={Christopher Hahn and Frederik Schmitt and Jens U. Kreber and Markus Norman Rabe and Bernd Finkbeiner},
booktitle={International Conference on Learning Representations},
year={2021}
}

@inproceedings{softreasoner_transformer,
author = {Clark, Peter and Tafjord, Oyvind and Richardson, Kyle},
title = {Transformers as soft reasoners over language},
year = {2020},
booktitle = {Proceedings of the Twenty-Ninth International Joint Conference on Artificial Intelligence},
location = {Yokohama, Yokohama, Japan}
}

@inproceedings{parmar2024logicbench,
  title={{LogicBench: Towards Systematic Evaluation of Logical Reasoning Ability of Large Language Models}},
  author={Parmar, Mihir and Patel, Nisarg and Varshney, Neeraj and Nakamura, Mutsumi and Luo, Man and Mashetty, Santosh and Mitra, Arindam and Baral, Chitta},
  booktitle={Proceedings of the 62nd Annual Meeting of the Association for Computational Linguistics (Volume 1: Long Papers)},
  pages={13679--13707},
  year={2024}
}

@inproceedings{koh2020concept,
  title={{Concept Bottleneck Models}},
  author={Koh, Pang Wei and Nguyen, Thao and Tang, Yew Siang and Mussmann, Stephen and Pierson, Emma and Kim, Been and Liang, Percy},
  booktitle={International Conference on Machine Learning},
  pages={5338--5348},
  year={2020},
  organization={PMLR}
}

@article{panousis2025coarse,
  title={{Coarse-to-Fine Concept Bottleneck Models}},
  author={Panousis, Konstantinos and Ienco, Dino and Marcos, Diego},
  journal={Advances in Neural Information Processing Systems},
  volume={37},
  pages={105171--105199},
  year={2025}
}

@article{kenny2024explainable,
  title={Explainable deep learning improves human mental models of self-driving cars},
  author={Kenny, Eoin M and Dharmavaram, Akshay and Lee, Sang Uk and Phan-Minh, Tung and Rajesh, Shreyas and Hu, Yunqing and Major, Laura and Tomov, Momchil S and Shah, Julie A},
  journal={arXiv preprint arXiv:2411.18714},
  year={2024}
}

@inproceedings{
yuksekgonul2023posthoc,
title={{Post-hoc Concept Bottleneck Models}},
author={Mert Yuksekgonul and Maggie Wang and James Zou},
booktitle={The Eleventh International Conference on Learning Representations },
year={2023}
}

@inproceedings{pang2024integrating,
  title={{Integrating Clinical Knowledge into Concept Bottleneck Models}},
  author={Pang, Winnie and Ke, Xueyi and Tsutsui, Satoshi and Wen, Bihan},
  booktitle={International Conference on Medical Image Computing and Computer-Assisted Intervention},
  pages={243--253},
  year={2024},
  organization={Springer}
}

@inproceedings{
sun2025concept,
title={{Concept Bottleneck Large Language Models}},
author={Chung-En Sun and Tuomas Oikarinen and Berk Ustun and Tsui-Wei Weng},
booktitle={The Thirteenth International Conference on Learning Representations},
year={2025}
}

@article{rudin2019stop,
  title={Stop explaining black box machine learning models for high stakes decisions and use interpretable models instead},
  author={Rudin, Cynthia},
  journal={Nature Machine Intelligence},
  volume={1},
  number={5},
  pages={206--215},
  year={2019},
  publisher={Nature Publishing Group UK London}
}

@article{kenny2024regulation,
  title={{Regulation of Language Models With Interpretability Will Likely Result In A Performance Trade-Off}},
  author={Kenny, Eoin M and Shah, Julie A},
  journal={arXiv preprint arXiv:2412.12169},
  year={2024}
}

@inproceedings{huang2022large,
  title={Are Large Pre-Trained Language Models Leaking Your Personal Information?},
  author={Huang, Jie and Shao, Hanyin and Chang, Kevin Chen-Chuan},
  booktitle={Findings of the Association for Computational Linguistics: EMNLP 2022},
  pages={2038--2047},
  year={2022}
}

@article{kim2024propile,
  title={{ProPILE: Probing Privacy Leakage in Large Language Models}},
  author={Kim, Siwon and Yun, Sangdoo and Lee, Hwaran and Gubri, Martin and Yoon, Sungroh and Oh, Seong Joon},
  journal={Advances in Neural Information Processing Systems},
  volume={36},
  year={2024}
}

@article{han2024medical,
  title={Medical large language models are susceptible to targeted misinformation attacks},
  author={Han, Tianyu and Nebelung, Sven and Khader, Firas and Wang, Tianci and M{\"u}ller-Franzes, Gustav and Kuhl, Christiane and F{\"o}rsch, Sebastian and Kleesiek, Jens and Haarburger, Christoph and Bressem, Keno K and others},
  journal={NPJ digital medicine},
  volume={7},
  number={1},
  pages={288},
  year={2024},
  publisher={Nature Publishing Group UK London}
}

@article{asimov1950runaround,
  title={{Runaround. I, robot}},
  author={Asimov, Isaac},
  journal={New York: Doubleday},
  year={1950}
}

@article{moor2006nature,
  title={{The Nature, Importance, and Difficulty of Machine Ethics}},
  author={Moor, James H},
  journal={IEEE Intelligent Systems},
  volume={21},
  number={4},
  pages={18--21},
  year={2006},
  publisher={IEEE}
}

@article{barman2024beyond,
  title={{Beyond transparency and explainability: on the need for adequate and contextualized user guidelines for LLM use}},
  author={Barman, Kristian Gonzalez and Wood, Nathan and Pawlowski, Pawel},
  journal={Ethics and Information Technology},
  volume={26},
  number={3},
  pages={47},
  year={2024},
  publisher={Springer}
}

@article{langer2021we,
  title={{What do we want from Explainable Artificial Intelligence (XAI)?--A stakeholder perspective on XAI and a conceptual model guiding interdisciplinary XAI research}},
  author={Langer, Markus and Oster, Daniel and Speith, Timo and Hermanns, Holger and K{\"a}stner, Lena and Schmidt, Eva and Sesing, Andreas and Baum, Kevin},
  journal={Artificial Intelligence},
  volume={296},
  pages={103473},
  year={2021},
  publisher={Elsevier}
}

@incollection{casonato2024ai,
  title={{AI Regulation in Europe: Exploring the Artificial Intelligence Act}},
  author={Casonato, Carlo and Olivato, Giulia},
  booktitle={Digital Environments and Human Relations},
  pages={87--111},
  year={2024},
  publisher={Springer}
}

@inproceedings{huang2024trustllm,
  title={{Position: TrustLLM: Trustworthiness in Large Language Models}},
  author={Huang, Yue and Sun, Lichao and Wang, Haoran and Wu, Siyuan and Zhang, Qihui and Li, Yuan and Gao, Chujie and Huang, Yixin and Lyu, Wenhan and Zhang, Yixuan and others},
  booktitle={International Conference on Machine Learning},
  pages={20166--20270},
  year={2024}
}

@article{liao2024ai,
  title={{AI Transparency in the Age of LLMs: A Human-Centered Research Roadmap}},
  author={Liao, Q Vera and Vaughan, Jennifer Wortman},
  journal={Harvard Data Science Review},
  number={Special Issue 5},
  year={2024},
  publisher={The MIT Press}
}

@article{hu2024unveiling,
  title={{Unveiling LLM Evaluation Focused on Metrics: Challenges and Solutions}},
  author={Hu, Taojun and Zhou, Xiao-Hua},
  journal={arXiv preprint arXiv:2404.09135},
  year={2024}
}

@inproceedings{lin2004rouge,
  title={{ROUGE: A Package for Automatic Evaluation of Summaries}},
  author={Lin, Chin-Yew},
  booktitle={Text summarization branches out},
  pages={74--81},
  year={2004}
}

@inproceedings{
wang2021adversarial,
title={{Adversarial {GLUE}: A Multi-Task Benchmark for Robustness Evaluation of Language Models}},
author={Boxin Wang and Chejian Xu and Shuohang Wang and Zhe Gan and Yu Cheng and Jianfeng Gao and Ahmed Hassan Awadallah and Bo Li},
booktitle={Thirty-fifth Conference on Neural Information Processing Systems Datasets and Benchmarks Track (Round 2)},
year={2021}
}

@inproceedings{zhu2023promptrobust,
  title={Promptrobust: Towards evaluating the robustness of large language models on adversarial prompts},
  author={Zhu, Kaijie and Wang, Jindong and Zhou, Jiaheng and Wang, Zichen and Chen, Hao and Wang, Yidong and Yang, Linyi and Ye, Wei and Zhang, Yue and Gong, Neil and others},
  booktitle={Proceedings of the 1st ACM Workshop on Large AI Systems and Models with Privacy and Safety Analysis},
  pages={57--68},
  year={2023}
}

@inproceedings{chen2023llm,
  title={LLM-EVAL: Unified Multi-Dimensional Automatic Evaluation for Open-Domain Conversations with Large Language Models},
  author={Chen, Yen-Ting Lin Yun-Nung},
  booktitle={The 5th Workshop on NLP for Conversational AI},
  pages={47},
  year={2023}
}

@inproceedings{liu2023g,
  title={{G-Eval: NLG Evaluation using GPT-4 with Better Human Alignment}},
  author={Liu, Yang and Iter, Dan and Xu, Yichong and Wang, Shuohang and Xu, Ruochen and Zhu, Chenguang},
  booktitle={Proceedings of the 2023 Conference on Empirical Methods in Natural Language Processing},
  pages={2511--2522},
  year={2023}
}

@inproceedings{yang2024rem,
  title={{REM: A Ranking-Based Automatic Evaluation Method for LLMs}},
  author={Yang, Jintao and Tan, Yushan and Hu, Wenpeng and Yang, Zonghao and Zhou, Xian and Luo, Zhunchen and Luo, Wei},
  booktitle={International Conference on Artificial Neural Networks},
  pages={371--385},
  year={2024},
  organization={Springer}
}

@inproceedings{
wang2024pandalm,
title={{Panda{LM}: An Automatic Evaluation Benchmark for {LLM} Instruction Tuning Optimization}},
author={Yidong Wang and Zhuohao Yu and Wenjin Yao and Zhengran Zeng and Linyi Yang and Cunxiang Wang and Hao Chen and Chaoya Jiang and Rui Xie and Jindong Wang and Xing Xie and Wei Ye and Shikun Zhang and Yue Zhang},
booktitle={The Twelfth International Conference on Learning Representations},
year={2024}
}

@article{
liang2023holistic,
title={Holistic Evaluation of Language Models},
author={Percy Liang and Rishi Bommasani and Tony Lee and Dimitris Tsipras and Dilara Soylu and Michihiro Yasunaga and Yian Zhang and Deepak Narayanan and Yuhuai Wu and Ananya Kumar and Benjamin Newman and Binhang Yuan and Bobby Yan and Ce Zhang and Christian Alexander Cosgrove and Christopher D Manning and Christopher Re and Diana Acosta-Navas and Drew Arad Hudson and Eric Zelikman and Esin Durmus and Faisal Ladhak and Frieda Rong and Hongyu Ren and Huaxiu Yao and Jue WANG and Keshav Santhanam and Laurel Orr and Lucia Zheng and Mert Yuksekgonul and Mirac Suzgun and Nathan Kim and Neel Guha and Niladri S. Chatterji and Omar Khattab and Peter Henderson and Qian Huang and Ryan Andrew Chi and Sang Michael Xie and Shibani Santurkar and Surya Ganguli and Tatsunori Hashimoto and Thomas Icard and Tianyi Zhang and Vishrav Chaudhary and William Wang and Xuechen Li and Yifan Mai and Yuhui Zhang and Yuta Koreeda},
journal={Transactions on Machine Learning Research},
issn={2835-8856},
year={2023}
}

@inproceedings{lewis2020bart,
  title={{BART: Denoising Sequence-to-Sequence Pre-training for Natural Language Generation, Translation, and Comprehension}},
  author={Lewis, Mike and Liu, Yinhan and Goyal, Naman and Ghazvininejad, Marjan and Mohamed, Abdelrahman and Levy, Omer and Stoyanov, Veselin and Zettlemoyer, Luke},
  booktitle={Proceedings of the 58th Annual Meeting of the Association for Computational Linguistics},
  pages={7871--7880},
  year={2020}
}

@inproceedings{zhong2024agieval,
  title={{AGIEval: A Human-Centric Benchmark for Evaluating Foundation Models}},
  author={Zhong, Wanjun and Cui, Ruixiang and Guo, Yiduo and Liang, Yaobo and Lu, Shuai and Wang, Yanlin and Saied, Amin and Chen, Weizhu and Duan, Nan},
  booktitle={Findings of the Association for Computational Linguistics: NAACL 2024},
  pages={2299--2314},
  year={2024}
}

@inproceedings{
ning2025pico,
title={{Pi{CO}: Peer Review in {LLM}s based on Consistency Optimization}},
author={Kun-Peng Ning and Shuo Yang and Yuyang Liu and Jia-Yu Yao and Zhenhui Liu and Yonghong Tian and Yibing Song and Li Yuan},
booktitle={The Thirteenth International Conference on Learning Representations},
year={2025}
}

@inproceedings{zhou2024alignment,
  title={{How Alignment and Jailbreak Work: Explain LLM Safety through Intermediate Hidden States}},
  author={Zhou, Zhenhong and Yu, Haiyang and Zhang, Xinghua and Xu, Rongwu and Huang, Fei and Li, Yongbin},
  booktitle={Findings of the Association for Computational Linguistics: EMNLP 2024},
  pages={2461--2488},
  year={2024}
}

@inproceedings{ferrando2024information,
  title={{Information Flow Routes: Automatically Interpreting Language Models at Scale}},
  author={Ferrando, Javier and Voita, Elena},
  booktitle={Proceedings of the 2024 Conference on Empirical Methods in Natural Language Processing},
  pages={17432--17445},
  year={2024}
}

@article{he2025learning,
  title={Learning to grok: Emergence of in-context learning and skill composition in modular arithmetic tasks},
  author={He, Tianyu and Doshi, Darshil and Das, Aritra and Gromov, Andrey},
  journal={Advances in Neural Information Processing Systems},
  volume={37},
  pages={13244--13273},
  year={2024}
}

@incollection{polanyi2009tacit,
  title={{The Tacit Dimension}},
  author={Polanyi, Michael},
  booktitle={Knowledge in organisations},
  pages={135--146},
  year={2009},
  publisher={Routledge}
}

@article{gabbatore2023silent,
  title={{Silent Finns and talkative Italians? An investigation of communicative differences and similarities as perceived by parents in typically developing children}},
  author={Gabbatore, Ilaria and Dindar, Katja and Pirinen, Veera and V{\"a}h{\"a}nikkil{\"a}, Hannu and M{\"a}mmel{\"a}, Laura and Kotila, Aija and Bosco, Francesca M and Leinonen, Eeva and Loukusa, Soile},
  journal={First Language},
  volume={43},
  number={3},
  pages={313--335},
  year={2023},
  publisher={SAGE Publications Sage UK: London, England}
}

@book{armstrong1973belief,
  title={{Belief, Truth and Knowledge
}},
  author={Armstrong, David Malet},
  year={1973},
  publisher={Cambridge University Press}
}

@book{price2002belief,
  title={Belief},
  author={Price, Henry Habberley},
  year={2002},
  publisher={Routledge}
}

@article{zhuang2021egg,
  title={{Egg and cholesterol consumption and mortality from cardiovascular and different causes in the United States: a population-based cohort study}},
  author={Zhuang, Pan and Wu, Fei and Mao, Lei and Zhu, Fanghuan and Zhang, Yiju and Chen, Xiaoqian and Jiao, Jingjing and Zhang, Yu},
  journal={PLoS Medicine},
  year={2021},
  publisher={Public Library of Science}

}

@article{rai2024practical,
  title={{A Practical Review of Mechanistic Interpretability for Transformer-Based Language Models}},
  author={Rai, Daking and Zhou, Yilun and Feng, Shi and Saparov, Abulhair and Yao, Ziyu},
  journal={arXiv preprint arXiv:2407.02646},
  year={2024}
}

@article{longo2024explainable,
  title={Explainable Artificial Intelligence (XAI) 2.0: A manifesto of open challenges and interdisciplinary research directions},
  author={Longo, Luca and Brcic, Mario and Cabitza, Federico and Choi, Jaesik and Confalonieri, Roberto and Del Ser, Javier and Guidotti, Riccardo and Hayashi, Yoichi and Herrera, Francisco and Holzinger, Andreas and others},
  journal={Information Fusion},
  volume={106},
  pages={102301},
  year={2024},
  publisher={Elsevier}
}

@inproceedings{gantla2025exploring,
  title={{Exploring Mechanistic Interpretability in Large Language Models: Challenges, Approaches, and Insights}},
  author={Gantla, Sandeep Reddy},
  booktitle={2025 International Conference on Data Science, Agents \& Artificial Intelligence (ICDSAAI)},
  pages={1--8},
  year={2025},
  organization={IEEE}
}

@article{chandna2025dissecting,
  title={{Dissecting Bias in LLMs: A Mechanistic Interpretability Perspective}},
  author={Chandna, Bhavik and Bashir, Zubair and Sen, Procheta},
  journal={arXiv preprint arXiv:2506.05166},
  year={2025}
}

@phdthesis{lad2024mechanistic,
  title={{Mechanistic Interpretability for Progress Towards Quantitative AI Safety}},
  author={Lad, Vedang K},
  year={2024},
  school={Massachusetts Institute of Technology}
}

@inproceedings{tanneru2024quantifying,
  title={Quantifying uncertainty in natural language explanations of large language models},
  author={Tanneru, Sree Harsha and Agarwal, Chirag and Lakkaraju, Himabindu},
  booktitle={International Conference on Artificial Intelligence and Statistics},
  pages={1072--1080},
  year={2024},
  organization={PMLR}
}

@inproceedings{lyu2023faithful,
  title={{Faithful Chain-of-Thought Reasoning}},
  author={Lyu, Qing and Havaldar, Shreya and Stein, Adam and Zhang, Li and Rao, Delip and Wong, Eric and Apidianaki, Marianna and Callison-Burch, Chris},
  booktitle={The 13th International Joint Conference on Natural Language Processing and the 3rd Conference of the Asia-Pacific Chapter of the Association for Computational Linguistics (IJCNLP-AACL 2023)},
  year={2023}
}

@article{turpin2023language,
  title={{Language Models Don’t Always Say What They Think:
Unfaithful Explanations in Chain-of-Thought Prompting}},
  author={Turpin, Miles and Michael, Julian and Perez, Ethan and Bowman, Samuel},
  journal={Advances in Neural Information Processing Systems},
  volume={36},
  pages={74952--74965},
  year={2023}
}

@article{arcuschin2025chain,
  title={Chain-of-thought reasoning in the wild is not always faithful},
  author={Arcuschin, Iv{\'a}n and Janiak, Jett and Krzyzanowski, Robert and Rajamanoharan, Senthooran and Nanda, Neel and Conmy, Arthur},
  journal={arXiv preprint arXiv:2503.08679},
  year={2025}
}

@article{Dai_Kim_Sutton_Mitchell_Goebel_Baumgart_2025, title={{Comparative analysis of natural language processing methodologies for classifying computed tomography enterography reports in Crohn’s disease patients}}, volume={8}, DOI={10.1038/s41746-025-01729-5}, number={1}, journal={npj Digital Medicine}, author={Dai, Jiayi and Kim, Mi-Young and Sutton, Reed T. and Mitchell, J. Ross and Goebel, Randolph and Baumgart, Daniel C.}, year={2025}}

@misc{barez-chain-2025,
  author = {Barez, Fazl and Wu, Tung-Yu and Arcuschin, Iván and Lan, Michael and Wang, Vincent and Siegel, Noah and Collignon, Nicolas and Neo, Clement and Lee, Isabelle and Paren, Alasdair and Bibi, Adel and Trager, Robert and Fornasiere, Damiano and Yan, John and Elazar, Yanai and Bengio, Yoshua},
  title = {Chain-of-Thought Is Not Explainability},
  publisher = {alphaXiv},
  year = {2025}
}

@article{yuan2024rag,
  title={{RAG-Driver: Generalisable Driving Explanations with
Retrieval-Augmented In-Context Learning in Multi-Modal
Large Language Model}},
  author={Yuan, Jianhao and Sun, Shuyang and Omeiza, Daniel and Zhao, Bo and Newman, Paul and Kunze, Lars and Gadd, Matthew},
  journal={Robotics: Science and Systems},
  year={2024}
}

@inproceedings{li2025g,
  title={{G-Refer: Graph Retrieval-Augmented Large Language Model for Explainable Recommendation}},
  author={Li, Yuhan and Zhang, Xinni and Luo, Linhao and Chang, Heng and Ren, Yuxiang and King, Irwin and Li, Jia},
  booktitle={Proceedings of the ACM on Web Conference 2025},
  pages={240--251},
  year={2025}
}

@inproceedings{wu2020perturbed,
  title={{Perturbed Masking: Parameter-free Probing for Analyzing and Interpreting BERT}},
  author={Wu, Zhiyong and Chen, Yun and Kao, Ben and Liu, Qun},
  booktitle={Proceedings of the 58th Annual Meeting of the Association for Computational Linguistics},
  pages={4166--4176},
  year={2020}
}

@inproceedings{mohebbi-etal-2021-exploring,
    title = "Exploring the Role of {BERT} Token Representations to Explain Sentence Probing Results",
    author = "Mohebbi, Hosein  and
      Modarressi, Ali  and
      Pilehvar, Mohammad Taher",

    booktitle = "Proceedings of the 2021 Conference on Empirical Methods in Natural Language Processing",
    year = "2021",
    address = "Online and Punta Cana, Dominican Republic",
    publisher = "Association for Computational Linguistics",
    pages = "792--806",
}

@article{enguehard2023sequential,
  title={{Sequential Integrated Gradients: a simple but effective method for explaining language models}},
  author={Enguehard, Joseph},
  journal={ACL Findings},
  year={2023}
}

@inproceedings{kokalj2021bert,
  title={{BERT meets Shapley: Extending SHAP Explanations to
Transformer-based Classifiers}},
  author={Kokalj, Enja and {\v{S}}krlj, Bla{\v{z}} and Lavra{\v{c}}, Nada and Pollak, Senja and Robnik-{\v{S}}ikonja, Marko},
  booktitle={Proceedings of the EACL hackashop on news media content analysis and automated report generation},
  pages={16--21},
  year={2021}
}

@article{wu2021explaining,
  title={On explaining your explanations of bert: An empirical study with sequence classification},
  author={Wu, Zhengxuan and Ong, Desmond C},
  journal={arXiv preprint arXiv:2101.00196},
  year={2021}
}

@article{bai2023qwen,
  title={Qwen Technical Report},
  author={Bai, Jinze and Bai, Shuai and Chu, Yunfei and Cui, Zeyu and Dang, Kai and Deng, Xiaodong and Fan, Yang and Ge, Wenbin and Han, Yu and Huang, Fei and others},
  journal={arXiv preprint arXiv:2309.16609},
  year={2023}
}

@article{comanici2025gemini,
  title={Gemini 2.5: Pushing the frontier with advanced reasoning, multimodality, long context, and next generation agentic capabilities},
  author={Comanici, Gheorghe and Bieber, Eric and Schaekermann, Mike and Pasupat, Ice and Sachdeva, Noveen and Dhillon, Inderjit and Blistein, Marcel and Ram, Ori and Zhang, Dan and Rosen, Evan and others},
  journal={arXiv preprint arXiv:2507.06261},
  year={2025}
}

@article{team2024qwen2,
  title={{Qwen2 Technical Report}},
  author={{Qwen Team}},
  journal={arXiv preprint arXiv:2407.10671},
  year={2024}
}

@inproceedings{volkov2024local,
  title={{Local Explanations for Large Language Models: a Brief Review of Methods}},
  author={Volkov, Egor N and Averkin, Alexey N},
  booktitle={2024 XXVII International Conference on Soft Computing and Measurements (SCM)},
  pages={189--192},
  year={2024},
  organization={IEEE}
}

@article{gabriel2020artificial,
  title={{Artificial Intelligence, Values, and Alignment}},
  author={Gabriel, Iason},
  journal={Minds and Machines},
  volume={30},
  number={3},
  pages={411--437},
  year={2020},
  publisher={Springer}
}

@article{ferrando2024primer,
  title={{A Primer on the Inner Workings of Transformer-based Language Models}},
  author={Ferrando, Javier and Sarti, Gabriele and Bisazza, Arianna and Costa-Juss{\`a}, Marta R},
  journal={arXiv preprint arXiv:2405.00208},
  year={2024}
}

@article{regulation2016regulation,
  title={{Regulation (EU) 2016/679 of the European Parliament and of the Council}},
  author={Regulation, Protection},
  journal={Regulation (EU)},
  volume={679},
  number={2016},
  pages={10--13},
  year={2016}
}

@article{gu2024survey,
  title={{A Survey on LLM-as-a-Judge}},
  author={Gu, Jiawei and Jiang, Xuhui and Shi, Zhichao and Tan, Hexiang and Zhai, Xuehao and Xu, Chengjin and Li, Wei and Shen, Yinghan and Ma, Shengjie and Liu, Honghao and others},
  journal={arXiv preprint arXiv:2411.15594},
  year={2024}
}

@inproceedings{dziri2022origin,
  title={{On the Origin of Hallucinations in Conversational Models: Is it the Datasets or the Models?}},
  author={Dziri, Nouha and Milton, Sivan and Yu, Mo and Zaiane, Osmar R and Reddy, Siva},
  booktitle={Proceedings of the 2022 Conference of the North American Chapter of the Association for Computational Linguistics: Human Language Technologies},
  pages={5271--5285},
  year={2022}
}

@article{bi2024deepseek,
  title={{DeepSeek LLM: Scaling Open-Source Language Models with Longtermism}},
  author={Bi, Xiao and Chen, Deli and Chen, Guanting and Chen, Shanhuang and Dai, Damai and Deng, Chengqi and Ding, Honghui and Dong, Kai and Du, Qiushi and Fu, Zhe and others},
  journal={arXiv preprint arXiv:2401.02954},
  year={2024}
}

@article{guo2025deepseek,
  title={{DeepSeek-R1: Incentivizing Reasoning Capability in LLMs via Reinforcement Learning}},
  author={Guo, Daya and Yang, Dejian and Zhang, Haowei and Song, Junxiao and Zhang, Ruoyu and Xu, Runxin and Zhu, Qihao and Ma, Shirong and Wang, Peiyi and Bi, Xiao and others},
  journal={arXiv preprint arXiv:2501.12948},
  year={2025}
}

@article{liu2024v3deepseek,
  title={{DeepSeek-V3 Technical Report}},
  author={Liu, Aixin and Feng, Bei and Xue, Bing and Wang, Bingxuan and Wu, Bochao and Lu, Chengda and Zhao, Chenggang and Deng, Chengqi and Zhang, Chenyu and Ruan, Chong and others},
  journal={arXiv preprint arXiv:2412.19437},
  year={2024}
}

@article{liu2024deepseek,
  title={{DeepSeek-V2: A Strong, Economical, and Efficient Mixture-of-Experts Language Model}},
  author={Liu, Aixin and Feng, Bei and Wang, Bin and Wang, Bingxuan and Liu, Bo and Zhao, Chenggang and Dengr, Chengqi and Ruan, Chong and Dai, Damai and Guo, Daya and others},
  journal={arXiv preprint arXiv:2405.04434},
  year={2024}
}

@article{team2025kimik2,
  title={{Kimi K2: Open Agentic Intelligence}},
  author={Team, Kimi and Bai, Yifan and Bao, Yiping and Chen, Guanduo and Chen, Jiahao and Chen, Ningxin and Chen, Ruijue and Chen, Yanru and Chen, Yuankun and Chen, Yutian and others},
  journal={arXiv preprint arXiv:2507.20534},
  year={2025}
}

@article{team2025kimi1.5,
  title={{Kimi k1.5: Scaling Reinforcement Learning with LLMs}},
  author={Team, Kimi and Du, Angang and Gao, Bofei and Xing, Bowei and Jiang, Changjiu and Chen, Cheng and Li, Cheng and Xiao, Chenjun and Du, Chenzhuang and Liao, Chonghua and others},
  journal={arXiv preprint arXiv:2501.12599},
  year={2025}
}

@misc{gpt5systemcard2025,
  title        = {{GPT-5 System Card}},
  author       = {OpenAI},
  year         = {2025},
  month        = Aug,
  url          = {https://cdn.openai.com/pdf/8124a3ce-ab78-4f06-96eb-49ea29ffb52f/gpt5-system-card-aug7.pdf}
}

@article{fanous2025syceval,
  title={{SycEval: Evaluating LLM Sycophancy}},
  author={Fanous, Aaron and Goldberg, Jacob and Agarwal, Ank A and Lin, Joanna and Zhou, Anson and Daneshjou, Roxana and Koyejo, Sanmi},
  journal={arXiv preprint arXiv:2502.08177},
  year={2025}
}

@inproceedings{wang2024navigating,
  title={{Navigating the Risks: A Review of Safety Issues in Large Language Models}},
  author={Wang, Haiyang and Li, Yihao and Wang, Yue and Liu, Pan and Li, Pengxiao},
  booktitle={2024 IEEE 24th International Conference on Software Quality, Reliability, and Security Companion (QRS-C)},
  pages={74--83},
  year={2024},
  organization={IEEE}
}

@article{cheng2024interactive,
  title={{Interactive Analysis of LLMs using Meaningful Counterfactuals}},
  author={Cheng, Furui and Zouhar, Vil{\'e}m and Chan, Robin Shing Moon and F{\"u}rst, Daniel and Strobelt, Hendrik and El-Assady, Mennatallah},
  journal={arXiv preprint arXiv:2405.00708},
  year={2024}
}

@article{paulus2024advprompter,
  title={{AdvPrompter: Fast Adaptive Adversarial Prompting for LLMs}},
  author={Paulus, Anselm and Zharmagambetov, Arman and Guo, Chuan and Amos, Brandon and Tian, Yuandong},
  journal={arXiv preprint arXiv:2404.16873},
  year={2024}
}

@inproceedings{matton2025walk,
title={{Walk the Talk? Measuring the Faithfulness of Large Language Model Explanations}},
author={Katie Matton and Robert Ness and John Guttag and Emre Kiciman},
booktitle={The Thirteenth International Conference on Learning Representations},
year={2025}
}

@article{yang2025qwen3,
  title={{Qwen3 Technical Report}},
  author={Yang, An and Li, Anfeng and Yang, Baosong and Zhang, Beichen and Hui, Binyuan and Zheng, Bo and Yu, Bowen and Gao, Chang and Huang, Chengen and Lv, Chenxu and others},
  journal={arXiv preprint arXiv:2505.09388},
  year={2025}
}

@misc{anthropic2025claude45,
  author       = {Anthropic},
  title        = {{Introducing Claude Sonnet 4.5}},
  year         = {2025},
  month        = {September},
  url          = {https://www.anthropic.com/news/claude-sonnet-4-5}
}

@article{huang2025survey,
  title={{A Survey on Hallucination in Large Language Models: Principles, Taxonomy, Challenges, and Open Questions}},
  author={Huang, Lei and Yu, Weijiang and Ma, Weitao and Zhong, Weihong and Feng, Zhangyin and Wang, Haotian and Chen, Qianglong and Peng, Weihua and Feng, Xiaocheng and Qin, Bing and others},
  journal={ACM Transactions on Information Systems},
  volume={43},
  number={2},
  pages={1--55},
  year={2025},
  publisher={ACM New York, NY}
}

@inproceedings{Lan2020ALBERT,
title={{ALBERT: A Lite BERT for Self-supervised Learning of Language Representations}},
author={Zhenzhong Lan and Mingda Chen and Sebastian Goodman and Kevin Gimpel and Piyush Sharma and Radu Soricut},
booktitle={International Conference on Learning Representations},
year={2020}
}

@article{liu2019roberta,
  title={{RoBERTa: A Robustly Optimized BERT Pretraining Approach}},
  author={Liu, Yinhan and Ott, Myle and Goyal, Naman and Du, Jingfei and Joshi, Mandar and Chen, Danqi and Levy, Omer and Lewis, Mike and Zettlemoyer, Luke and Stoyanov, Veselin},
  journal={arXiv preprint arXiv:1907.11692},
  year={2019}
}

@article{hurst2024gpt,
  title={{GPT-4o System Card}},
  author={Hurst, Aaron and Lerer, Adam and Goucher, Adam P and Perelman, Adam and Ramesh, Aditya and Clark, Aidan and Ostrow, AJ and Welihinda, Akila and Hayes, Alan and Radford, Alec and others},
  journal={arXiv preprint arXiv:2410.21276},
  year={2024}
}

@article{sharkey2025open,
  title={{Open Problems in Mechanistic Interpretability
}},
  author={Sharkey, Lee and Chughtai, Bilal and Batson, Joshua and Lindsey, Jack and Wu, Jeff and Bushnaq, Lucius and Goldowsky-Dill, Nicholas and Heimersheim, Stefan and Ortega, Alejandro and Bloom, Joseph and others},
  journal={arXiv preprint arXiv:2501.16496},
  year={2025}
}

@article{verma2025teaching,
  title={{Teaching LLMs to Plan: Logical Chain-of-Thought Instruction Tuning for Symbolic Planning}},
  author={Verma, Pulkit and La, Ngoc and Favier, Anthony and Mishra, Swaroop and Shah, Julie A},
  journal={arXiv preprint arXiv:2509.13351},
  year={2025}
}

@article{hossain2025explainable,
  title={{Explainable AI for Medical Data: Current Methods, Limitations, and Future Directions}},
  author={Hossain, Md Imran and Zamzmi, Ghada and Mouton, Peter R and Salekin, Md Sirajus and Sun, Yu and Goldgof, Dmitry},
  journal={ACM Computing Surveys},
  volume={57},
  number={6},
  pages={1--46},
  year={2025},
  publisher={ACM New York, NY}
}

@article{lundberg2017unified,
  title={{A Unified Approach to Interpreting Model Predictions
}},
  author={Lundberg, Scott M and Lee, Su-In},
  journal={Advances in Neural Information Processing Systems},
  volume={30},
  year={2017}
}

@article{scherlis2022polysemanticity,
  title={{Polysemanticity and Capacity in Neural Networks
}},
  author={Scherlis, Adam and Sachan, Kshitij and Jermyn, Adam S and Benton, Joe and Shlegeris, Buck},
  journal={arXiv preprint arXiv:2210.01892},
  year={2022}
}

@inproceedings{pal2023med,
  title={{Med-HALT: Medical Domain Hallucination Test for Large Language Models}},
  author={Pal, Ankit and Umapathi, Logesh Kumar and Sankarasubbu, Malaikannan},
  booktitle={Proceedings of the 27th Conference on Computational Natural Language Learning (CoNLL)},
  pages={314--334},
  year={2023}
}

@article{prina2015community,
  title={Community-acquired pneumonia},
  author={Prina, Elena and Ranzani, Otavio T and Torres, Antoni},
  journal={The Lancet},
  volume={386},
  number={9998},
  pages={1097--1108},
  year={2015},
  publisher={Elsevier}
}

@inproceedings{shen2025stressprompt,
  title={{StressPrompt: Does Stress Impact Large Language Models and Human Performance Similarly?}},
  author={Shen, Guobin and Zhao, Dongcheng and Bao, Aorigele and He, Xiang and Dong, Yiting and Zeng, Yi},
  booktitle={Proceedings of the AAAI Conference on Artificial Intelligence},
  volume={39},
  number={1},
  pages={711--719},
  year={2025}
}

@misc{xai2024grok,
  author       = {xAI},
  title        = {{Open Release of Grok-1}},
  year         = {2024},
  url          = {https://x.ai/news/grok-os},
  howpublished = {\url{https://x.ai/news/grok-os}}
}

@misc{xai2025grok4,
  author       = {xAI},
  title        = {Grok 4},
  year         = {2025},
  url          = {https://x.ai/news/grok-4},
  howpublished = {\url{https://x.ai/news/grok-4}}
}

@misc{xai2024grok2,
  author       = {xAI},
  title        = {Grok-2 Beta Release},
  year         = {2024},
  month        = {August},
  url          = {https://x.ai/news/grok-2},
  note         = {Accessed October 12, 2025},
  howpublished = {\url{https://x.ai/news/grok-2}}
}

@article{bengio2025singapore,
  title={{The Singapore Consensus on Global AI Safety Research Priorities}},
  author={Bengio, Yoshua and Maharaj, Tegan and Ong, Luke and Russell, Stuart and Song, Dawn and Tegmark, Max and Xue, Lan and Zhang, Ya-Qin and Casper, Stephen and Lee, Wan Sie and others},
  journal={arXiv preprint arXiv:2506.20702},
  year={2025}
}

@misc{eu2024aiact,
  author       = {{European Parliament and Council of the European Union}},
  title        = {{Artificial Intelligence Act: European Parliament legislative resolution of 13 March 2024 on the proposal for a regulation of the European Parliament and of the Council on laying down harmonised rules on Artificial Intelligence (Artificial Intelligence Act) and amending certain Union Legislative Acts (COM(2021)0206 – C9-0146/2021 – 2021/0106(COD))}},
  year         = {2024},
  month        = {March},
  howpublished = {\url{https://artificialintelligenceact.eu/wp-content/uploads/2024/04/TA-9-2024-0138_EN.pdf}}
}

@inproceedings{malmqvist2025sycophancy,
  title={{Sycophancy in Large Language Models: Causes and Mitigations}},
  author={Malmqvist, Lars},
  booktitle={Intelligent Computing-Proceedings of the Computing Conference},
  pages={61--74},
  year={2025},
  organization={Springer}
}

@inproceedings{
li2025causally,
title={Causally Motivated Sycophancy Mitigation for Large Language Models},
author={Haoxi Li and Xueyang Tang and Jie ZHANG and Song Guo and Sikai Bai and Peiran Dong and Yue Yu},
booktitle={The Thirteenth International Conference on Learning Representations},
year={2025},
url={https://openreview.net/forum?id=yRKelogz5i}
}

@misc{openai2022chatgpt,
  author       = {OpenAI},
  title        = {{Introducing ChatGPT}},
  year         = {2022},
  url          = {https://openai.com/index/chatgpt/}
}

@article{sun2025and,
  title={{Why and How LLMs Hallucinate: Connecting the Dots with Subsequence Associations}},
  author={Sun, Yiyou and Gai, Yu and Chen, Lijie and Ravichander, Abhilasha and Choi, Yejin and Song, Dawn},
  journal={arXiv preprint arXiv:2504.12691},
  year={2025}
}

@article{jiang2023mistral,
  title={Mistral 7B},
  author={Jiang, Albert Q and Sablayrolles, Alexandre and Mensch, Arthur and Bamford, Chris and Chaplot, Devendra Singh and Casas, Diego de las and Bressand, Florian and Lengyel, Gianna and Lample, Guillaume and Saulnier, Lucile and others},
  journal={arXiv preprint arXiv:2310.06825},
  year={2023}
}

@inproceedings{marcu2024lingoqa,
  title={{LingoQA: Visual Question Answering for Autonomous Driving}},
  author={Marcu, Ana-Maria and Chen, Long and H{\"u}nermann, Jan and Karnsund, Alice and Hanotte, Benoit and Chidananda, Prajwal and Nair, Saurabh and Badrinarayanan, Vijay and Kendall, Alex and Shotton, Jamie and others},
  booktitle={European Conference on Computer Vision},
  pages={252--269},
  year={2024},
  organization={Springer}
}

@article{wang2024towards,
  title={{Towards Data-And Knowledge-Driven AI: A Survey
on Neuro-Symbolic Computing}},
  author={Wang, Wenguan and Yang, Yi and Wu, Fei},
  journal={IEEE Transactions on Pattern Analysis and Machine Intelligence},
  year={2024},
  publisher={IEEE}
}

@article{chaudhari2025rlhf,
  title={{RLHF Deciphered: A Critical Analysis of Reinforcement Learning from Human Feedback for LLMs}},
  author={Chaudhari, Shreyas and Aggarwal, Pranjal and Murahari, Vishvak and Rajpurohit, Tanmay and Kalyan, Ashwin and Narasimhan, Karthik and Deshpande, Ameet and Castro da Silva, Bruno},
  journal={ACM Computing Surveys},
  volume={58},
  number={2},
  pages={1--37},
  year={2025},
  publisher={ACM New York, NY}
}

@inproceedings{garouani2024investigating,
  title={{Investigating the Duality of Interpretability and Explainability in Machine Learning}},
  author={Garouani, Moncef and Mothe, Josiane and Barhrhouj, Ayah and Aligon, Julien},
  booktitle={2024 IEEE 36th International Conference on Tools with Artificial Intelligence (ICTAI)},
  pages={861--867},
  year={2024},
  organization={IEEE}
}

@article{li2026persona,
  title={{Persona Non Grata: Single-Method Safety Evaluation Is Incomplete for Persona-Imbued LLMs}},
  author={Li, Wenkai and Yang, Fan and Mehta, Shaunak A and Onoue, Koichi},
  journal={arXiv preprint arXiv:2604.11120},
  year={2026}
}

@article{ghandeharioun2024s,
  title={Who's asking? User personas and the mechanics of latent misalignment},
  author={Ghandeharioun, Asma and Yuan, Ann and Guerard, Marius and Reif, Emily and Lepori, Michael A and Dixon, Lucas},
  journal={Advances in Neural Information Processing Systems},
  volume={37},
  pages={125967--126003},
  year={2024}
}

@article{gyevnar2025ai,
  title={{AI safety for everyone}},
  author={Gyevnar, Balint and Kasirzadeh, Atoosa},
  journal={Nature Machine Intelligence},
  volume={7},
  number={4},
  pages={531--542},
  year={2025},
  publisher={Nature Publishing Group UK London}
}

@article{dubey2024nested,
  title={{A nested model for AI design and validation}},
  author={Dubey, Akshat and Yang, Zewen and Hattab, Georges},
  journal={IScience},
  volume={27},
  number={9},
  year={2024},
  publisher={Elsevier}
}

@inproceedings{hattab2024persona,
  title={Persona adaptable strategies make large language models tractable},
  author={Hattab, Georges and An{\v{z}}el, Aleksandar and Dubey, Akshat and Ezekannagha, Chisom and Yang, Zewen and Ilgen, Bahar},
  booktitle={Proceedings of the 2024 8th International Conference on Natural Language Processing and Information Retrieval},
  pages={24--31},
  year={2024}
}

@inproceedings{karny2026neural,
  title={{Neural Transparency: Mechanistic Interpretability Interfaces for Anticipating Model Behaviors for Personalized AI}},
  author={Karny, Sheer and Baez, Anthony and Pataranutaporn, Pat},
  booktitle={Proceedings of the 31st International Conference on Intelligent User Interfaces},
  pages={868--884},
  year={2026}
}

@inproceedings{zhou2026improving,
  title={{Improving Human Verification of LLM Reasoning through Interactive Explanation Interfaces}},
  author={Zhou, Runtao and Nguyen, Giang and Kharya, Nikita and Nguyen, Anh and Agarwal, Chirag},
  booktitle={Proceedings of the 31st International Conference on Intelligent User Interfaces},
  pages={456--473},
  year={2026}
}

@inproceedings{jeck2025tell,
  title={{TELL-ME: Toward Personalized Explanations of Large Language Models}},
  author={Jeck, Jakub and Leiser, Florian and H{\"u}sges, Anne and Sunyaev, Ali},
  booktitle={Proceedings of the Extended Abstracts of the CHI Conference on Human Factors in Computing Systems},
  pages={1--18},
  year={2025}
}

@inproceedings{mindlin2025towards,
  title={Towards Co-Constructed Explanations: A Multi-Agent Reasoning-Based Conversational System for Adaptive Explanations},
  author={Mindlin, Dimitry and Booshehri, Meisam and Cimiano, Philipp},
  booktitle={Proceedings of the 13th International Conference on Human-Agent Interaction},
  pages={148--157},
  year={2025}
}

@inproceedings{he2025conversational,
  title={{Is Conversational XAI All You Need? Human-AI Decision Making With a Conversational XAI Assistant}},
  author={He, Gaole and Aishwarya, Nilay and Gadiraju, Ujwal},
  booktitle={Proceedings of the 30th International Conference on Intelligent User Interfaces},
  pages={907--924},
  year={2025}
}

@article{alon2026faithfulness,
  title={{Faithfulness Serum: Mitigating the Faithfulness Gap in Textual Explanations of LLM Decisions via Attribution Guidance}},
  author={Alon, Bar and Zimerman, Itamar and Wolf, Lior},
  journal={arXiv preprint arXiv:2604.14325},
  year={2026}
}

@inproceedings{sui2024confabulation,
  title={Confabulation: The surprising value of large language model hallucinations},
  author={Sui, Peiqi and Duede, Eamon and Wu, Sophie and So, Richard},
  booktitle={Proceedings of the 62nd Annual Meeting of the Association for Computational Linguistics (Volume 1: Long Papers)},
  pages={14274--14284},
  year={2024}
}

@inproceedings{parcalabescu2024measuring,
  title={{On Measuring Faithfulness or Self-consistency
of Natural Language Explanations}},
  author={Parcalabescu, Letitia and Frank, Anette},
  booktitle={Proceedings of the 62nd Annual Meeting of the Association for Computational Linguistics (Volume 1: Long Papers)},
  pages={6048--6089},
  year={2024}
}

@article{naseem2026mechanistic,
  title={{Mechanistic Interpretability for Large Language Model Alignment: Progress, Challenges, and Future Directions}},
  author={Naseem, Usman},
  journal={arXiv preprint arXiv:2602.11180},
  year={2026}
}

@article{somvanshi2026bridging,
  title={Bridging the Black Box: A Survey on Mechanistic Interpretability in AI},
  author={Somvanshi, Shriyank and Islam, Md Monzurul and Rafe, Amir and Tusti, Anannya Ghosh and Chakraborty, Arka and Baitullah, Anika and Chowdhury, Tausif Islam and Alnawmasi, Nawaf and Dutta, Anandi and Das, Subasish},
  journal={ACM Computing Surveys},
  volume={58},
  number={8},
  pages={1--35},
  year={2026},
  publisher={ACM New York, NY}
}

@inproceedings{ferrario2022explainability,
  title={{How Explainability Contributes to Trust in AI}},
  author={Ferrario, Andrea and Loi, Michele},
  booktitle={Proceedings of the 2022 ACM Conference on Fairness, Accountability, and Transparency},
  pages={1457--1466},
  year={2022}
}

@inproceedings{jacovi2021formalizing,
  title={{Formalizing Trust in Artificial Intelligence: Prerequisites, Causes and Goals of Human Trust in AI}},
  author={Jacovi, Alon and Marasovi{\'c}, Ana and Miller, Tim and Goldberg, Yoav},
  booktitle={Proceedings of the 2021 ACM Conference on Fairness, Accountability, and Transparency},
  pages={624--635},
  year={2021}
}

@article{afroogh2024trust,
  title={{Trust in AI: progress, challenges, and future directions}},
  author={Afroogh, Saleh and Akbari, Ali and Malone, Emmie and Kargar, Mohammadali and Alambeigi, Hananeh},
  journal={Humanities and Social Sciences Communications},
  volume={11},
  number={1},
  pages={1568},
  year={2024},
  publisher={Palgrave}
}

@article{miller2019explanation,
  title={Explanation in artificial intelligence: Insights from the social sciences},
  author={Miller, Tim},
  journal={Artificial intelligence},
  volume={267},
  pages={1--38},
  year={2019},
  publisher={Elsevier}
}

@article{lindsey2025biology,
  title        = {On the Biology of a Large Language Model},
  author       = {Jack Lindsey and Wes Gurnee and Emmanuel Ameisen and Brian Chen and Adam Pearce and Nicholas L. Turner and Craig Citro and David Abrahams and Shan Carter and Basil Hosmer and Jonathan Marcus and Michael Sklar and Adly Templeton and Trenton Bricken and Callum McDougall and Hoagy Cunningham and Thomas Henighan and Adam Jermyn and Andy Jones and Andrew Persic and Zhenyi Qi and T. Ben Thompson and Sam Zimmerman and Kelley Rivoire and Thomas Conerly and Chris Olah and Joshua Batson},
  journal      = {Transformer Circuits},
  year         = {2025},
  month        = {March},
  url          = {https://transformer-circuits.pub/2025/attribution-graphs/biology.html},
  publisher    = {Anthropic}
}

@article{chen2025persona,
  title={{Persona Vectors: Monitoring and Controlling Character Traits in Language Models
}},
  author={Chen, Runjin and Arditi, Andy and Sleight, Henry and Evans, Owain and Lindsey, Jack},
  journal={arXiv preprint arXiv:2507.21509},
  year={2025}
}

@article{lindsey2025introspection,
  title        = {{Emergent Introspective Awareness in Large Language Models}},
  author       = {Lindsey, Jack},
  year         = {2025},
  journal={Transformer Circuits Thread},
  url          = {https://transformer-circuits.pub/2025/introspection/index.html}
}

@article{ahmad2026beyond,
  title={{Beyond Components: Singular Vector-Based Interpretability of Transformer Circuits}},
  author={Ahmad, Areeb and Joshi, Abhinav and Modi, Ashutosh},
  journal={Advances in Neural Information Processing Systems},
  volume={38},
  pages={167772--167811},
  year={2025}
}

@inproceedings{im2026how,
title={How Do Transformers Learn to Associate Tokens: Gradient Leading Terms Bring Mechanistic Interpretability},
author={Shawn Im and Changdae Oh and Zhen Fang and Sharon Li},
booktitle={The Fourteenth International Conference on Learning Representations},
year={2026}
}

@inproceedings{golimblevskaia2026circuit,
title={{Circuit Insights: Towards Interpretability Beyond Activations}},
author={Elena Golimblevskaia and Aakriti Jain and Bruno Puri and Ammar Ibrahim and Wojciech Samek and Sebastian Lapuschkin},
booktitle={The Fourteenth International Conference on Learning Representations},
year={2026}
}

@inproceedings{chen2026decomposing,
title={{Decomposing {LLM} Computation with Jets}},
author={Yihong Chen and Xiangxiang Xu and Pontus Stenetorp and Sebastian Riedel and Luca Franceschi},
booktitle={The Fourteenth International Conference on Learning Representations},
year={2026}
}

@article{fraser-taliente2026nla,
  title   = {{Natural Language Autoencoders Produce Unsupervised Explanations of LLM Activations}},
  author  = {Kit Fraser-Taliente and Subhash Kantamneni and Euan Ong and Dan Mossing and Christina Lu and Paul C. Bogdan and Emmanuel Ameisen and James Chen and Dzmitry Kishylau and Adam Pearce and Julius Tarng and Alex Wu and Jeff Wu and Yang Zhang and Daniel M. Ziegler and Evan Hubinger and Joshua Batson and Jack Lindsey and Samuel Zimmerman and Samuel Marks},
  booktitle = {Transformer Circuits Thread},
  year    = {2026},
}

@inproceedings{chen2025towards,
  title={{Towards Consistent Natural-Language Explanations via Explanation-Consistency Finetuning}},
  author={Chen, Yanda and Singh, Chandan and Liu, Xiaodong and Zuo, Simiao and Yu, Bin and He, He and Gao, Jianfeng},
  booktitle={Proceedings of the 31st International Conference on Computational Linguistics},
  pages={7558--7568},
  year={2025}
}

@inproceedings{chuang2026faithlm,
  title={{FaithLM: Towards Faithful Explanations for Large Language Models}},
  author={Chuang, Yu-Neng and Wang, Guanchu and Chang, Chia-Yuan and Tang, Ruixiang and Zhong, Shaochen and Yang, Fan and Wen, Andrew and Du, Mengnan and Cai, Xuanting and Braverman, Vladimir and others},
  booktitle={Proceedings of the 19th Conference of the European Chapter of the Association for Computational Linguistics (Volume 1: Long Papers)},
  pages={3802--3824},
  year={2026}
}

@inproceedings{chen2026does,
  title={{How Does Chain of Thought Think? Mechanistic Interpretability of Chain-of-Thought Reasoning with Sparse Autoencoding}},
  author={Chen, Xi and Plaat, Aske and van Stein, Niki},
  booktitle={Proceedings of the AAAI Conference on Artificial Intelligence},
  volume={40},
  number={36},
  pages={30297--30305},
  year={2026}
}

@article{zheng2023judging,
  title={{Judging LLM-as-a-Judge with MT-Bench and Chatbot Arena}},
  author={Zheng, Lianmin and Chiang, Wei-Lin and Sheng, Ying and Zhuang, Siyuan and Wu, Zhanghao and Zhuang, Yonghao and Lin, Zi and Li, Zhuohan and Li, Dacheng and Xing, Eric and others},
  journal={Advances in Neural Information Processing Systems},
  volume={36},
  pages={46595--46623},
  year={2023}
}

@misc{alpaca_eval,
  author = {Xuechen Li and Tianyi Zhang and Yann Dubois and Rohan Taori and Ishaan Gulrajani and Carlos Guestrin and Percy Liang and Tatsunori B. Hashimoto },
  title = {{AlpacaEval: An Automatic Evaluator of Instruction-following Models}},
  year = {2023},
  publisher = {GitHub},
  journal = {GitHub repository},
  howpublished = {\url{https://github.com/tatsu-lab/alpaca_eval}}
}

@article{bengio2025international,
  title={{International AI Safety Report}},
  author={Bengio, Yoshua and Mindermann, S{\"o}ren and Privitera, Daniel and Besiroglu, Tamay and Bommasani, Rishi and Casper, Stephen and Choi, Yejin and Fox, Philip and Garfinkel, Ben and Goldfarb, Danielle and others},
  journal={arXiv preprint arXiv:2501.17805},
  year={2025}
}

@inproceedings{basu2025on,
title={{On Mechanistic Circuits for Extractive Question-Answering}},
author={Samyadeep Basu and Vlad I Morariu and Ryan A. Rossi and Nanxuan Zhao and Zichao Wang and Soheil Feizi and Varun Manjunatha},
booktitle={Second Conference on Language Modeling},
year={2025}
}

\end{document}